\documentclass[letterpaper]{article} 
\usepackage{aaai2026}  
\usepackage{times}  
\usepackage{helvet}  
\usepackage{courier}  
\usepackage[hyphens]{url}  
\usepackage{graphicx} 
\usepackage{natbib}  
\usepackage{caption} 
\usepackage{algorithm}
\usepackage{algorithmic}

\usepackage{algorithm}
\usepackage{algorithmic}
\usepackage{times}
\usepackage{latexsym}
\usepackage{amsmath}
\usepackage{booktabs}
\usepackage{soul}

\usepackage[T1]{fontenc}

\usepackage{xspace}
\usepackage{graphicx}

\def\onedot{\ifx\@thefnmark\relax\else.\fi\xspace}
\def\eg{\emph{e.g}\onedot} 
\def\ie{\emph{i.e}\onedot}

\def\etal{\emph{et al}\onedot}

\usepackage{graphicx}
\usepackage{latexsym}
\usepackage{xcolor}
\usepackage{booktabs}
\usepackage{amsmath}
\usepackage{amsfonts} 
\usepackage{amssymb} 
\usepackage{graphicx}  
\usepackage{subfig}    
\usepackage{soul}
\usepackage{multirow}
\definecolor{cvprblue}{rgb}{0.21,0.49,0.74}

\usepackage[table]{xcolor}    
\usepackage{colortbl}         
\usepackage{pgf}     

\usepackage{pgf} 
\usepackage[table]{xcolor}   
\usepackage{colortbl}        
\usepackage{pgf}             

\usepackage{mdframed}
\usepackage{xcolor}

\newcounter{finding}

\usepackage[table]{xcolor}  
\usepackage{pgf}    

\definecolor{level0}{RGB}{255,245,240}  
\definecolor{level1}{RGB}{254,230,220}  
\definecolor{level2}{RGB}{254,215,200}  
\definecolor{level3}{RGB}{253,190,180}  
\definecolor{level4}{RGB}{252,165,160}  
\definecolor{level5}{RGB}{251,140,130}  
\definecolor{level6}{RGB}{244,110,100}  
\definecolor{level7}{RGB}{230,80,80}    

\definecolor{green}{RGB}{127,255,127}
\definecolor{mediumgreen}{RGB}{63,167,63}
\definecolor{darkgreen}{RGB}{0,127,0}
\definecolor{blue}{RGB}{127,127,255}
\definecolor{lightblue}{RGB}{150,150,255}

\usepackage{xspace}

\makeatletter
\DeclareRobustCommand\onedot{\futurelet\@let@token\@onedot}
\def\@onedot{\ifx\@let@token.\else.\null\fi\xspace}

\def\eg{\emph{e.g}\onedot} 
\def\ie{\emph{i.e}\onedot}

\def\etal{\emph{et al}\onedot}
\makeatother

\newcommand{\inlinefinding}[1]{\begingroup\setlength{\fboxsep}{1pt}\colorbox{gray!20}{\textbf{#1}}\endgroup}

\newcommand{\colorBinsCell}[1]{%
    \pgfmathparse{#1}%
    \ifdim \pgfmathresult pt < 0.2pt
        \cellcolor{level0}{#1}%
    \else
        \ifdim \pgfmathresult pt < 0.4pt
            \cellcolor{level1}{#1}%
        \else
            \ifdim \pgfmathresult pt < 0.6pt
                \cellcolor{level2}{#1}%
            \else
                \ifdim \pgfmathresult pt < 0.8pt
                    \cellcolor{level3}{#1}%
                \else
                    \cellcolor{level4}{#1}%
                \fi
            \fi
        \fi
    \fi
}

\usepackage{mdframed}
\usepackage{xcolor}
\newcounter{Prompt}

\usepackage{newfloat}
\usepackage{listings}
\DeclareCaptionStyle{ruled}{labelfont=normalfont,labelsep=colon,strut=off} 
\floatstyle{ruled}
\newfloat{listing}{tb}{lst}{}
\floatname{listing}{Listing}
\title{The Confidence Trap: Gender Bias and Predictive
Certainty in LLMs}

\author{
  Ahmed Sabir\textsuperscript{\rm 1}, 
  Markus K\"angsepp\textsuperscript{\rm 1}, 
  Rajesh Sharma\textsuperscript{\rm 1,\rm 2}
}

\affiliations{
  \textsuperscript{\rm 1}Institute of Computer Science, University of Tartu, Estonia\\
  \textsuperscript{\rm 2}School of AI and CS, Plaksha University, India\\
}

\usepackage{bibentry}

\begin{document}

\maketitle

\begin{abstract}
The increased use of Large Language Models (LLMs) in sensitive domains leads to growing interest in how their confidence scores correspond to fairness and bias. This study examines the alignment between LLM-predicted confidence and human-annotated bias judgments. Focusing on gender bias, the research investigates probability confidence calibration in contexts involving gendered pronoun resolution. The goal is to evaluate if calibration metrics based on predicted confidence scores effectively capture fairness-related disparities in LLMs. The results show that, among the six state-of-the-art models, Gemma-2 demonstrates the worst calibration according to the gender bias benchmark. The primary contribution of this work is a fairness-aware evaluation of LLMs' confidence calibration, offering guidance for ethical deployment. In addition, we introduce a new calibration metric, Gender-ECE, designed to measure gender disparities in resolution tasks. Our code is publicly available at \url{https://github.com/ahmedssabir/GECE}.

\end{abstract}

\section{Introduction}

Large pretrained LLMs have demonstrated exceptional performance across a variety of tasks, yet their trustworthiness remains a pressing concern, especially regarding gender bias. These models not only inherit biases from their training data but can also amplify stereotypes, leading to unfair or inaccurate outputs in critical applications  \cite{zhao2018gender,rudinger-EtAl:2018:N18,nangia2020crows, felkner2023winoqueer, gallegos2024bias,sabir2025exploring}. The challenge lies not only in detecting bias but also in ensuring that users can reliably interpret and trust model predictions, particularly in high-stakes domains such as hiring, healthcare, and legal decision-making.

Calibration, which ensures that an LLM’s confidence in its predictions accurately reflects the likelihood of being correct \cite{niculescu2005predicting,si2022re}, is a crucial aspect of model reliability. Without proper calibration, an LLM might provide overconfident yet incorrect outputs, misleading users into trusting biased or unreliable responses. This issue becomes even more critical in cases where models exhibit skewed confidence distributions across different demographic groups, further exacerbating fairness concerns. Therefore, in this work, we aim to answer the question regarding calibration in a gender bias scenario. Here, calibration refers to how well a model’s probability estimates match reality. For instance, when a model assigns an 80\% probability to its prediction (\eg for the pronoun 'he'), then, ideally, for those 80\% confidence predictions, the model should be correct approximately 80\% of the time.

The importance of calibration is particularly relevant in gender bias tasks. In many real-world settings, the objective is not only to measure which pronoun best matches human bias but also to assess the model’s confidence in its predictions. Poor calibration means that the model might generate seemingly “confident” predictions even when it is frequently incorrect, posing a significant risk when deploying these models in production or high-stakes scenarios. In addition, advancing the understanding of fairness in the context of misrepresented groups contributes to a more comprehensive evaluation of model behavior \cite{ding2024fairness}. 

While many studies have explored the biases and stereotypes propagated by LLMs \cite{gallegos2024bias}, little to no research has examined how well these models are calibrated in their predictions or how closely they align with human biases. In this work, we investigate these gaps by analyzing the confidence calibration of LLMs in gendered and gender-neutral contexts. Specifically, inspired by model calibration evaluation methods such as Expected Calibration Error (ECE) \cite{guo2017calibration}, which help evaluate the model’s predicted confidence with its actual outcomes, we seek to answer the following question: To what extent is the predictive confidence of LLMs calibrated in gendered pronoun resolution tasks?  

To summarize our contributions: (1) we conduct a comprehensive analysis of widely used open-weight LLMs, offering practical guidance for their ethical deployment, and (2) we propose Gender-ECE, a novel adaptation of ECE designed to quantify gender-related calibration disparities in coreference resolution tasks.

\section{Related Work}

\noindent{\bf{Gender Bias in LLM.}} Large language models have demonstrated remarkable capabilities in natural language understanding and generation. However, they also inherit and amplify societal biases, including gender bias. To measure Gender bias in downstream tasks in NLP applications, a variety of research efforts have investigated gender bias using templates based on structured sentences such as "He/She is a/an [occupation]," in which blanks are filled with occupations that convey either positive or negative stereotypes or associations \cite{Staczak2021ASO}.

Additional research has been conducted using the format of Winograd Schemas \cite{levesque2012winograd}. This method has been applied to various datasets such as WinoBias \cite{zhao2018gender}, Winogender \cite{rudinger-EtAl:2018:N18}, and WinoMT \cite{stanovsky2019evaluating}. The Winograd Schema Challenge focuses on the task of coreference resolution, which requires common-sense reasoning. This challenge has been used to examine whether the interpretation of pronoun references in sentences is influenced by gender. This method has also been applied to evaluate gender-based stereotypes and neutral associations across various professions \cite{zhao2018gender,rudinger-EtAl:2018:N18, sabir2025exploring}.  More recently, most research has shifted beyond template-based approaches to address gender bias by evaluating the outputs of LLMs, which can encode and propagate societal stereotypes across various contexts, including personality traits, social roles, and cultural representations. For example, the work of Cheng \etal \cite{cheng2023marked} employs a method to assess stereotypes in LLMs by generating persona descriptions for various demographic groups. Extending this direction, prompt-driven frameworks such as RUTEd \cite{lum2025bias} measure bias by instructing LLMs to perform in-context tasks \eg writing stories or adopting personas, shifting evaluation from short template-based sentences to longer-form generations that better surface bias.

However, the most dominant method to measure bias in LLMs is the use of template-based methods, where controlled sentence structures allow for systematic evaluation of gender bias in model predictions. These templates help isolate bias by ensuring that the only difference between sentence variations is the gender-related attribute while keeping all other linguistic components identical. In this study, we adopt a template-based approach.

\noindent{\textbf{Calibration in LLMs.}} Large language models are stated to be rather well-calibrated~\cite{kadavath2022language}. Mostly, the studies involve using question-answer datasets with multiple answers \cite{krause2023confidently, kapoor2024calibration, yoon2025reasoning}. There are many ways to check how well the model is calibrated, for example, using (1) prompting, (2) model token probabilities, and (3) training a model on top of the model's internal state or output. Firstly, many works on prompting~\cite{kadavath2022language, xiong2023can, tian2023just, yoon2025reasoning} have relied solely on the model’s ability, whether a LLM or a reasoning model, to convey confidence based on their answers or chain of thought. For example, Kadavath \etal ~\cite{kadavath2022language} query a model about its answer confidence $P(true)$ after the model has answered a question in a question-answering setting. Secondly, token probabilities are a straightforward way to determine the model's uncertainty for each token.
However, these probabilities might not be well-calibrated, for example, for low-resource languages~\cite{krause2023confidently} in the case of multilingual models. Also, having bigger or pretrained and fine-tuned models does not grant better calibrated models~\cite{chen2022close}.  Lastly, training an extra model on top of the LLM to assess the model's confidence more precisely. For example, \cite{kadavath2022language} proposes to train models with an extra head to predict the probability that a model knows $P(IK)$ the answer. More recently, the work of \cite{kapoor2024calibration} proposes a fine-tuning protocol to calibrate model confidences in question-answer tasks.

In this work, we are interested in token probabilities, more precisely, how well a model's probabilities for gender bias align with human bias. Thus, extracting the probability for pronouns and using simple post-hoc calibration methods suffices for our analysis.

\section{Datasets} 

\noindent{\textbf{WinoBias \cite{zhao2018gender}.}} WinoBias is a benchmark dataset for evaluating gender bias in LLMs. The dataset contains 3,160 sentences using templates inspired by Winograd-schema style sentences  \cite{levesque2012winograd} with entities corresponding to people referred by their occupation (\textit{e.g.} doctor). For each sentence, there are three variables person, occupation, and pronoun. The dataset is structured to test whether LLMs make gendered assumptions about professions by evaluating coreference resolution performance, \eg "The developer argued with the designer and slapped her in the face".

\noindent{\textbf{Winogender \cite{rudinger-EtAl:2018:N18}.}} Winogender is a dataset similar to WinoBias, which also includes template sentences that refer to occupation and person. The dataset contains 720 template-based sentences. Unlike WinoBias,  Winogender introduced a second entity to the gendered occupation, using gender neutral \texttt{someone} to avoid gender stereotype association. Each sentence in Winogender is carefully designed to evaluate coreference resolution accuracy in a way that minimizes unintended gender stereotypes. 

\noindent{\textbf{GenderLex \cite{sabir2025exploring}.}}  A similar template-based dataset to WinoBias and Winogender contains pronouns and biased occupations but also includes nouns and verbs directly related to the pronouns. This dataset uses ChatGPT 3.5-turbo \cite{OpenAI} to generate 1,676 sentence pairs (838 unique sentences), followed by human annotation and correction. However, unlike previous datasets, this template relies on the \texttt{last cloze} structure, \eg "The lawyer stated that the legal case was won by him". This approach, adding the pronoun at the end, mimics how humans and LLMs generate text, postpones gender clues, and therefore provides a strong method to probe gender bias in LLMs.

\section{Methods}
The main idea of this work is to understand whether the model’s LLM confidence (probability estimates) aligns with model prediction and its human alignment  (\ie human bias). In other words, we aim to evaluate the extent to which the model’s self-assessed confidence (\ie token probability) reflects both the correctness of predictions and the degree of correlation between these confidence levels and human judgment or biases, within the context of gender bias research. By examining this alignment, we can assess the extent to which the model’s probability estimates serve as reliable indicators of actual performance in detecting, mitigating, or perpetuating gender bias.

To do this, for a given sentence with a pronoun, we extract the probability confidence of the model assigning to different pronouns. For example, in "The surgeon called the nurse and instructed [her/him] to prepare for surgery.", the "her" pronoun is more commonly expected here, which reflects occupational gender bias. However, we also evaluate the model’s confidence when replacing the pronoun with "him" and check how confident the model is in its prediction. The idea is to evaluate the model's confidence by examining predictions involving a given pronoun and a stereotypically biased occupation (\eg nurse). Therefore, the aim is to investigate the model's certainty in making biased predictions.

Let a sentence $S$ contain $T$ words, $S = (w_1, w_2, ..., w_T)$, If a pronoun $w_p$ in the given sentence occurs at position $k$ (\ie $w_k = w_p$), 
the probability is computed as follows:

\begin{equation}
P(w_p \mid w_1, ..., w_{k-1}) = \frac{e^{z_{k-1, w_p}}}{\sum_{j=1}^{V} e^{z_{k-1, j}}},
\end{equation} 

\noindent{where} $z_{k-1, w_p}$  is the model logit score for the pronoun $w_{p}$ at the position $k$.  After extracting the probability, we investigate model's confidence using different calibration metrics:

\noindent{\textbf{Expected Calibration Error (ECE).}} ECE~\cite{naeini2015obtaining} measures how well the model-predicted confidence matches with the actual outcome and accuracy of the model.  The model predictions are divided into $M$ bins $B_m$.
Bins can be divided into equal-width bins (each bin takes an equal amount of probability space) or equal-size bins (each bin has an equal number of instances).
For each bin, the average confidence $\text{conf}(B_m)$ and average accuracy $\text{acc}(B_m)$ are calculated, and ECE is the average absolute bin-wise difference between average confidence and average accuracy. More precisely,

\begin{equation}
\text{ECE} = \sum_{m=1}^{M} \frac{|B_m|}{n} \left| \text{acc}(B_m) - \text{conf}(B_m) \right|,
\end{equation}
where $n$ is the total number of instances. The smaller the ECE score, the more trustworthy the model’s confidence (\ie confidences reflect reality).

\begin{figure*}[ht]
    \centering
    \subfloat[ECE: \underline{0.076}]{\includegraphics[width=0.20\textwidth]{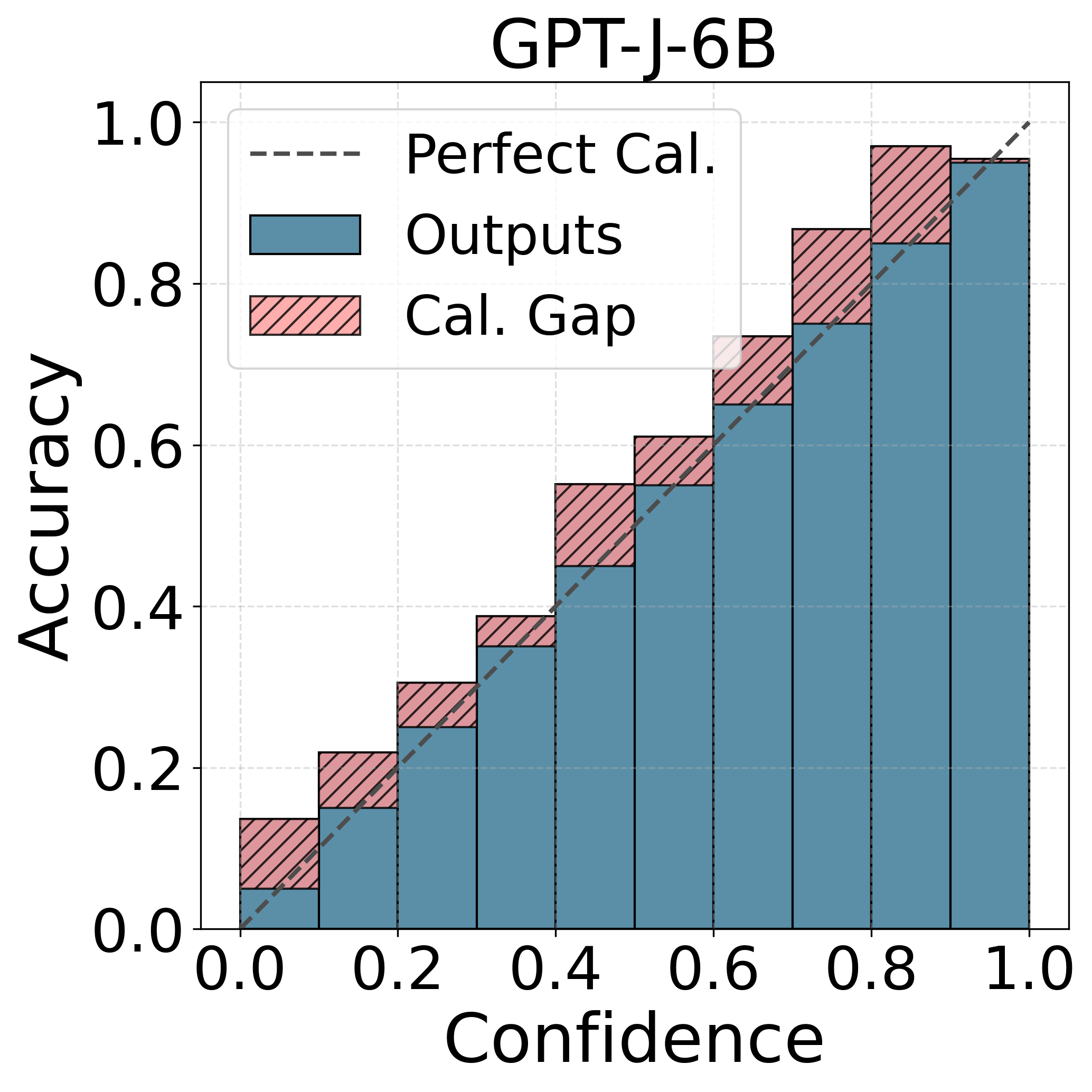}}\hfill
   \subfloat[ECE: 0.111]{\includegraphics[width=0.20\textwidth]{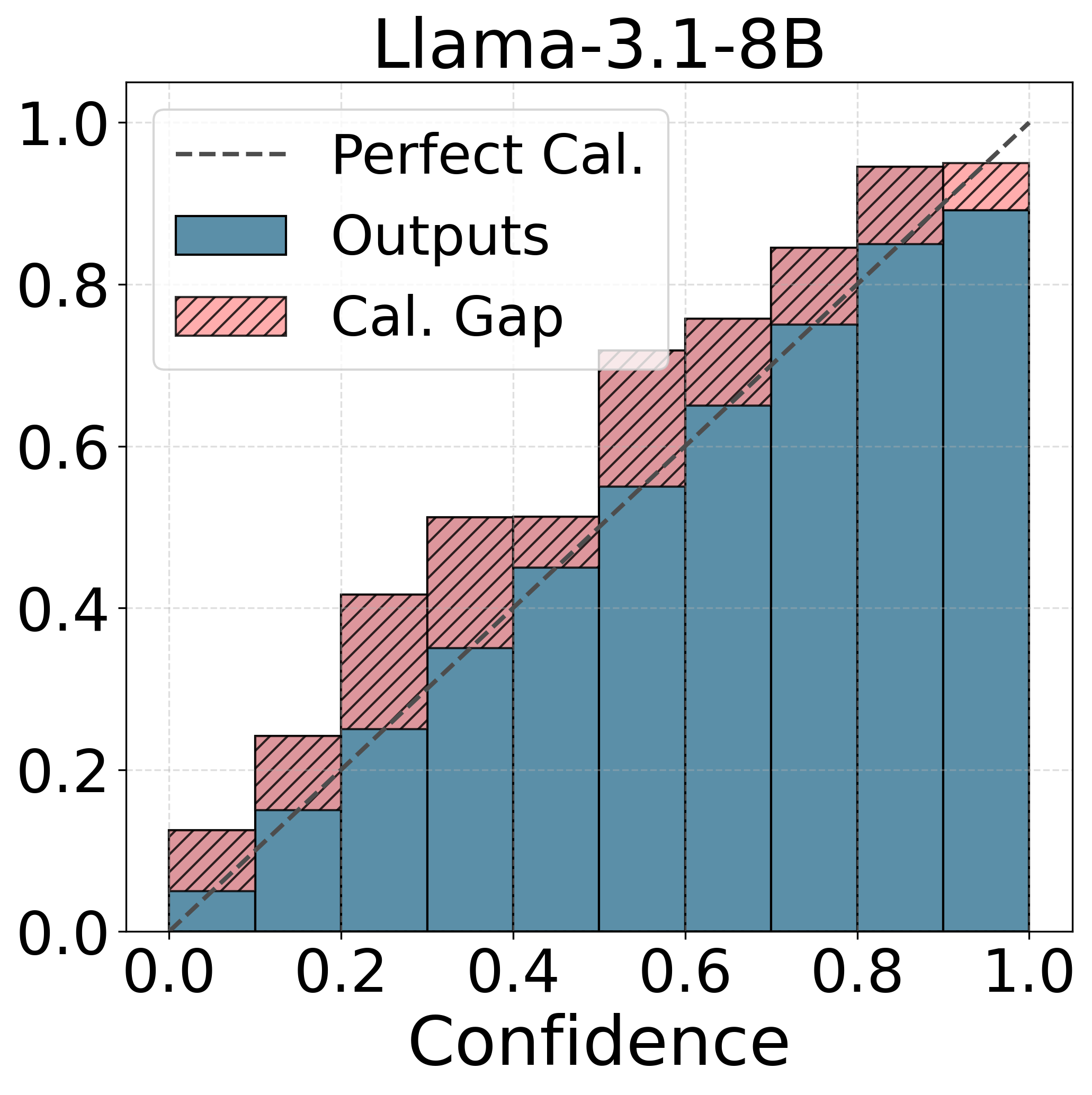}}\hfill
  \subfloat[ECE: 0.085]{\includegraphics[width=0.20\textwidth]{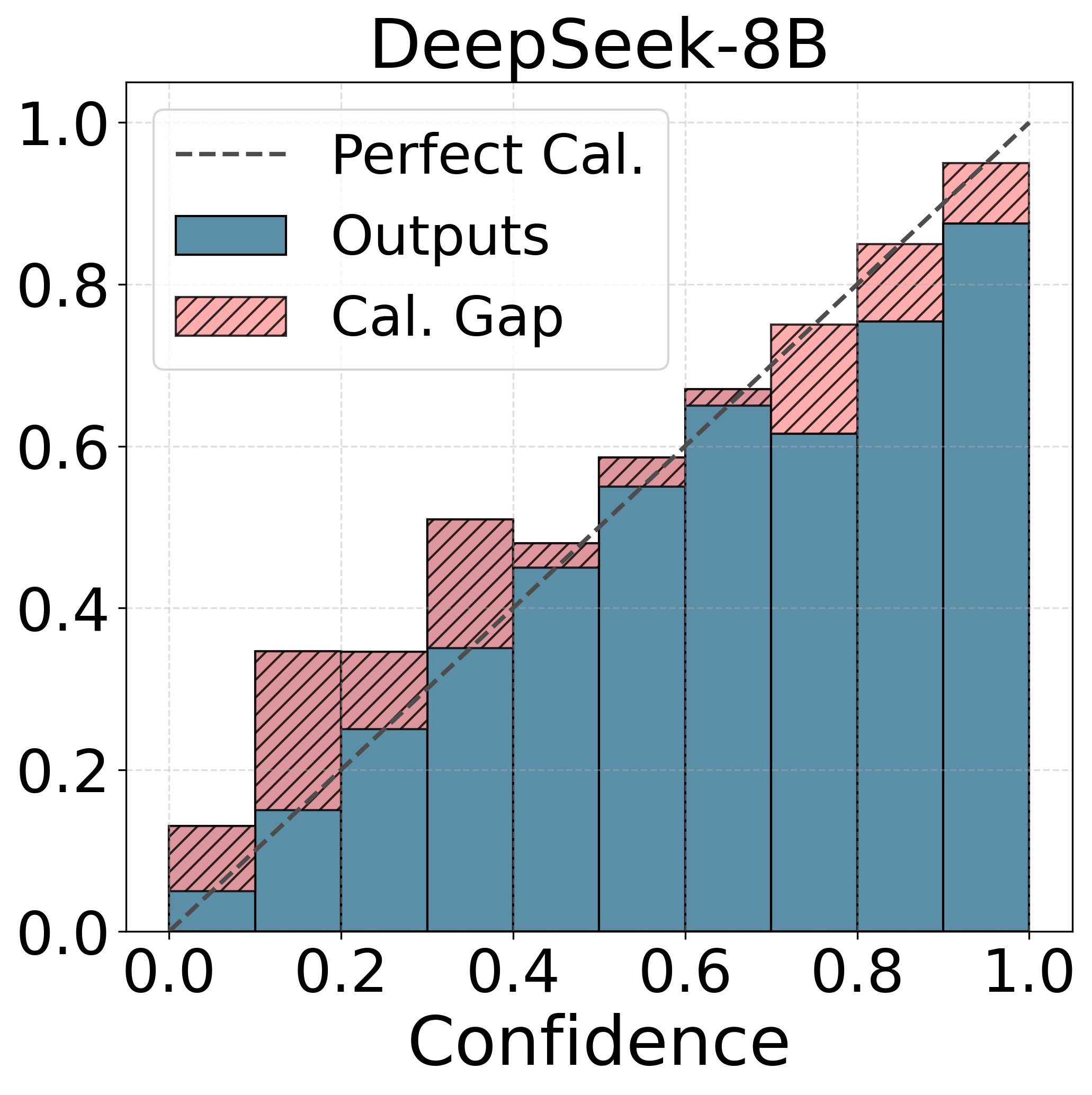}}\hfill
    \subfloat[ECE: \textbf{0.327}]{\includegraphics[width=0.20\textwidth]{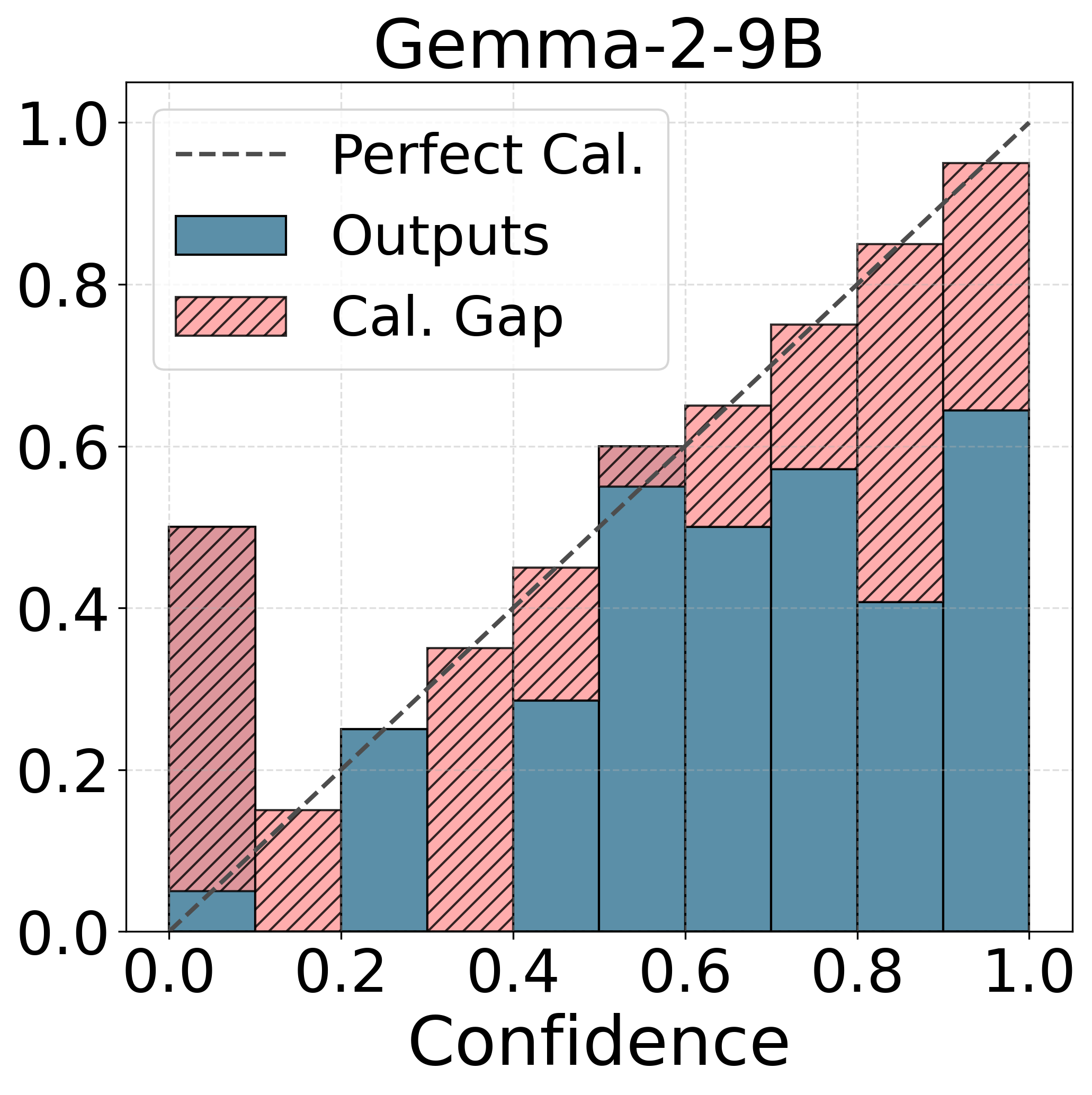}}\hfill
     \subfloat[ECE: 0.106]{\includegraphics[width=0.20\textwidth]{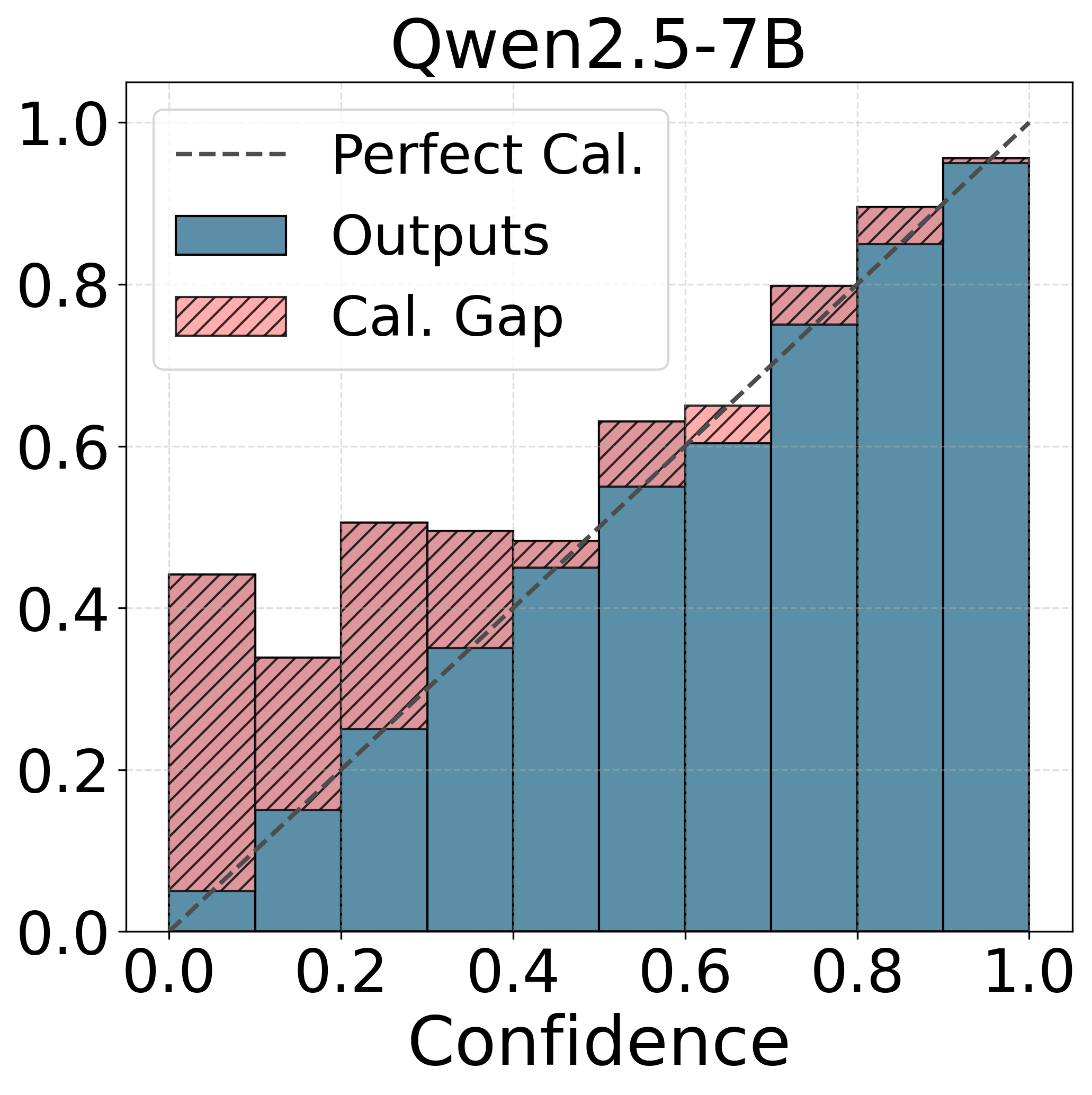}}
       \caption{Reliability diagrams for the GenderLex dataset with the gender appearing at the end of the sentence having all context \texttt{last cloze} structure. Gemma-2-9B exhibits the \textbf{highest} ECE while GPT-J-6B is \underline{the most calibrated model.} }
    \label{fig:GenderVocab-reliability diagrams}
\end{figure*}

\noindent{\textbf{Reliability Diagrams.}} Reliability diagrams~\cite{murphy1977reliability} are a visualization method based on ECE (\eg Figure \ref{fig:GenderVocab-reliability diagrams}). Each bin $B_m$ is plotted on the y-axis based on the confidence. The blue bar shows the average accuracy of a bin $\text{acc}(B_m)$, visible on the x-axis. The red gap shows the difference between the average accuracy $\text{acc}(B_m)$ and average confidence $\text{conf}(B_m)$. Indicating the expected calibration error in the bin $B_m$. Ideally, the blue bars should be aligned with the dashed diagonal line to have a well-calibrated model (perfect calibration).

\noindent{\textbf{Instance Calibration Error (ICE).}} ICE~\cite{si2022re} measures the mean absolute difference between model confidence $\hat{p}_i$ and true label $y_i$ for each instance $i$. 
\begin{equation}
    \text{ICE} = \frac{1}{n} \sum_{i=1}^{n} \left| y_i - \hat{p}_i \right|.
\end{equation}

\noindent{\textbf{Macro Calibration Error (MacroCE).}} MacroCE~\cite{si2022re} is similar to ICE. Except that it divides the instances into two subgroups, correct and incorrect predictions, and ICE is separately calculated for these subgroups. Finally, the average of $\text{ICE}_{pos}$ and $\text{ICE}_{neg}$ is taken.
Formally:
\begin{equation}
    \begin{split}
    \text{MacroCE} &= \frac{1}{2}(\text{ICE}_{pos} + \text{ICE}_{neg}), \\
    \text{ICE}_{pos} &= \frac{1}{n_p} \sum_{i=1}^{n_p}(1 - \hat{p}_i), \forall \hat{y}_i = y_i, \\
    \text{ICE}_{neg} &= \frac{1}{n_n} \sum_{i=1}^{n_n}(\hat{p}_i - 0), \forall \hat{y}_i \neq y_i,
    \end{split}
\end{equation}
where $n_p$ and $n_n$ are number of positive and negative instances, $\hat{p}_i$ and $\hat{y}_i$ are predicted probability and predicted class-label of $i$th instance.

\noindent{{\textbf{Brier Score}.}} Brier score~\cite{brier1950verification} is a proper scoring rule that measures the mean squared difference between model predicted confidence $\hat{p}_i$ and true label $y_i$ for each instance $i$. Brier score is commonly used as a loss measure for training machine learning models. The Brier score is defined as:

\begin{equation}
    \text{Brier Score} = \frac{1}{n} \sum_{i=1}^{n} ( \hat{p}_i - y_i )^2.
\end{equation}

\begin{table*}[t]
\centering
\renewcommand{\arraystretch}{1.2}
\setlength{\tabcolsep}{6pt}
\resizebox{0.8\textwidth}{!}{%
\begin{tabular}{lcccccccc}
\toprule
 & \multicolumn{4}{c}{\textbf{Standard Calibration Metrics}} & \multicolumn{3}{c}{\textbf{Gender-ECE}} & \textbf{Human} \\
\cmidrule(lr){2-5} \cmidrule(lr){6-8}
\textbf{Model} 
& \textbf{ECE} 
& \textbf{MacroCE} 
& \textbf{ICE} 
& \textbf{Brier Score} 
& \textbf{Group} 
& \textbf{M} 
& \textbf{F} 
&  \\
\midrule

GPT-J-6B \cite{gpt-j} 
& \underline{0.076} 
& \underline{0.453} 
& 0.374 & 0.432 & \underline{0.076} & 0.085 & \underline{0.066} & 0.715 \\
Llama-3.1-8B \cite{llama3modelcard} 
& 0.111 
& 0.466 
& \underline{0.371} 
& 0.446 
& 0.111 
& 0.112 
& 0.109 
& \textbf{0.727} \\
Gemma-2-9B \cite{team2024gemma}
& \textbf{0.327} 
& \textbf{0.493} 
& 0.390 
& \textbf{0.559} 
& \textbf{0.267}
& \textbf{0.330} 
& 0.204
& 0.617 \\
Qwen2.5-7B \cite{qwen2}
& 0.106 
& 0.476 
& 0.422 
& 0.385 
& 0.107 
& \underline{0.052} 
& 0.162 
& 0.637 \\
Falcon3-7B \cite{falcon3} 
& 0.161 
& 0.491 
& \textbf{0.449} 
& \underline{0.356} 
& 0.149 
& 0.081 
& \textbf{0.217} 
& \underline{0.605} \\
DeepSeek-8B \cite{deepseekai2025} 
& 0.085 & 0.461 & 0.369 & 0.470 & 0.090 & 0.074 & 0.106 & 0.686 \\

\bottomrule
\end{tabular}
}

\caption{Comparison of calibration metrics across models on the GenderLex dataset. GPT-J-6B shows the best calibration (lowest ECE and balanced Gender-ECE). Gemma-2-9B performs the worst overall. \textbf{Bold} = highest (worst calibration and best human alignment), \underline{Underlined} = lowest (best calibration and worst human alignment).}
\label{tab:genderlex_gender_split}
\end{table*}

\noindent{{\textbf{Gender-Aware ECE}.}} All previous metrics for evaluating calibration in LLMs focus on comparing two gendered sentence completions, one with a male pronoun and one with a female pronoun. However, overall calibration, such as ECE, does not reveal how the model behaves differently with respect to male versus female pronouns.  To address this gap, we propose Gender-Aware Group-ECE, a metric designed to capture calibration disparities across gendered pronouns.

\begin{equation}
{\small
\text{Gender-ECE} = \frac{1}{2}\left(\text{ECE}_{\text{male}} + \text{ECE}_{\text{female}}\right)
}
\label{eq:gender-ece}
\end{equation}
\begin{equation*}
{\small
\begin{aligned}
\text{ECE}_{\text{male}}   &= \sum_{m=1}^{M} \frac{|B_m|}{n} \left|\text{acc}(B_m) - \text{conf}(B_m)\right|, && \forall \hat{y}_i = 1, \\
\text{ECE}_{\text{female}} &= \sum_{m=1}^{M} \frac{|B_m|}{n} \left|\text{acc}(B_m) - \text{conf}(B_m)\right|, && \forall \hat{y}_i = 0.
\end{aligned}
}
\end{equation*}

\noindent{where} $\hat{y}_i$ is predicted class, and [1/0] correspond to male/female label. Gender-ECE is conceptually inspired by MacroCE, however, in MacroCE the subdivision into groups (positive and negative) is based on whether prediction was correct or incorrect. Therefore, in the positive group are all the instances where model prediction is aligned with human bias (true label). On the other hand, Gender-ECE divides data into two groups (\ie male and female) based on the predicted label, not relying on the true label for the division. By doing this, it shows how well-calibrated the model's scores are compared to the human bias for gender pronouns. In addition, MacroCE is calculated instance-wise, rather than bin-wise, like ECE and our proposed Gender-ECE. Instance-wise calibration error measuring leads to less reliable results, as it introduces more noise and instability compared to bin-wise measuring, which is more interpretable and provides an understanding of model confidence and calibration behavior.

\begin{table*}[t]
\centering
\renewcommand{\arraystretch}{1}
\setlength{\tabcolsep}{6pt}
\resizebox{\textwidth}{!}{%
\begin{tabular}{lcccccccccccccccc}
\toprule
\textbf{Model} 
& \multicolumn{8}{c}{\textbf{WinoBias}} 
& \multicolumn{8}{c}{\textbf{Winogender}} \\
\cmidrule(lr){2-9} \cmidrule(lr){10-17}
& \multicolumn{4}{c}{\textbf{Standard Metrics}} 
& \multicolumn{3}{c}{\textbf{Gender-ECE}} & \textbf{Human} 
& \multicolumn{4}{c}{\textbf{Standard Metrics}} 
& \multicolumn{3}{c}{\textbf{Gender-ECE}} & \textbf{Human} \\
\cmidrule(lr){2-5} \cmidrule(lr){6-8} \cmidrule(lr){10-13} \cmidrule(lr){14-16}
& \textbf{ECE} 
& \textbf{MacroCE} 
& \textbf{ICE} 
& \textbf{Brier} 
& \textbf{Group} 
& \textbf{M} 
& \textbf{F} 
& 
& \textbf{ECE} 
& \textbf{MacroCE} 
& \textbf{ICE} 
& \textbf{Brier} 
& \textbf{Group} 
& \textbf{M} 
& \textbf{F} 
& \\
\midrule
GPT-J-6B 
& 0.157 & 0.444  & \underline{0.356} & 0.481  
& 0.164 & 0.150 & 0.179 & \textbf{0.686} 
& \underline{0.086} & 0.473  & 0.400  & 0.406  
& 0.118 & 0.066 & 0.170 & 0.685 \\[0.5em]

Llama-3.1-8B
& 0.193  & 0.460  & 0.377  & \underline{0.460} 
& 0.214 & 0.179 & \textbf{0.249} & 0.662 
& 0.099  & 0.475  & 0.387  & 0.428  
& 0.138 & 0.076 & 0.200 & \textbf{0.707} \\[0.5em]

Gemma-2-9B 
& \textbf{0.429} & \textbf{0.490} & \textbf{0.482} & 0.467  
& \textbf{0.297} & \textbf{0.438} & 0.156 & \underline{0.509} 
& \textbf{0.373} & \textbf{0.486} & \textbf{0.422} & \textbf{0.533}  
& \textbf{0.396} & \textbf{0.372} & \textbf{0.421} & \underline{0.573} \\[0.5em]

Qwen2.5-7B
& 0.234  & \underline{0.442} & 0.362  & \textbf{0.510}  
& 0.190 & 0.259 & \underline{0.121} & 0.630 
& 0.136  & \underline{0.461} & \underline{0.379} & 0.463  
& 0.129 & 0.139 & \underline{0.119} & 0.657 \\[0.5em]

Falcon3-7B
& \underline{0.154}  & 0.452 & 0.357  & 0.487  
& \underline{0.149} & 0.160 & 0.138 & 0.684 
& 0.112  & 0.474  & 0.404  & \underline{0.392} 
& 0.176 & 0.079 & 0.273 & 0.696 \\[0.5em]

DeepSeek-8B 
& 0.240  & 0.478  & 0.382  & 0.496  
& 0.218 & 0.255 & 0.182 & 0.648 
& 0.131  & 0.470  & 0.380  & 0.453  
& 0.135 & 0.129 & 0.141 & 0.679 \\

\bottomrule

\end{tabular}
}
\caption{WinoBias and Winogender benchmark datasets calibration evaluation using different metrics across different models. Gemma-2-9B shows the \textbf{worst} calibration overall, and GPT-J-6B is the \underline{more} calibrated model.}

\label{tab:winobias_winogender_split}
\end{table*}

\begin{table}[h]
\small
    \centering
    \begin{tabular}{lcccc}
        \toprule
        \multirow{2}{*}{\textbf{Model}} 
          & \multicolumn{2}{c}{\textbf{WinoBias}} 
          & \multicolumn{2}{c}{\textbf{GenderLex}} \\
        \cmidrule(lr){2-3}\cmidrule(lr){4-5}
        & \textbf{Male} & \textbf{Female} & \textbf{Male} & \textbf{Female} \\
        \midrule
        GPT-J-6B       
          & \colorBinsCell{0.206} 
          & \colorBinsCell{0.508} 
          & \colorBinsCell{0.373} 
          & \colorBinsCell{0.377} \\
        Llama-3.1-8B   
          & \colorBinsCell{0.197} 
          & \colorBinsCell{0.559} 
          & \colorBinsCell{0.396} 
          & \colorBinsCell{0.333} \\
        Gemma-2-9B     
          & \colorBinsCell{0.067} 
          & \colorBinsCell{0.895} 
          & \colorBinsCell{0.056} 
          & \colorBinsCell{0.901} \\
        Qwen2.5-7B    
          & \colorBinsCell{0.130} 
          & \colorBinsCell{0.596} 
          & \colorBinsCell{0.426} 
          & \colorBinsCell{0.416} \\
        Falcon3-7B
          & \colorBinsCell{0.215} 
          & \colorBinsCell{0.502} 
          & \colorBinsCell{0.505} 
          & \colorBinsCell{0.363} \\       
        DeepSeek-8B 
          & \colorBinsCell{0.158} 
          & \colorBinsCell{0.606} 
          & \colorBinsCell{0.303} 
          & \colorBinsCell{0.469} \\
        \bottomrule
    \end{tabular}
    \caption{ECE for male and female pronouns in the WinoBias and GenderLex datasets. Most models exhibit worse calibration for female pronouns, with Gemma-2-9B showing an extreme disparity, having the lowest ECE for male and the highest for female pronouns.}
    \label{tab:bias_ece_male_female}
\end{table}

\begin{table}[t!]
\small
    \centering
  \begin{tabular}{lcccc}
        \toprule
        \multirow{2}{*}{\textbf{Model}} 
        & \multicolumn{4}{c}{\textbf{WinoQueer}} \\
        \cmidrule(lr){2-5}
        & \textbf{Gay} 
        & \textbf{Lesbian} 
        & \textbf{Trans} 
        & \textbf{Queer} \\
        \midrule
        GPT-J-6B       
          & \colorBinsCell{0.121}
          & \colorBinsCell{0.790}
          & \colorBinsCell{0.816}
          & \colorBinsCell{0.700} \\
        Llama-3.1-8B   
          & \colorBinsCell{0.083}
          & \colorBinsCell{0.730}
          & \colorBinsCell{0.760}
          & \colorBinsCell{0.657} \\
        Gemma-2-9B     
          & \colorBinsCell{0.026}
          & \colorBinsCell{0.221}
          & \colorBinsCell{0.586}
          & \colorBinsCell{0.182} \\
          Qwen2.5-7B    
          & \colorBinsCell{0.189}
          & \colorBinsCell{0.898}
          & \colorBinsCell{0.919}
          & \colorBinsCell{0.788} \\
        Falcon3-7B     
          & \colorBinsCell{0.059}
          & \colorBinsCell{0.715}
          & \colorBinsCell{0.686}
          & \colorBinsCell{0.432} \\
        DeepSeek-8B 
          & \colorBinsCell{0.277}
          & \colorBinsCell{0.838}
          & \colorBinsCell{0.258}
          & \colorBinsCell{0.910} \\
        \bottomrule
    \end{tabular}
    \caption{ECE scores on WinoQueer dataset show that Qwen2.5-7B has the highest calibration errors in most categories, while Gemma-2-9B is the most calibrated overall.}
    \label{tab:bias_ece_winoqueer}
\end{table}

\section{Experimental Result}

To answer our research questions and validate our findings, we experiment with a variety of gender bias benchmark datasets:  WinoBias, Winogender, and GenderLex, In each dataset, there are two sentence pairs with a different pronoun. For example, in the WinoBias dataset: "The developer argued with the designer and slapped [him/her] in the face". Each sentence in the dataset is used directly as the input, and we perform a deterministic forward pass (w/o decoding parameters involved) and utilize offset mappings for precise token alignment of the pronoun, ensuring reproducibility.

\noindent{\textbf{Human Alignment.}} Each sentence is assigned a human-labeled bias score, indicating which sentence is more biased: "1" for male bias and "0" for female bias. Human alignment was conducted by three human subjects and one expert annotator, with final decisions made through majority voting. The expert also assessed the quality of the annotations. For example, if a human subject was unfamiliar with a particular occupation (\eg demographer), they would consult the expert annotator to avoid making a random annotation. Notably, the Genderlex dataset showed the most annotation disagreement, likely due to many rare occupation terms, yielding moderate inter-annotator agreement (avg pairwise Cohen’s $\kappa = 0.51$) \cite{cohen1960coefficient}. Overall, this process helps reduce noise from guessing or confusion in the final labels.

\noindent{\textbf{Model.}} We evaluate six state-of-the-art open-weight LLMs: GPT-J-6B \cite{gpt-j}, Falcon3-7B-base \cite{falcon3}, Llama-3.1-8B \cite{llama3modelcard}, Gemma-2-9B  \cite{team2024gemma}, Qwen2.5-7B \cite{qwen2} and  Llama-based DeepSeek-R1-8B or DeepSeek-8B  (Distill from Llama-3.1-8B) \cite{deepseekai2025}.

\noindent{\textbf{RQ1: How well do  LLMs calibrate their confidence in predicting pronouns at the end of the sequence?}}

In this research question, the primary focus of the analysis is to assess the models' bias and confidence in predicting pronouns at the end of sentences, GenderLex dataset (\texttt{last cloze}), meaning that the model has access to the full context of the sentence before scoring a biased pronoun.

\begin{figure*}[ht!]
    \centering
    \begin{minipage}{\textwidth}
        \centering
        \subfloat[ECE: \underline{0.157}]{\includegraphics[width=0.20\textwidth]{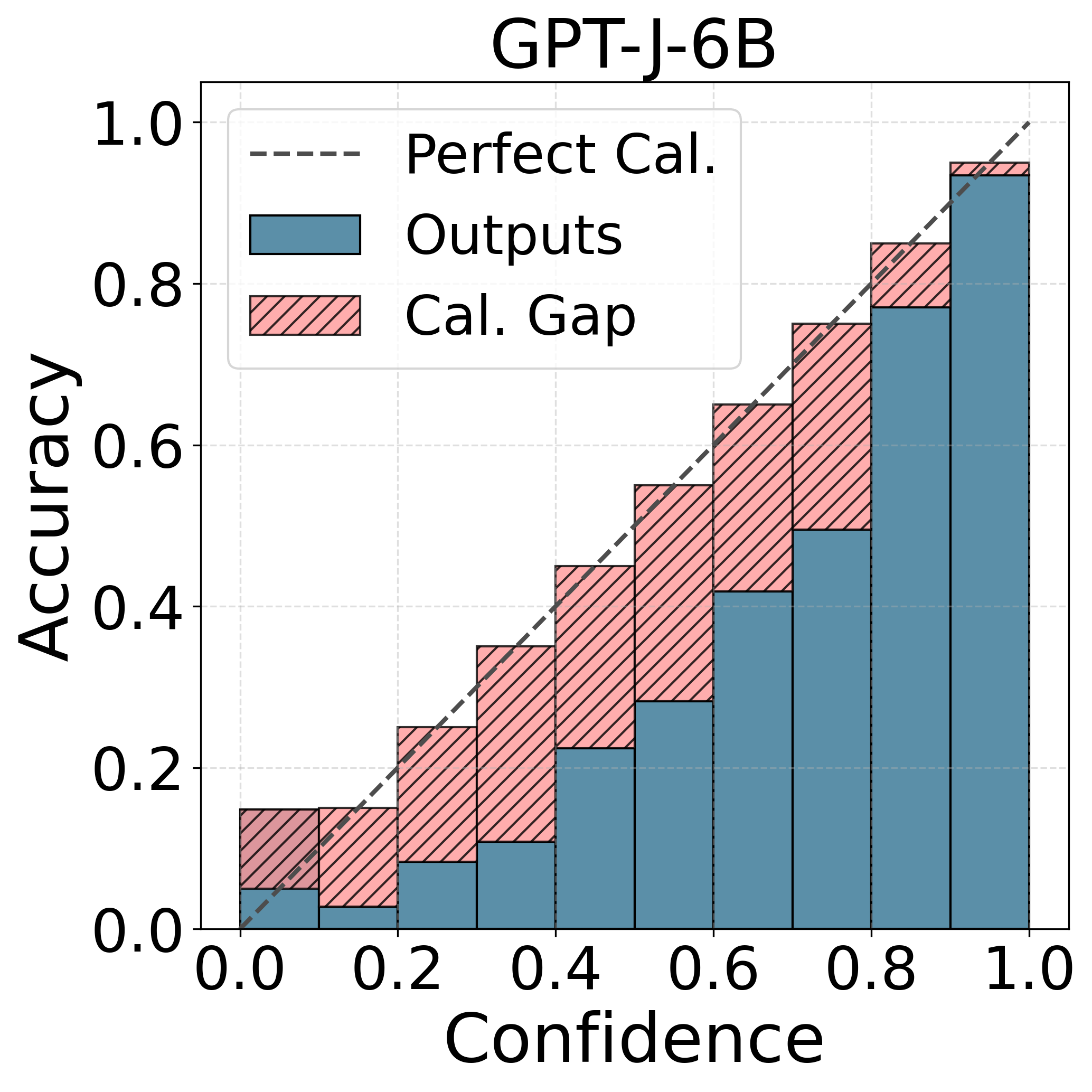}}\hfill
        \subfloat[ECE: 0.193]{\includegraphics[width=0.20\textwidth]{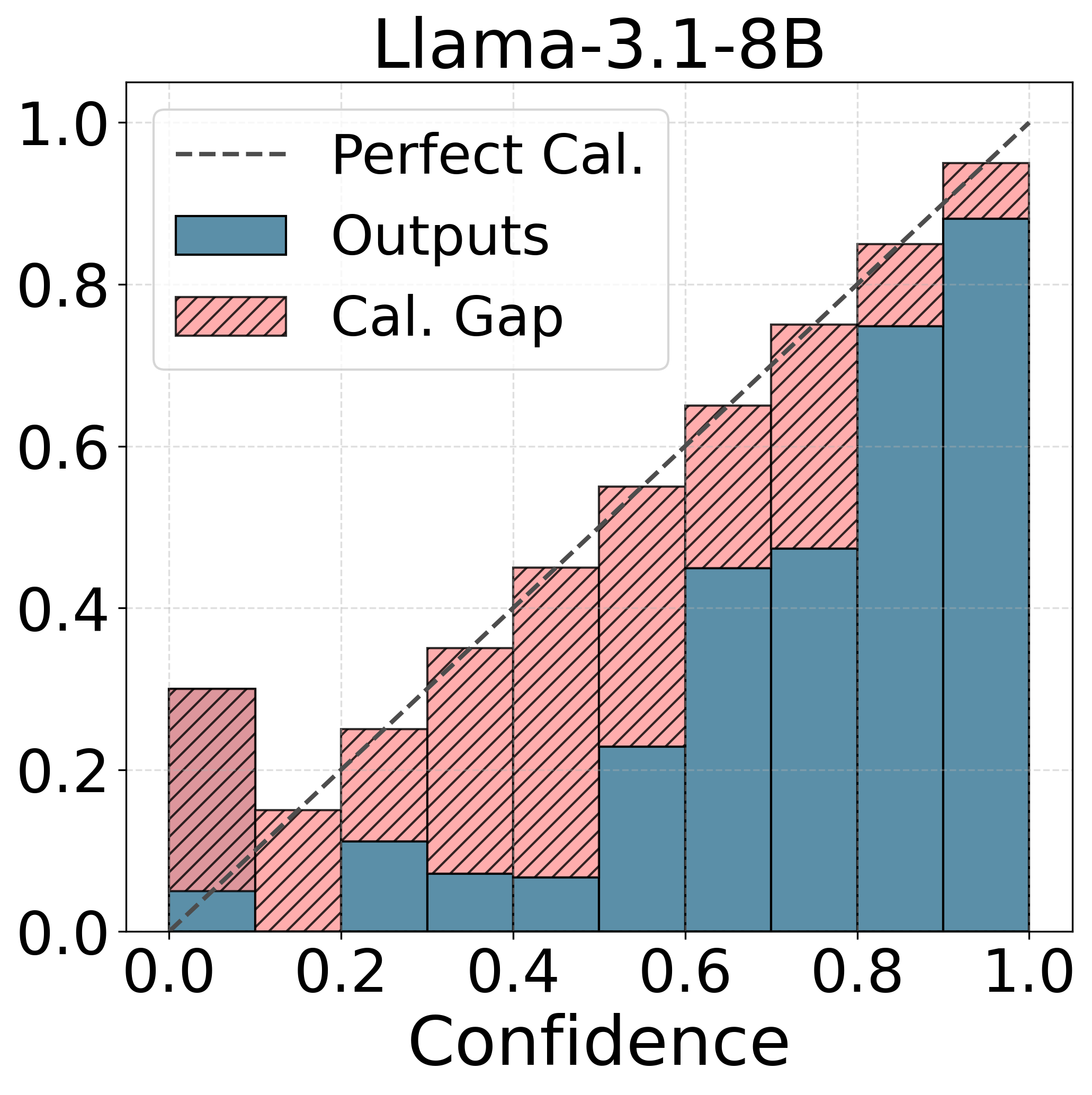}}\hfill
        \subfloat[ECE: 0.240]{\includegraphics[width=0.20\textwidth]{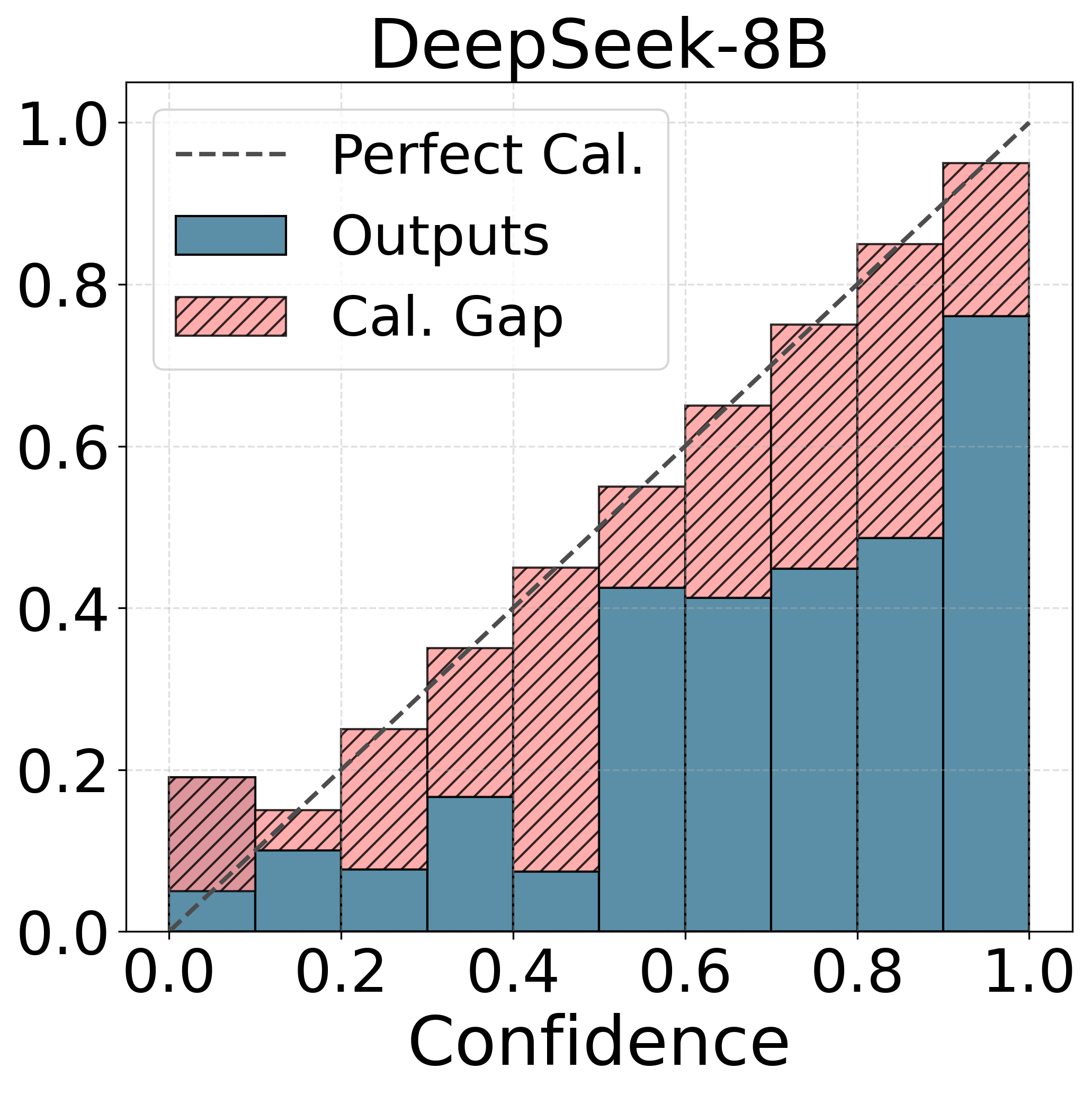}}\hfill
        \subfloat[ECE: \textbf{0.429}]{\includegraphics[width=0.20\textwidth]{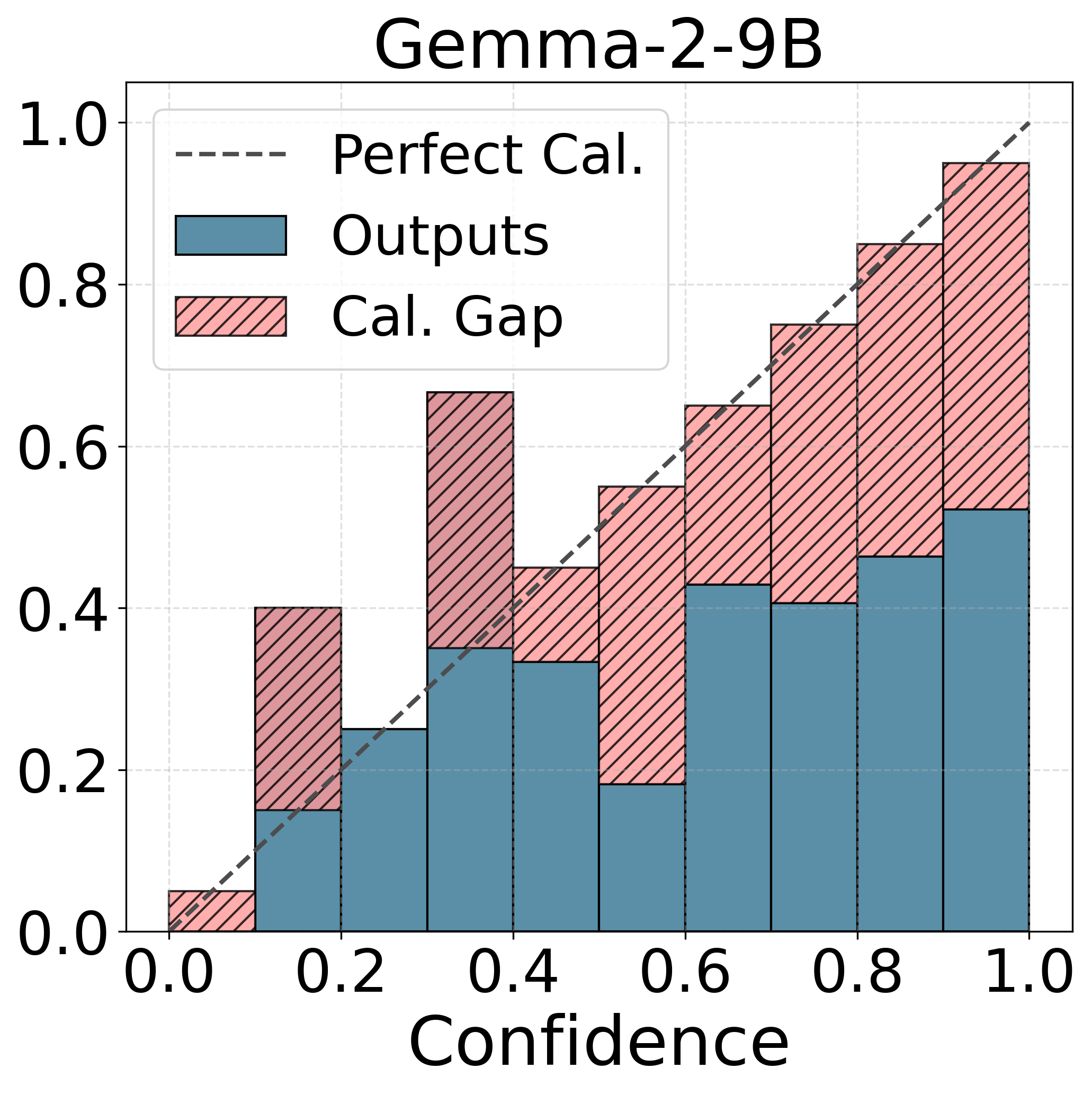}}\hfill
        \subfloat[ECE: 0.234]{\includegraphics[width=0.20\textwidth]{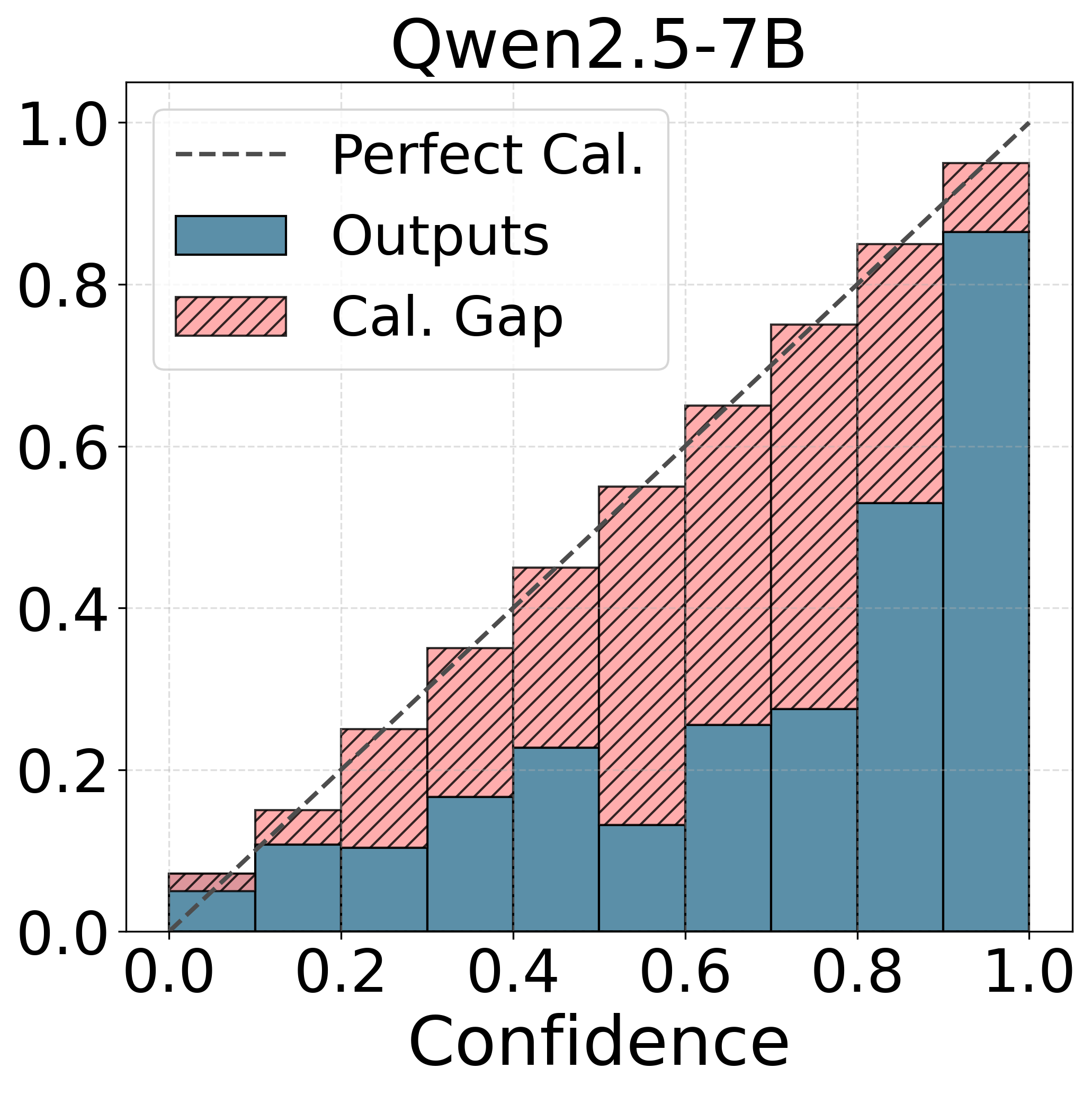}}
    \end{minipage}

    \vspace{-5pt} 

    \begin{minipage}{\textwidth}
        \centering
        \subfloat[ECE: \underline{0.086}]{\includegraphics[width=0.20\textwidth]{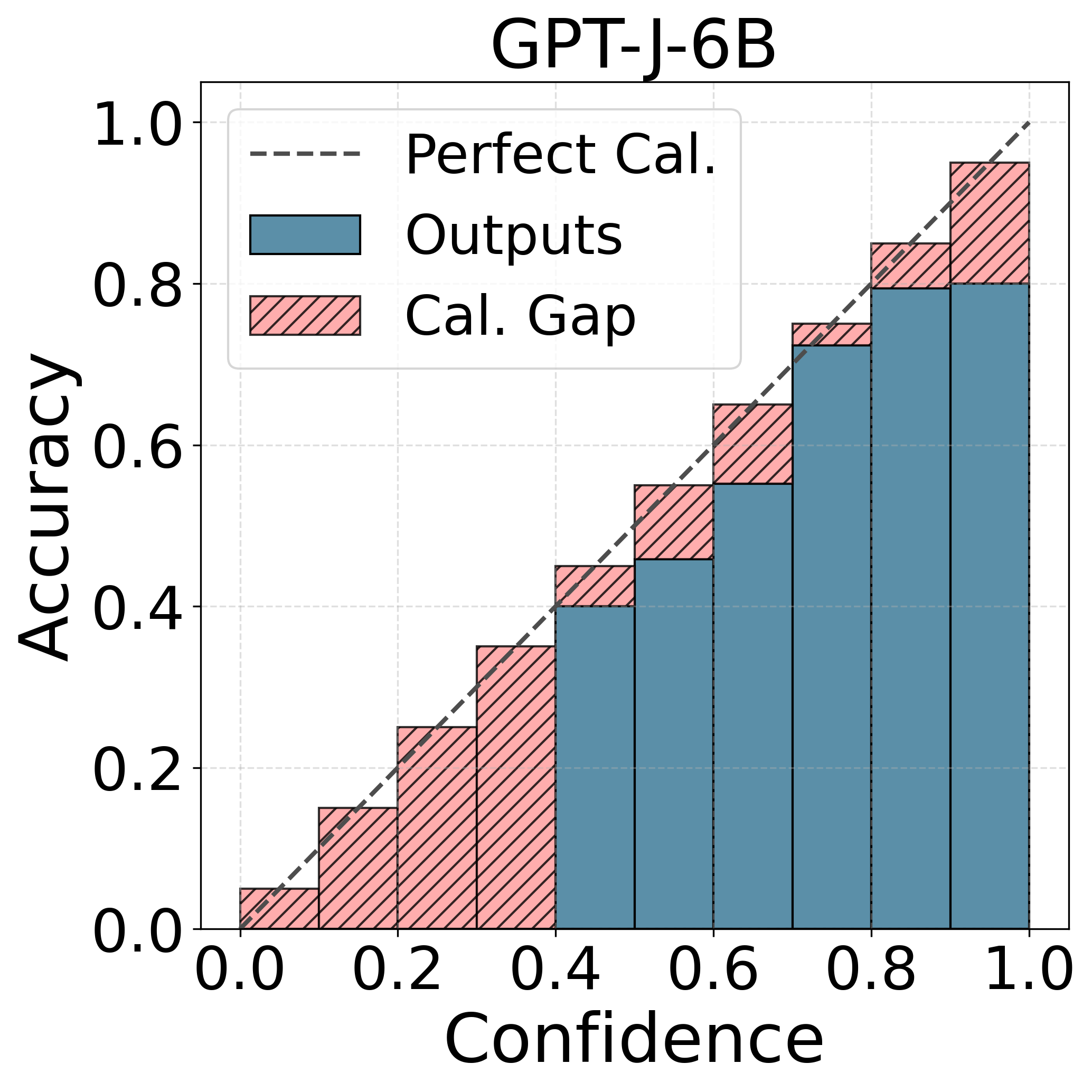}}\hfill
        \subfloat[ECE: 0.099]{\includegraphics[width=0.20\textwidth]{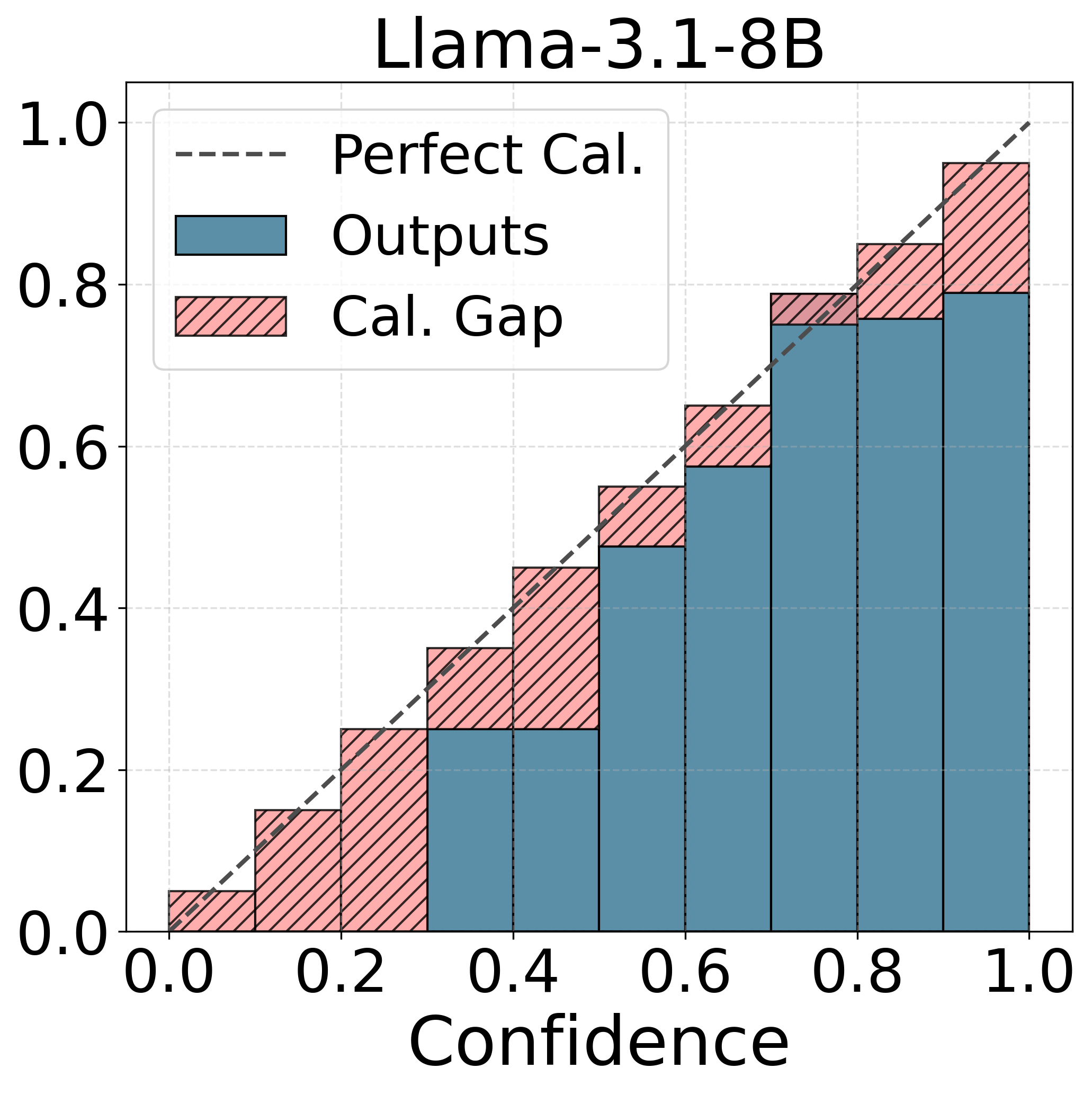}}\hfill
        \subfloat[ECE: 0.131]{\includegraphics[width=0.20\textwidth]{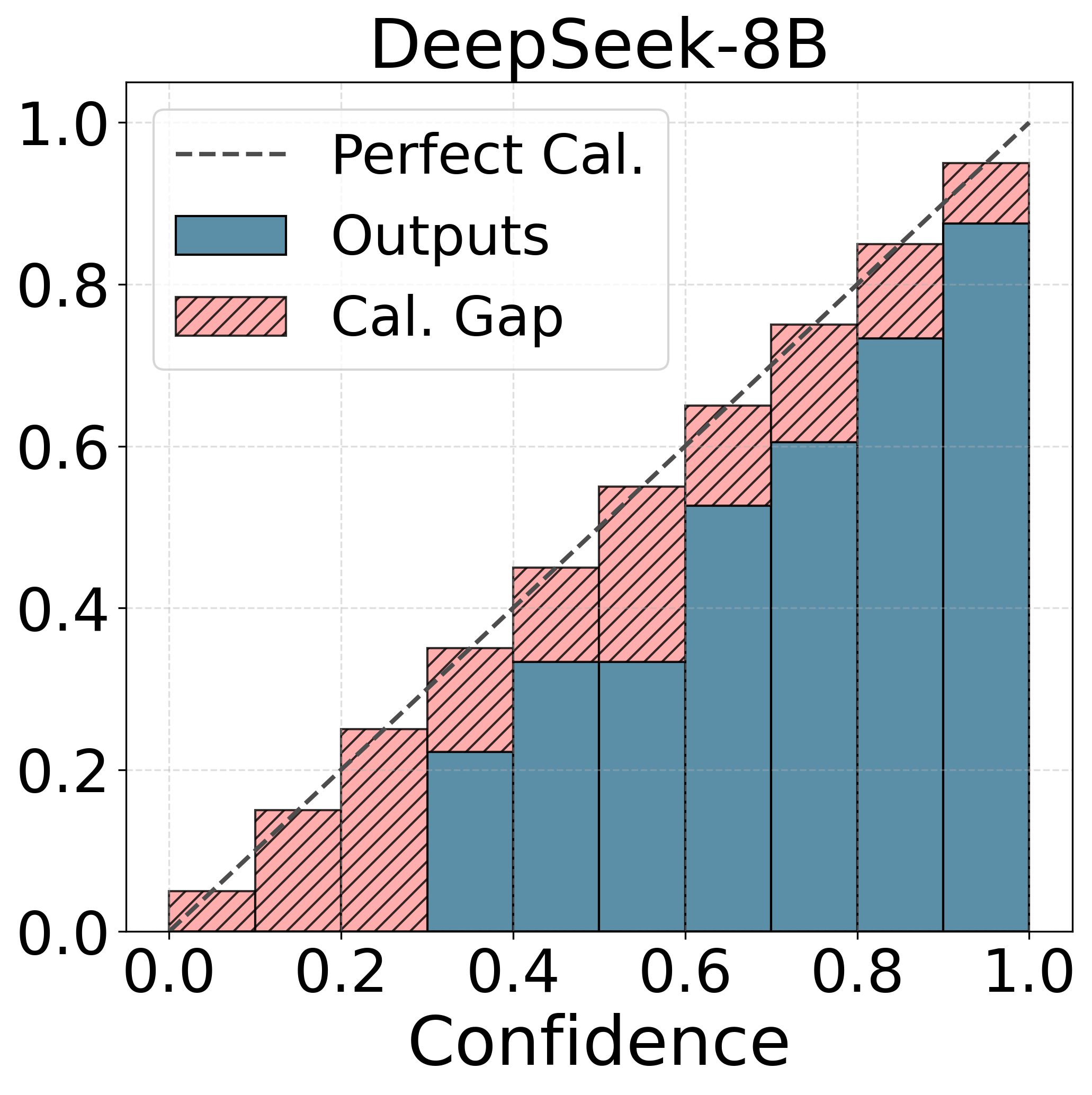}}\hfill
        \subfloat[ECE: \textbf{0.373}]{\includegraphics[width=0.20\textwidth]{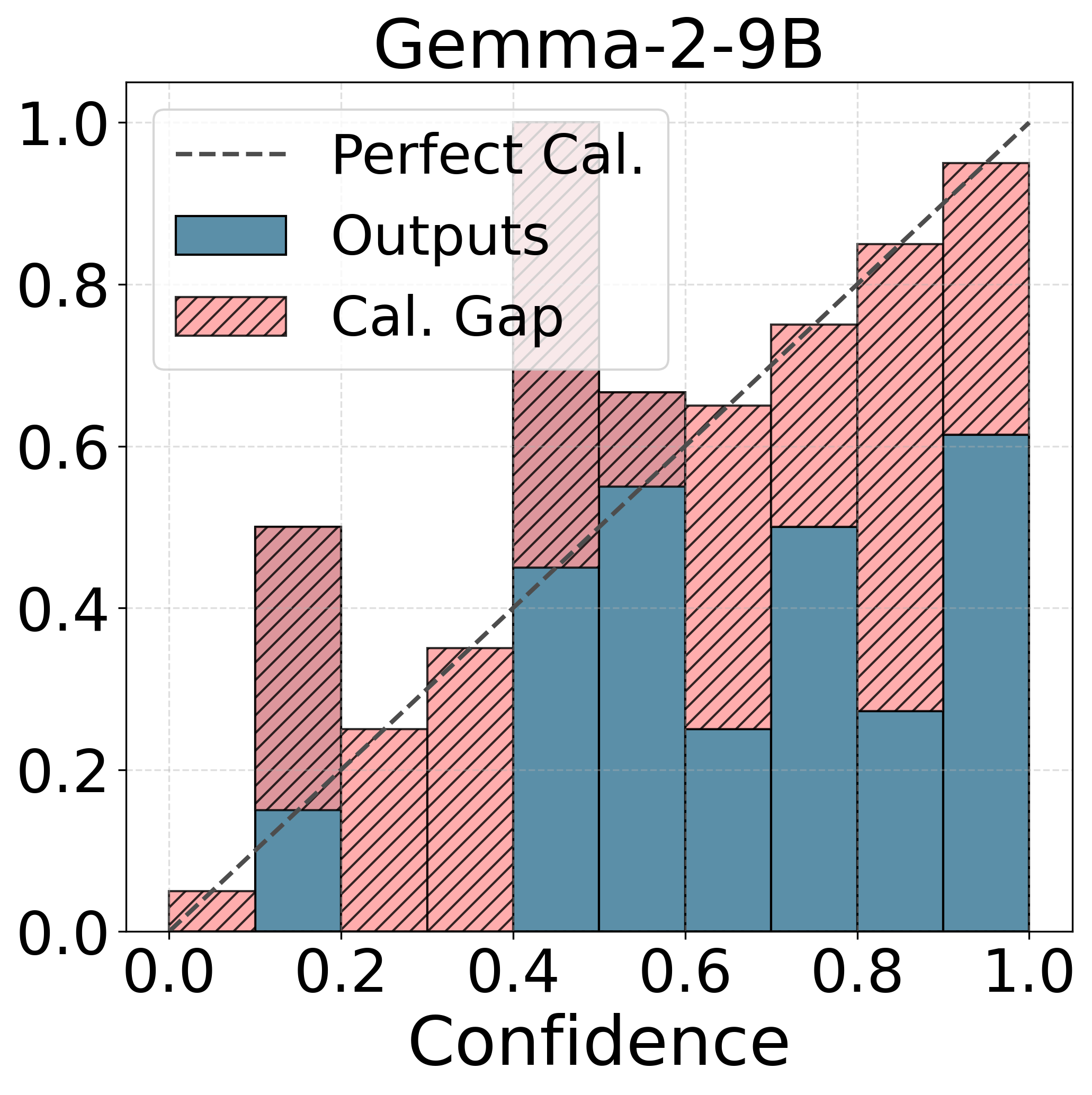}}\hfill
            \subfloat[ECE: 0.136]{\includegraphics[width=0.20\textwidth]{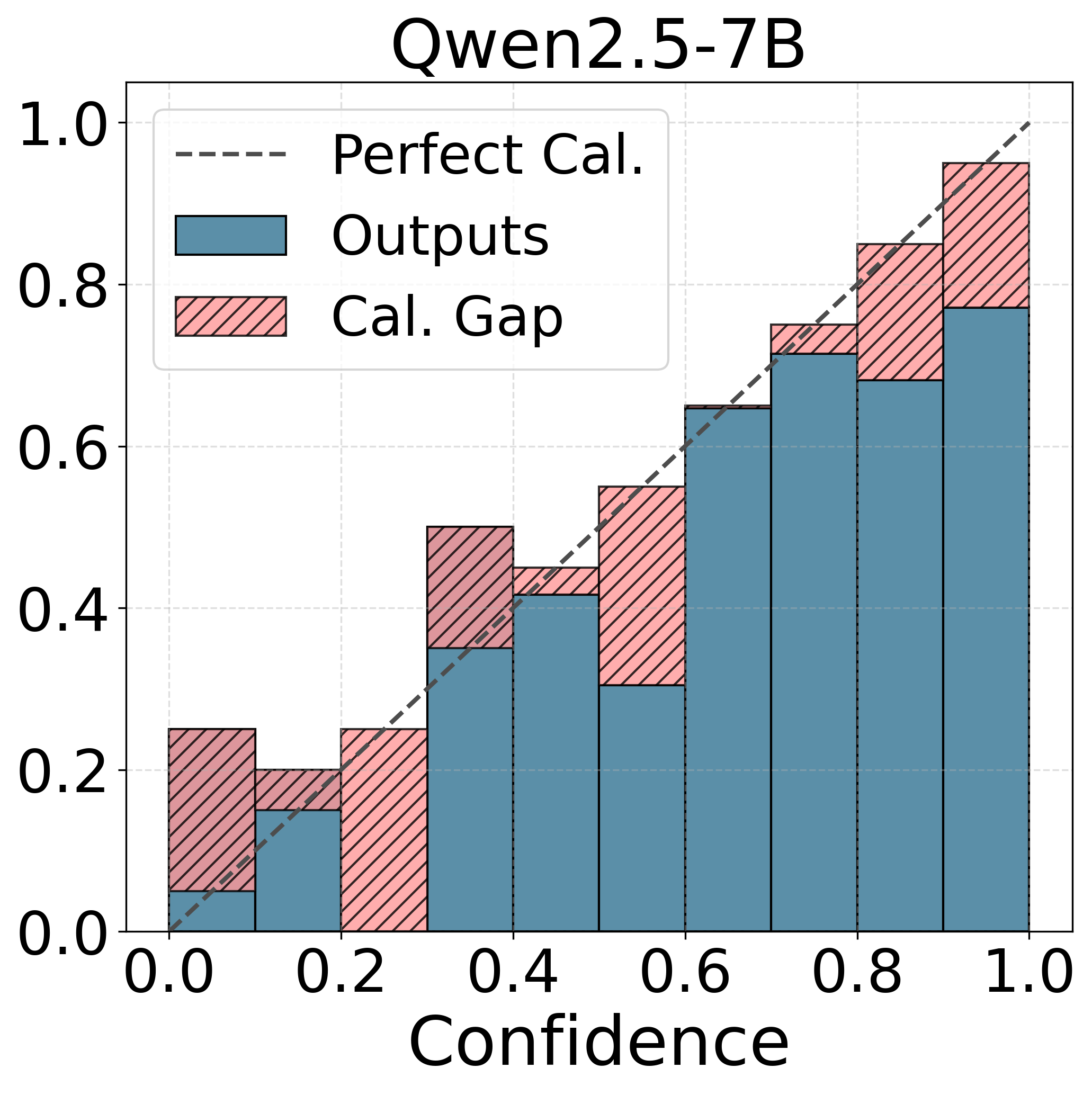}}
    \end{minipage}
    \caption{(Top) Reliability diagrams for WinoBias. (Bottom) Reliability diagrams for Winogender. GPT-J-6B shows the \underline{most} calibrated model performance, while Gemma-2-9B demonstrates the \textbf{worst} human-alignment and calibration.}
    \label{fig:four_side_by_side}
\end{figure*}

Table~\ref{tab:genderlex_gender_split} compares gender bias calibration metrics across language models. Gemma-2-9B exhibits the highest overall miscalibration and a notable disparity between female- and male-associated pronouns. GPT-J-6B demonstrates the strongest calibration and fairness, with balanced performance across genders. Llama-3.1-8B achieves the best alignment with human bias judgments, showing nearly identical calibration across genders. Falcon3-7B provides the most accurate probabilistic predictions (lowest Brier Score) but shows a large female–male gap. Qwen2.5-7B is well-calibrated for male pronouns, while DeepSeek-8B maintains a lower calibration error compared to Llama-3.1-8B.

\noindent{\inlinefinding{Finding~1}} As shown in reliability diagrams (Figure \ref{fig:GenderVocab-reliability diagrams}), GPT-J-6B shows the best calibration (lowest ECE), with relatively small Gender-ECE gap differences. Gemma-2-9B performs worst overall, being consistently incorrect and high risk for deployment. Llama-3.1-8B is the most consistently human-aligned, while Falcon3-7B, though under-calibrated, achieves the most accurate probabilities (lowest Brier) but exhibits a wider gender gap.

\noindent{\textbf{RQ2: How well are LLMs calibrated when resolving pronouns in gender-biased coreference tasks?}} 

In this experiment, we rely on two benchmark datasets designed to evaluate gender bias in coreference resolution.

\noindent{\textbf{WinoBias Dataset.}}  This dataset is designed to evaluate gender bias in language models, particularly in coreference resolution tasks in which the model predicts pronouns in the middle of sentences rather than at the end. We employ the WinoBias-syntax (type-2) dataset, which consists of sentence pairs where answers can be resolved using syntactic information alone (without grounding or world knowledge), comprising 1,542 sentence pairs (771 unique sentences, each containing a single pronoun). We replace occupational titles with a gender-neutral term \texttt{person} (\eg “The \st{developer} person argued with the [designer] and slapped [her] in the face”) to reduce ambiguity in pronoun prediction, when two occupational entities co-occur in a sentence.

Table \ref{tab:winobias_winogender_split} (Left) shows that Gemma-2-9B has the worst calibration. In contrast, Falcon3-7B and GPT-J-6B achieve the best calibration among the models evaluated. For gender calibration, Gemma-2-9B shows the largest gap, with much worse calibration for male pronouns. Llama-3.1-8B exhibits the highest miscalibration for female pronouns.
Falcon3-7B and GPT-J-6B again show the lowest Gender-ECE scores, reflecting a better balance across genders. Human alignment scores diverge notably for Gemma-2-9B, indicating weaker alignment with human bias and a higher risk of biased behavior toward the male group.

\noindent{\inlinefinding{Finding 2} Figure \ref{fig:four_side_by_side}  (\texttt{a} to \texttt{e}) show reliability diagrams for WinoBias dataset. Gemma-2-9B exhibits the worst calibration overall and a pronounced bias toward male pronouns, as indicated by its high Gender-ECE (male) score. These patterns indicate that Gemma-2-9B is both overconfident and inconsistent in gender-sensitive contexts. GPT-J-6B and Falcon3-7B are the fairest models, with the lowest Gender-ECE and minimal differences between pronouns.

\noindent{\textbf{Winogender Dataset.}} Unlike WinoBias dataset, which tests stereotypical occupational roles, Winogender evaluates gender preference in pronoun resolution without an inherent gender stereotype.
\eg "The paramedic helped the passenger because [he/she] was injured."

Table~\ref{tab:winobias_winogender_split} (Right) shows that Gemma-2-9B has the weakest calibration, with the highest Gender-ECE and the lowest human alignment. GPT-J-6B achieves the lowest overall ECE but shows higher calibration error for female pronouns. Llama-3.1-8B shows moderate calibration with a higher  Gender-ECE female error, yet the highest agreement with human judgments. Qwen2.5-7B exhibits balanced calibration with moderate gender gaps, while Falcon3-7B shows moderate ECE  but a pronounced female–male gap.

\noindent{\inlinefinding{Finding~3}} 
Figure~\ref{fig:four_side_by_side} (\texttt{f}--\texttt{j}) shows reliability diagrams for the Winogender dataset.  Gemma-2-9B exhibits the worst calibration, with the highest ECE and Gender-ECE, indicating poor handling of both male and female pronouns. 
GPT-J-6B achieves the lowest overall ECE and Gender-ECE but shows substantially higher error for female pronouns, indicating a notable gender imbalance in calibration.

\noindent{\textbf{RQ3: How well-calibrated are LLMs in predicting male and female gender-specific pronouns?}}

In this experiment, we examine model confidence in pronoun prediction for gender-biased occupations. We use human-labeled data to split the WinoBias and GenderLex datasets into male and female subsets.

Table \ref{tab:bias_ece_male_female} shows that most models have higher ECE (\ie worse calibration) for female pronouns, suggesting that LLMs are generally less reliable when handling female pronouns. In particular, Gemma-2-9B achieves low ECE for male pronouns 7\% in the WinoBias dataset, but extremely high ECE for female pronouns  93\%, indicating a significant calibration gap between male and female references. A similar pattern is observed in GenderLex, where the male ECE is 6\% and the female ECE is 94\%. Moreover, GenderLex dataset exhibits higher ECE scores for male pronouns compared to the WinoBias dataset, demonstrating substantial miscalibration when pronouns are cloze last.

\noindent{\inlinefinding{Finding~4}} Most LLMs are less calibrated on female pronouns, revealing bias toward male references.  Gemma-2-9B is well-calibrated for male pronouns but poorly calibrated for female, exhibiting a clear gender-specific bias.

\noindent{\textbf{RQ4: How well are LLMs calibrated on predictions about marginalized groups (\ie LGBTQ community)?}}

In this experiment, we use the WinoQueer dataset \cite{felkner2023winoqueer}. A Winograd-inspired benchmark designed to measure anti-LGBTQ+ biases in LLMs. From the 45,540 available sentences, we extract only 1,488 sentence pairs that, similar to GenderLex, end with a \texttt{last cloze} (\eg "everyone hates Wren because they are [LGBTQ]"). 

Table~\ref{tab:bias_ece_winoqueer} presents ECE scores for the WinoQueer dataset. Overall, the majority of models exhibit significantly greater miscalibration for Trans and Lesbian identity mentions, indicating increased bias and lower reliability when handling these identity groups. Interestingly, Qwen2.5-7B exhibits the highest ECE overall, particularly for Trans and Lesbian identities, revealing significant calibration issues. DeepSeek-8B also shows poor calibration for Queer identities. In contrast, Gemma-2-9B stands out as the most well-calibrated model in this setting, achieving the lowest ECE for Gay and Queer identities and relatively low scores overall. This suggests that Gemma-2-9B is better at maintaining confidence consistency across queer identity mentions. Meanwhile, GPT-J-6B and Falcon3-7B also perform poorly, especially on Lesbian and Trans instances, further highlighting the challenges these models face in equitably handling LGBTQ+ representations.

\noindent{\inlinefinding{Finding~5}} All Models exhibit the highest miscalibration for Lesbian and Trans identities in the WinoQueer dataset, with Qwen2.5-7B  showing the worst overall.

\begin{table}[t!]
\small
    \centering
     \begin{tabular}{lcccc}
        \toprule
        \textbf{Model} 
        & \textbf{Metric} 
        & \textbf{Someone} 
        & \textbf{Person} 
        & \textbf{Occ} \\
        \midrule
        \multirow{2}{*}{GPT-J-6B}       
          & ECE        & \colorBinsCell{0.144} & 0.063 & 0.076 \\
          & GECE       & \colorBinsCell{0.138} & 0.077 & 0.076 \\
        \midrule
        \multirow{2}{*}{Llama-3.1-8B}   
          & ECE        & 0.134 & \colorBinsCell{0.138} & 0.111 \\
          & GECE       & \colorBinsCell{0.132} & 0.130 & 0.111 \\
        \midrule
        \multirow{2}{*}{Gemma-2-9B}     
          & ECE        & 0.364 & \colorBinsCell{0.367} & 0.327 \\
          & GECE       & \colorBinsCell{0.450} & 0.351 & 0.267 \\
          \midrule
        \multirow{2}{*}{DeepSeek-8B}    
          & ECE        & \colorBinsCell{0.139} & 0.130 & 0.085 \\
          & GECE       & \colorBinsCell{0.138} & 0.137 & 0.090 \\
        \bottomrule
    \end{tabular}
    \caption{Gender-ECE (Group) and ECE scores on the GenderLex dataset show that models exhibit high calibration error with gender-neutral terms: \texttt{Someone} and \texttt{Person}.} 
    \label{tab:group_ece_scores}
\end{table}

\begin{figure*}[t!]
    \centering
    \begin{minipage}{\textwidth}
        \centering
        \subfloat[ECE: 0.148]{\includegraphics[width=0.20\textwidth]{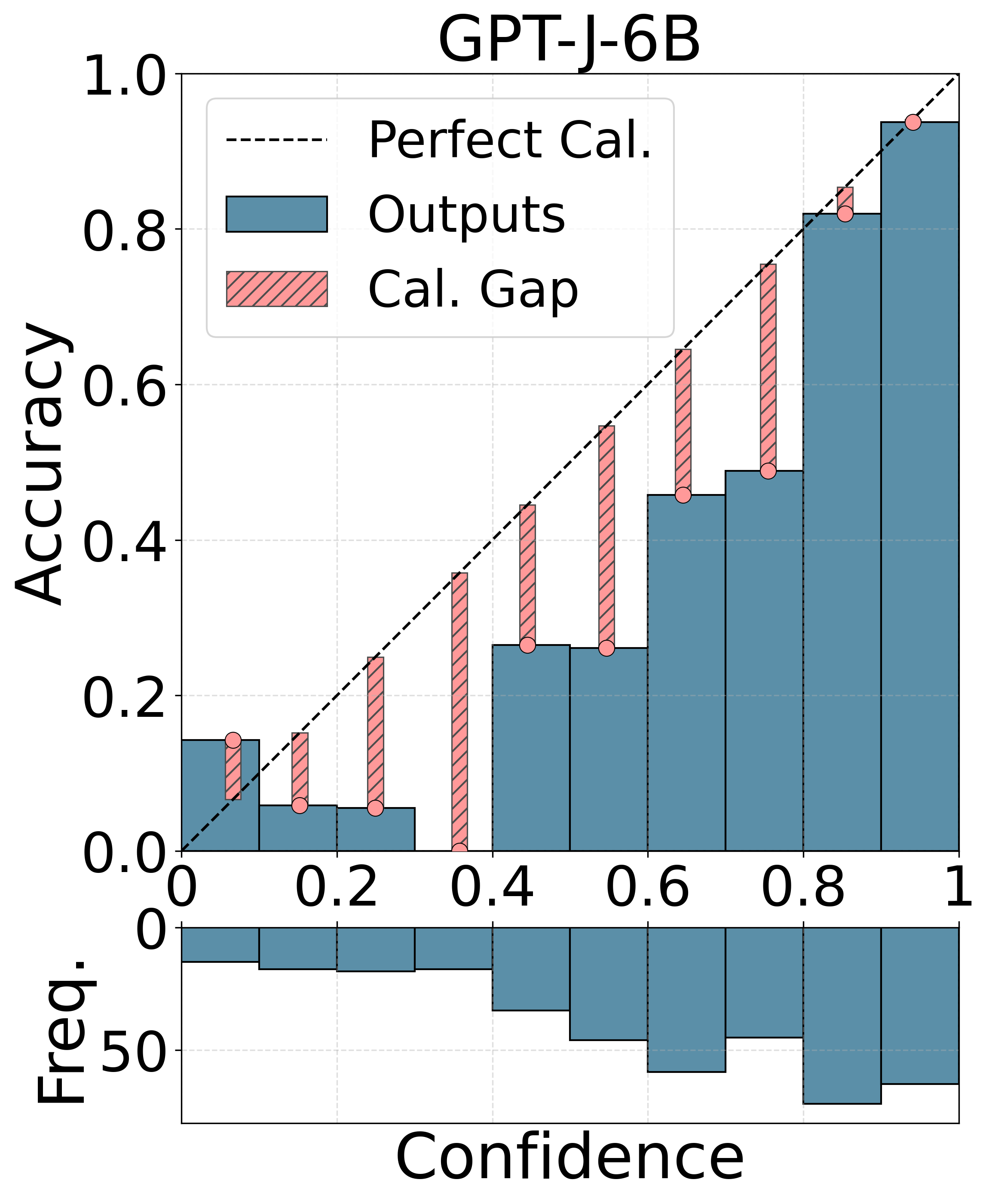}}\hfill
        \subfloat[ECE: 0.196]{\includegraphics[width=0.20\textwidth]{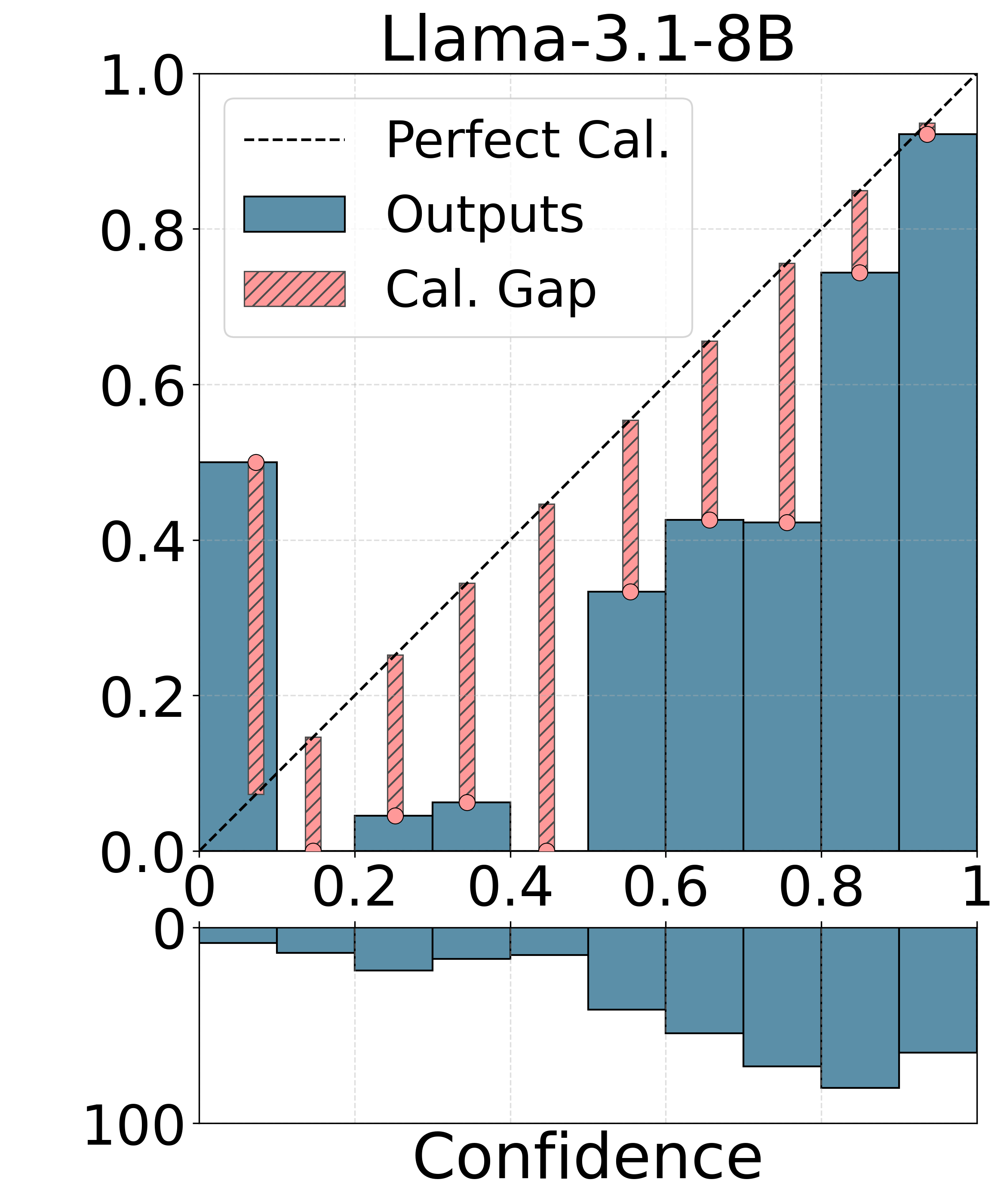}}\hfill
        \subfloat[ECE: 0.235]{\includegraphics[width=0.20\textwidth]{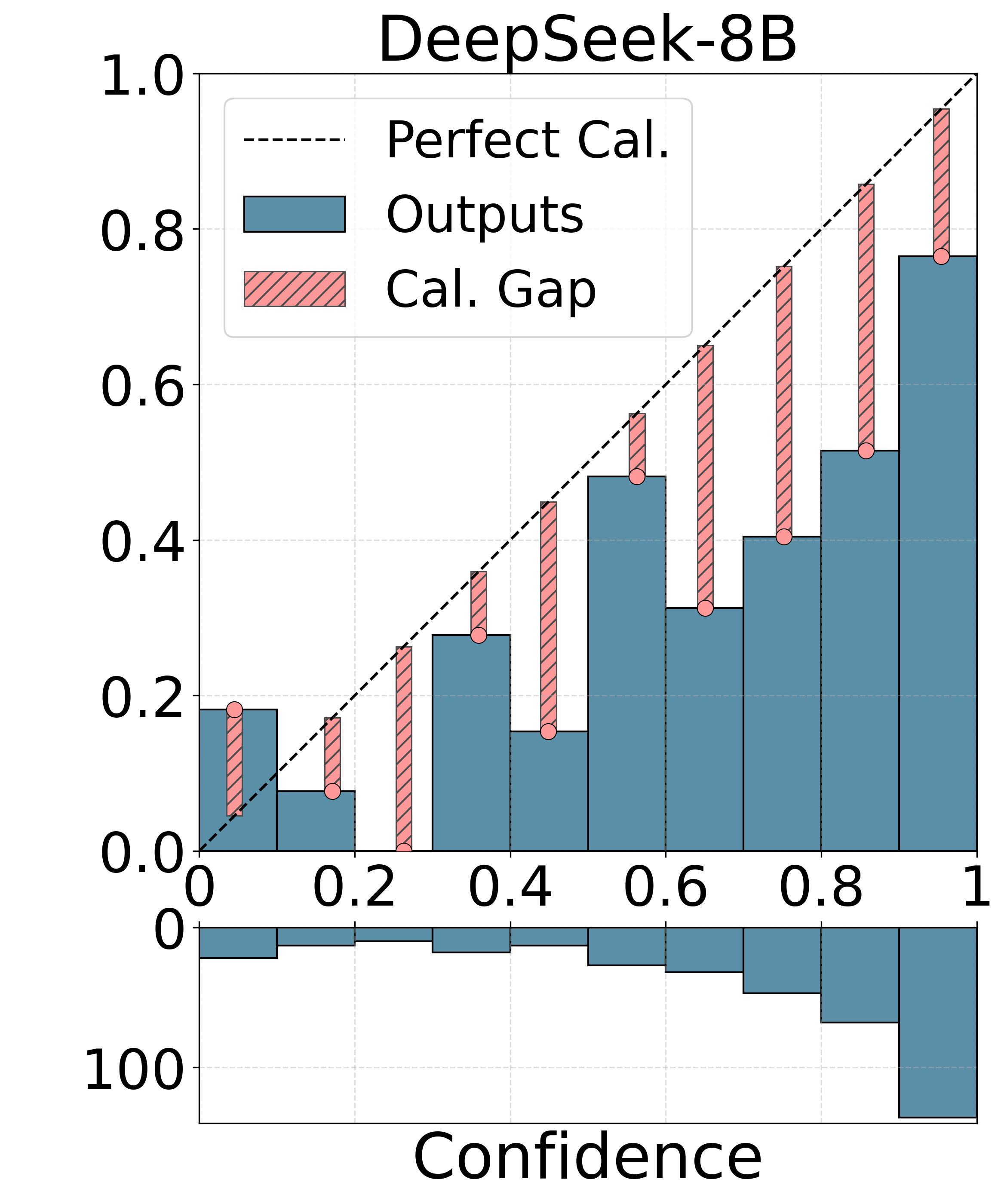}}\hfill
        \subfloat[ECE: 0.429]{\includegraphics[width=0.20\textwidth]{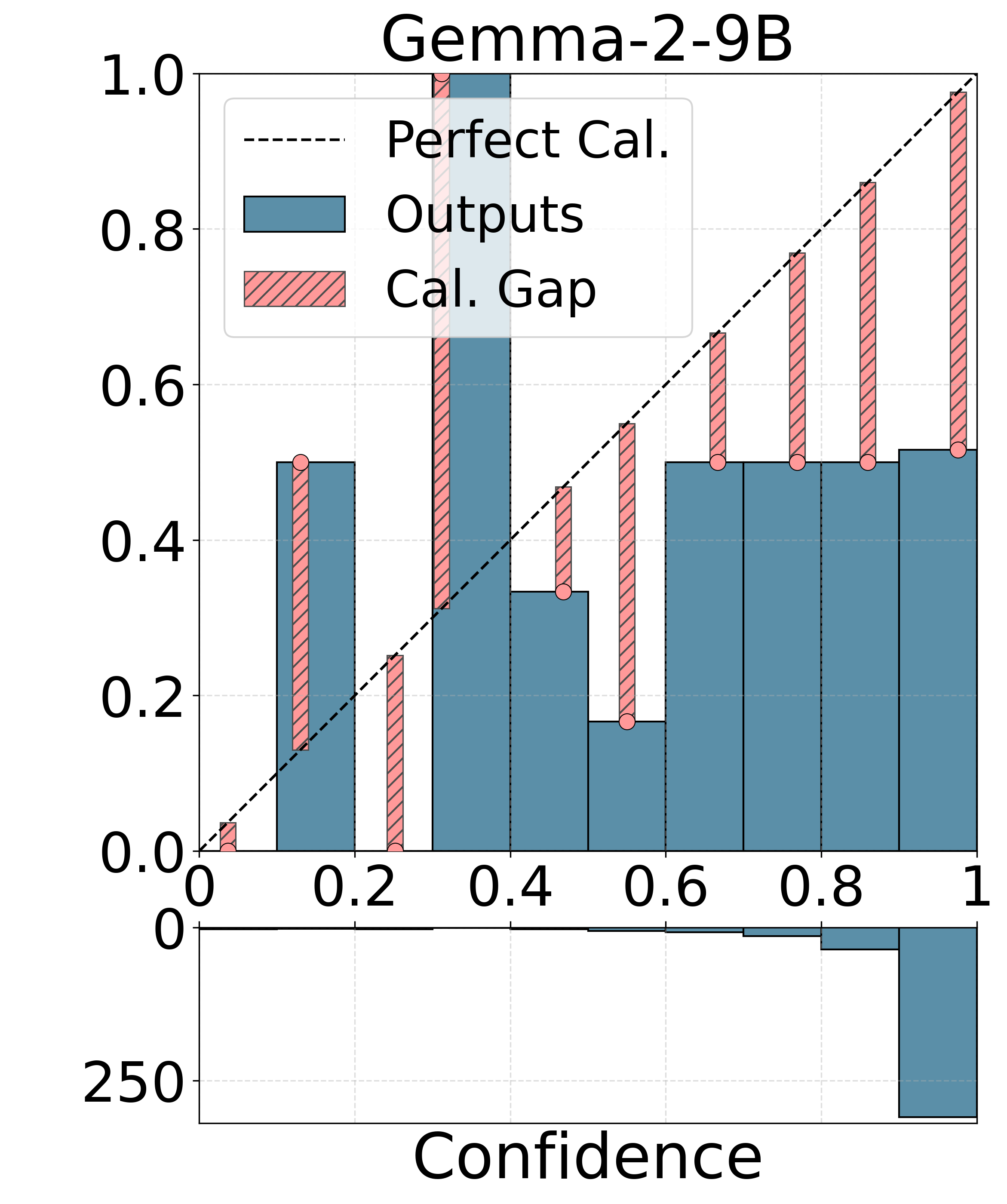}}\hfill
        \subfloat[ECE: 0.234]{\includegraphics[width=0.20\textwidth]{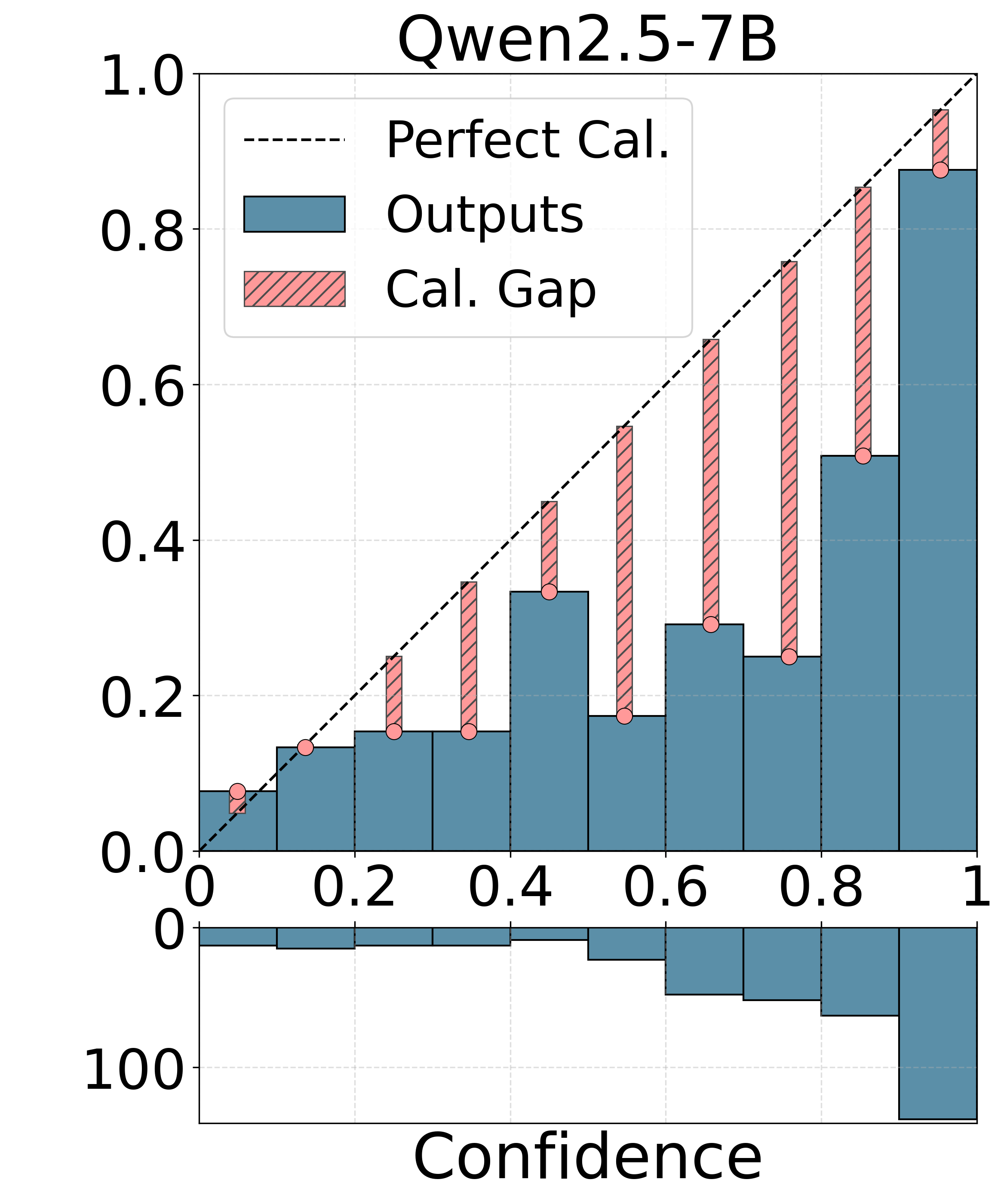}}
    \end{minipage}

    \vspace{-5pt} 

    \begin{minipage}{\textwidth}
        \centering
        \subfloat[ECE: 0.045]{\includegraphics[width=0.20\textwidth]{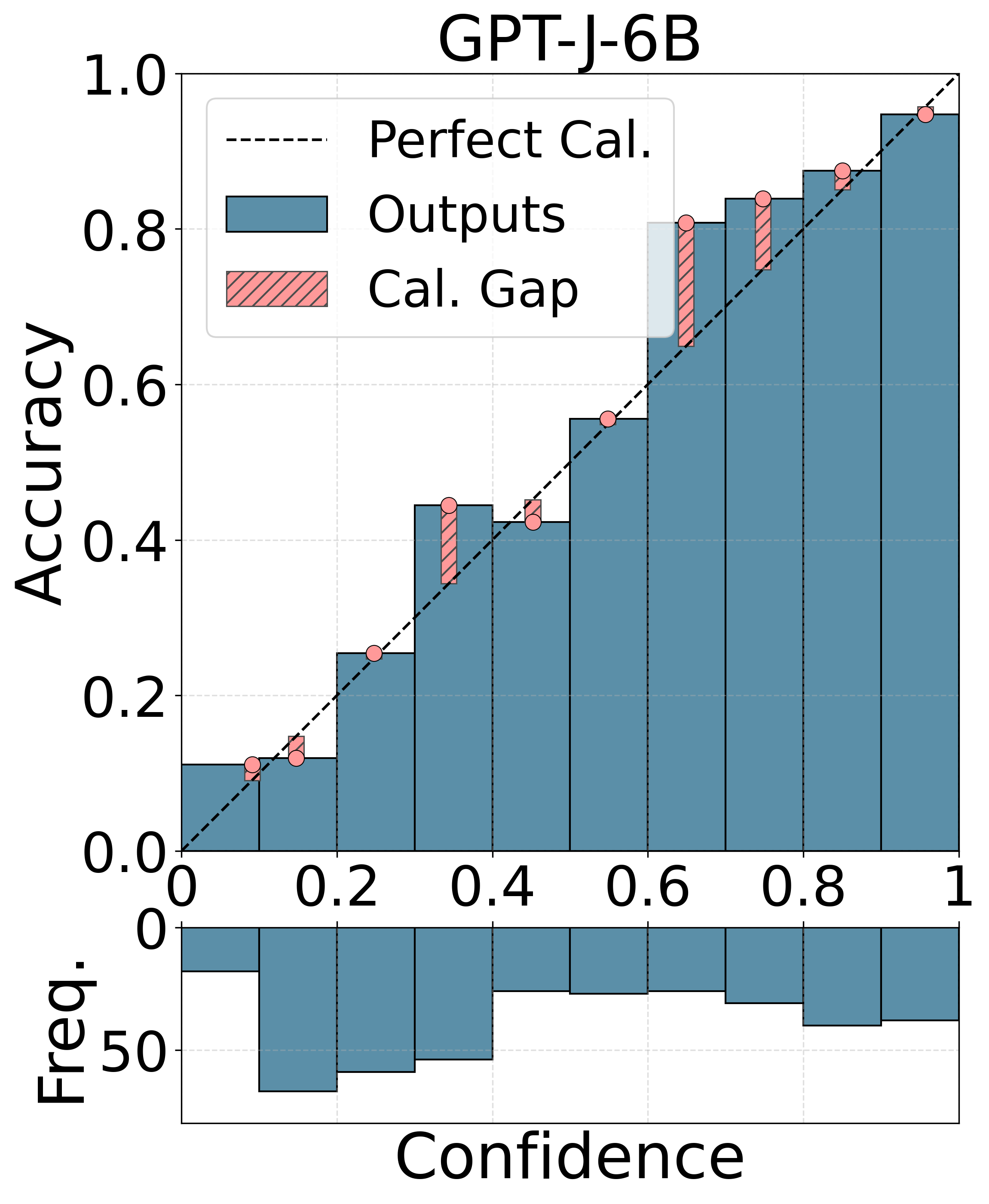}}\hfill
        \subfloat[ECE: 0.067]{\includegraphics[width=0.20\textwidth]{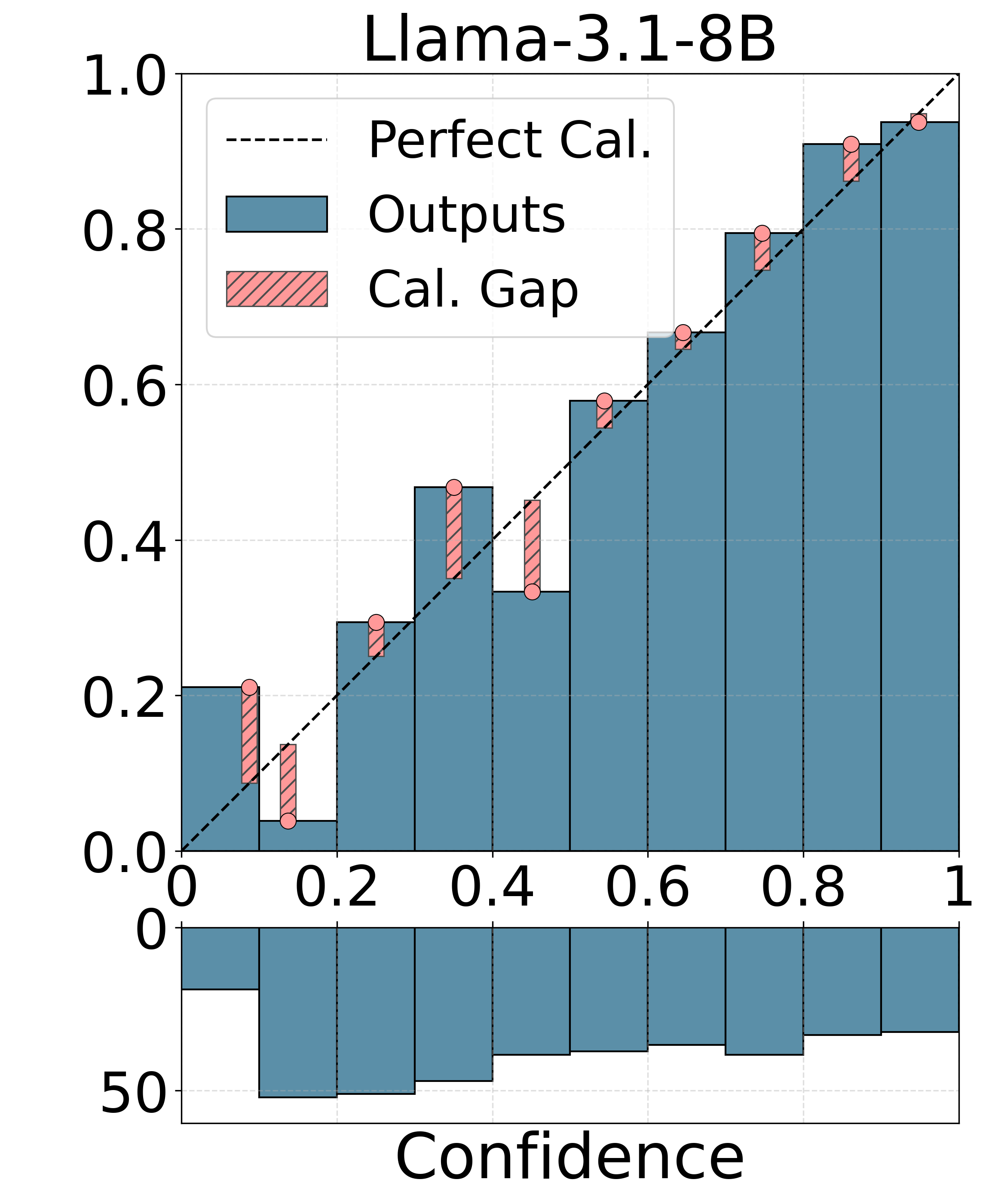}}\hfill
        \subfloat[ECE: 0.062]{\includegraphics[width=0.20\textwidth]{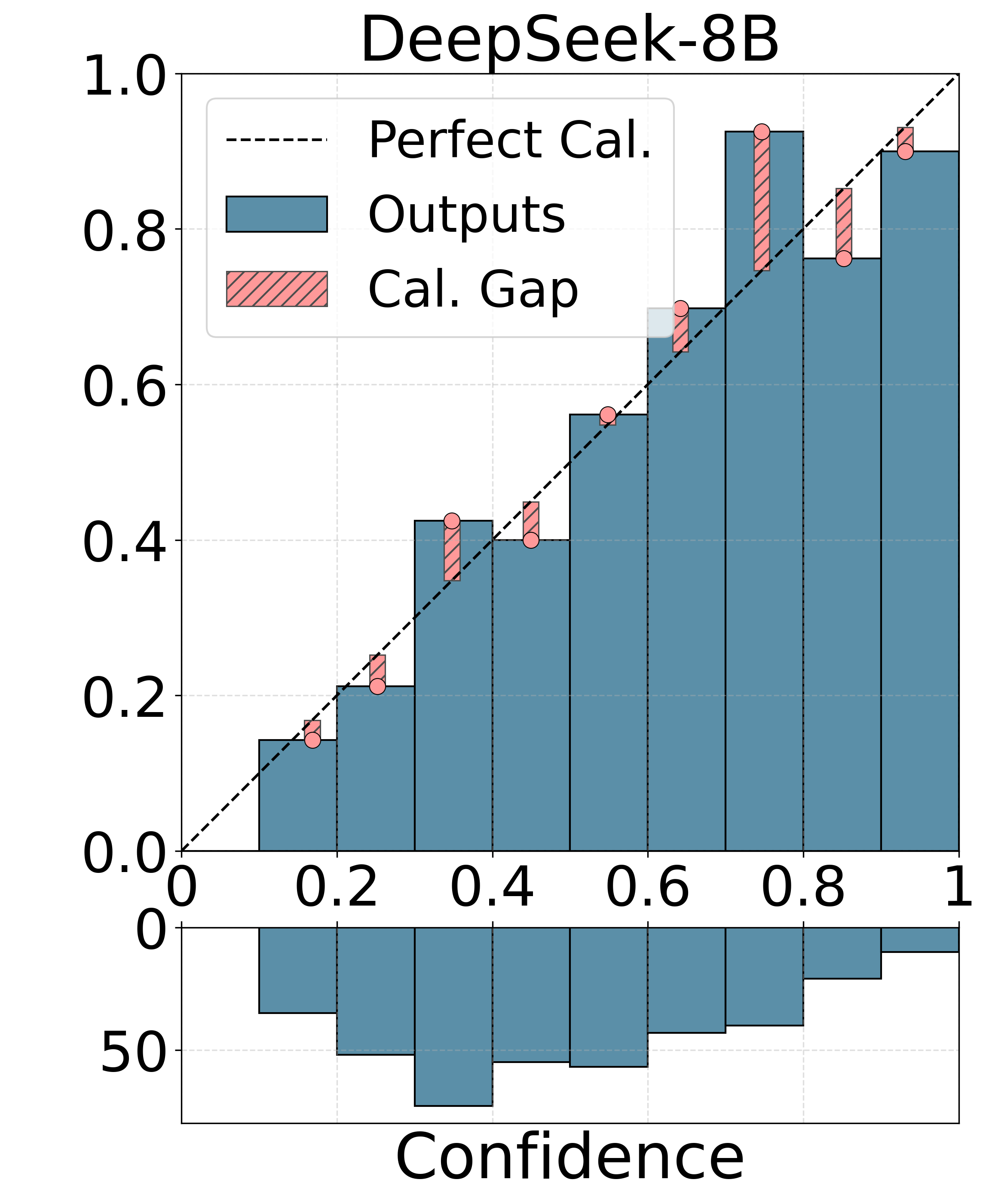}}\hfill
        \subfloat[ECE: 0.025]{\includegraphics[width=0.20\textwidth]{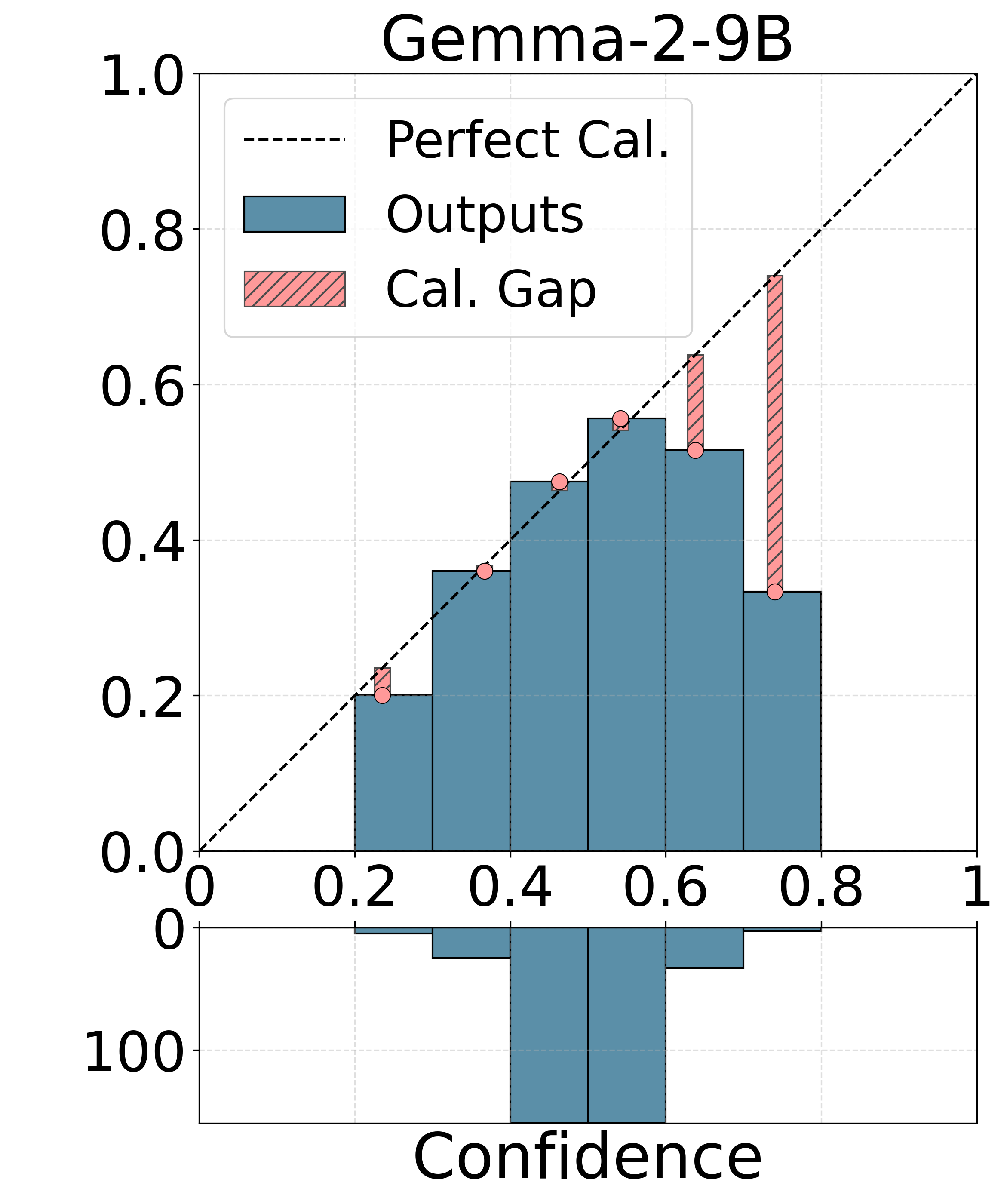}}\hfill
         \subfloat[ECE: 0.076]{\includegraphics[width=0.20\textwidth]{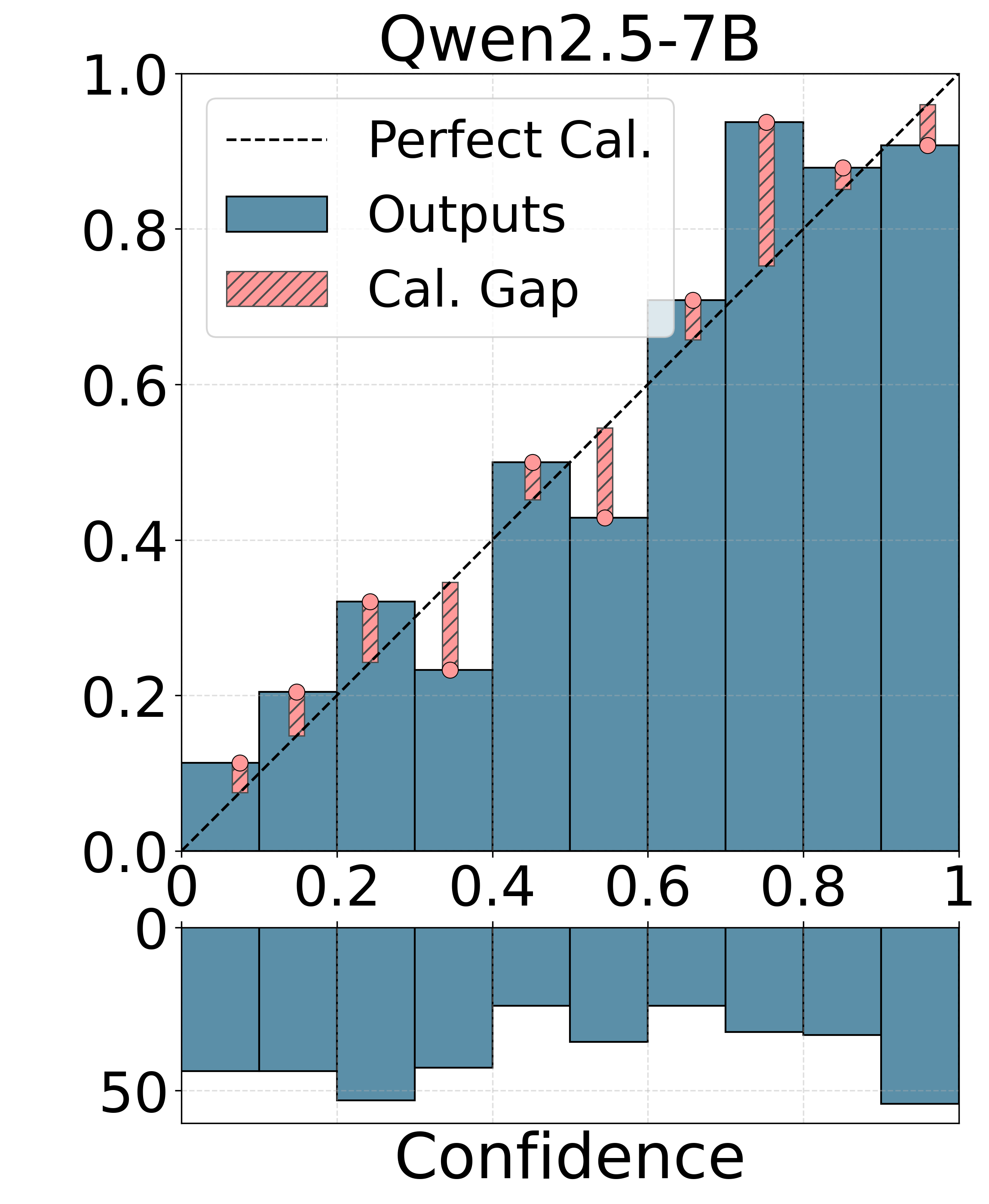}}
    \end{minipage}

      \caption{Comparison of uncalibrated (Top) and Beta calibrated (Bottom) gender bias probabilities for WinoBias dataset. Instances are divided into 10 equal-width bins. The red bar shows the distance to perfectly calibrated probabilities in the bin, and the blue bar shows the accuracy.  The histogram below the reliability diagram shows how many instances are in each bin.}

    \label{fig:calibration_winobias}
\end{figure*}

\begin{table}[t]
\centering
\resizebox{\columnwidth}{!}{%
\begin{tabular}{lcccc}
\toprule
\textbf{Model} 
& \textbf{ECE} 
& \multicolumn{3}{c}{\textbf{Gender-ECE}} \\
\cmidrule(lr){3-5}
             &      & \textbf{Group} & \textbf{M} & \textbf{F} \\
\midrule
Llama-3.1-8B   & $0.193_{\textcolor{white}{+00.0\% \uparrow}}$ 
               & $0.214_{\textcolor{white}{+0.0\% \uparrow}}$  
               & $0.179_{\textcolor{white}{+00.0\% \uparrow}}$ 
               & $0.249_{\textcolor{white}{+0.0\% \uparrow}}$ \\
Llama-3.1-70B  & $\colorBinsCell{0.216}_{\textcolor{red}{+11.9\% \uparrow}}$   
               & $\colorBinsCell{0.230}_{\textcolor{red}{+7.5\% \uparrow}}$    
               & $\colorBinsCell{0.207}_{\textcolor{red}{+15.6\% \uparrow}}$   
               & $\colorBinsCell{0.253}_{\textcolor{red}{+1.6\% \uparrow}}$ \\

\addlinespace
\hline  
\addlinespace
Gemma-2-9B     & $\color{black}{0.429}_{\textcolor{white}{+00.0\% \uparrow}}$ 
               & $0.297_{\textcolor{white}{+00.0\% \uparrow}}$ 
               & $0.438_{\textcolor{white}{+00.0\% \uparrow}}$ 
               & $0.156_{\textcolor{white}{+000.0\% \uparrow}}$ \\

Gemma-2-27B    & ${0.366}_{\textcolor{blue}{-14.7\% \downarrow}}$   
               & $\colorBinsCell{0.361}_{\textcolor{red}{+21.5\% \uparrow}}$   
               & ${0.341}_{\textcolor{blue}{-22.1\% \downarrow}}$   
               & $\colorBinsCell{0.381}_{\textcolor{red}{+144.2\% \uparrow}}$ \\

\bottomrule
\end{tabular}%
}
\caption{ECE and Gender-ECE on the WinoBias dataset. 
Red values indicate the \% change in ECE score when increasing the model size within the same family.}
\label{tab:bigger_LLMs}
\end{table}

\section{Discussion}

The main finding of this study is that Gemma-2 is the least calibrated model across genders, with larger calibration errors for female entities and worse human alignment. Results from all datasets indicate the model is biased toward male pronouns. Increasing model size improves confidence calibration for male-associated predictions but increases miscalibration for female pronouns, as shown in Table \ref{tab:bigger_LLMs}.

We selected commonly used models in the community, with DeepSeek chosen to analyze bias propagation and confidence during distillation. GPT-J-6B was included due to its minimally filtered training data, enabling analysis of how gender balanced data augmentation affects model confidence. Results indicate that less filtered models, such as GPT-J-6B, exhibit better calibration, suggesting that data filtering or augmentation, \eg gender swapping \cite{zhao2018gender} or using gender neutral, influence model confidence.  For example, using gender-neutral terms results in a less calibrated model, as shown in Table \ref{tab:group_ece_scores} with the GenderLex dataset. Therefore, providing explicit role context, such as the occupational title, reduces uncertainty and improves calibration. In contrast, gender-neutral \ie \texttt{someone} and \texttt{person} result in the highest calibration error due to their inherent semantic ambiguity. DeepSeek-8B, distilled from Llama-3.1-8B, exhibits an overall higher gender calibration error in coreference tasks and less human alignment compared to the base model overall. This indicates that the distillation process negatively impacts model calibration.\footnote{This effect diminishes in larger models \eg DeepSeek-70B.}  

These findings highlight a gap between model confidence and its actual accuracy (\ie human bias) that varies by predicted gender. This confidence--accuracy alignment serves as an indicator of model calibration, reflecting the extent to which the model’s confidence corresponds to its true likelihood of being correct. Accordingly, in our analysis, we complement accuracy-based evaluation by examining confidence calibration disaggregated by gender, which reveals confidence-related bias (\eg overconfidence in predictions for one gender despite comparable correctness). This is particularly important in trust-sensitive decision-making applications, \eg AI-assisted resume screening.  

\section{Calibration} 

As shown in the previous experiments, model output probabilities are often not well-calibrated, especially in our case, a deep neural network \cite{guo2017calibration}. Expected calibration error (ECE) and reliability diagrams can be used to check how well models are calibrated. Therefore, post-hoc calibration methods (\eg Beta calibration \cite{kull2017beta}) are needed, which are used to tune model predicted probabilities to match actual accuracies. For example, let's have 100 predictions with a probability of 80 per cent; in order to have well-calibrated probabilities, about 80 of 100 predictions should be correct. This kind of calibration helps to make the model more reliable and trustworthy. 

Analysis of Table~\ref{tab:winobias_winogender_split} and Figure~\ref{fig:calibration_winobias} shows that the ECE is relatively high, suggesting poor calibration performance on the WinoBias dataset. In order to make the model predictions about its gender bias more reliable and trustworthy, we employ a simple post-hoc calibration method, Beta calibration~\cite{kull2017beta} on top of the data. The data is split into two: 385 validation and 386 test instances.  As a result, Beta calibration method is able to alleviate calibration issues with all the models, resulting in about three times lower calibration error (ECE) (Figure~\ref{fig:calibration_winobias}). 

Also, data is more evenly spread across confidences, visible from the frequency histogram below the reliability diagrams.
Interestingly, Gemma-2-9B, the model before calibration, is predicting with high confidence towards one or another gender; however, based on human bias, the model should be much less confident about its predictions. As the model tunes its confidence predictions to be around 50 per cent. The model accuracies increase after calibration as follows: GPT-J-6B 69.2\% to 76.9\%, Llama-3.1-8B 65.8\% to 74.9\%, DeepSeek-8B 63.5\% to 69.9\%, Gemma-2-9B 51.6\% to 54.7\%, and Qwen2.5-7B 61.1\% to 76.4\%. This indicates that post-hoc calibration can also increase the model's accuracy (\ie reflecting human bias)  for all models. This is expected, as the post-hoc calibrator is fitted on extra validation data, aligning the model predictions better with human bias. 

Although calibration improves the trustworthiness of a model’s confidence estimates, particularly their reliability and interpretability, techniques such as Beta calibration do not serve as bias mitigation strategies, as these procedures enhance confidence calibration without addressing underlying sources of bias.

\section{Ablation Study}
The effect of sample size on ECE has been studied in prior work on classifier uncertainty calibration~\cite{minderer2021revisiting, roelofs2022mitigating}.
To illustrate this effect in our setting, we conduct a small-scale experiment using a dataset of varying sizes and compute the corresponding ECE.   

Table~\ref{tab:ablation} reports the mean and standard deviation of ECE for five different sample sizes: 50, 100, 150, 250, 500.
For each sample size, we draw 100 subsets without replacement from the full WinoBias dataset, which contains 771 instances. 
As a result, the standard deviation is substantially greater for smaller sample sizes, indicating increased instability when fewer instances are available.

Thus, it is important to notice that with a small number of instances, the bins have less samples and the calibration error estimates are less reliable. As a consequence, both confidence estimates and derived measures, such as gender bias in our case, become unreliable under conditions of limited sample size. Therefore, it is essential to check the combination of complementary measures and ensure that each bin contains a sufficient number of instances to produce stable and meaningful estimates.

\begin{table}
  \centering
  \small                            
  \begin{tabular}{lccccc}
    \toprule
    & \multicolumn{5}{c}{Sample size $N$} \\
    \cmidrule(lr){2-6}
    & 50 & 100 & 150 & 250 & 500 \\
    \midrule
    ECE (mean) & 0.2630 & 0.2522 & 0.2429 & 0.2421 & 0.2401 \\
    ECE (std)  & 0.0381 & 0.0256 & 0.0182 & 0.0154 & 0.0076 \\
    \bottomrule
  \end{tabular}
  \caption{Comparison of ECE means and standard deviations across different sample sizes. Each subset of size
$N$ is sampled 100 times without replacement. Results are reported for the DeepSeek-8B model on the WinoBias dataset.}
  
  \label{tab:ablation}
\end{table}

\section{Conclusion}

In this work, we investigate the alignment between predicted confidence scores and gender bias in large language models. Our results show that calibration error does not always correlate with human bias, and the degree of misalignment varies across models. We also propose a new metric, Gender-ECE, that can serve as an additional tool alongside existing calibration methods to better evaluate gender disparities in pronoun resolution tasks.

 \section{Ethical Statement}
 ECE and other calibration metrics measure how well a model’s predicted confidence aligns with actual outcomes: how accurately the predicted probabilities reflect real-world (\ie human bias). These metrics are primarily designed to assess trustworthiness, focusing on the relationship between confidence and accuracy. Gender-ECE extends this by evaluating calibration separately for subgroups, allowing comparison between, for example, male and female samples. However, these metrics do not fully capture other forms of bias, such as differences in prediction frequency, stereotypical language patterns, or unequal distribution of outcomes. Therefore, complementary fairness-oriented measures are required to comprehensively evaluate whether the model’s behavior is fair across gender groups.

\section*{Acknowledgments}

This work was supported by the Estonian Research Council grant “Developing human-centric digital solutions” (TEM-TA120) and by the Estonian Centre of Excellence in Artificial Intelligence (EXAI), funded by the Estonian Ministry of Education and Research and co-funded by the European Union and the Estonian Research Council via project TEM-TA119. It was funded by the EU H2020 program under the SoBigData++ project (grant agreement No. 871042) and partially funded by the HAMISON project.

\section{Appendix}

\section{Implantation details}} We outline the implementation details of the models used in the experiments. All experiments were done on an A40 GPU (48GB VRAM). Larger models (Llama-3.1-70B) utilized 4 A100 GPUs (each with 80GB of VRAM). We employ all default parameters. Note that because probabilities were extracted directly from logits, sampling settings and other generation parameters had no effect on the results.

\noindent{\textbf{GPT-J-6B}:} \texttt{EleutherAI/gpt-j-6B}\footnote{\url{https://huggingface.co/EleutherAI/gpt-j-6B}}\\
\noindent{\textbf{Llama-3.1-8B}:} \texttt{meta-llama/Meta-Llama-3.1-8B}\footnote{\url{https://huggingface.co/meta-llama/Meta-Llama-3.1-8B}}\\
\noindent{\textbf{Llama-3.1-70B}:} \texttt{meta-llama/Meta-Llama-3.1-70B}\footnote{\url{https://huggingface.co/meta-llama/Meta-Llama-3.1-70B}}\\
\textbf{Gemma-2-9B}: \texttt{google/gemma-2-9b}\footnote{\url{https://huggingface.co/google/gemma-2-9b}}\\
\textbf{Gemma-2-27B}: \texttt{google/gemma-2-27b}\footnote{\url{https://huggingface.co/google/gemma-2-27b}}\\
\textbf{Qwen2.5-7B}: \texttt{Qwen/Qwen2.5-7B}\footnote{\url{https://huggingface.co/Qwen/Qwen2.5-7B}}\\
\textbf{Falcon3-7B}: \texttt{tiiuae/Falcon3-7B-Base}\footnote{\url{https://huggingface.co/tiiuae/Falcon3-7B-Base}}\\
\noindent{\textbf{DeepSeek-R1-Distill-Llama-8B}:} \texttt{deepseek-ai/\\DeepSeek-R1-Distill-Llama-8B}\footnote{\url{https://huggingface.co/deepseek-ai/DeepSeek-R1-Distill-Llama-8B}}\\

\section{Limitations and Future Work}
In this research, we focus solely on pronoun prediction and rely on template-based or structural methods to measure gender bias. Although calibration helps improve the trustworthiness of the model, these techniques do not serve as true bias mitigation strategies. In future work, we plan to address this by proposing post-processing mitigation techniques. In addition, we aim to extend our work to address biases beyond gender, such as stereotypes related to nationality and disability  \cite{nangia2020crows}.  Another limitation is that our framework is limited to English and uses gendered pronouns. Future work will investigate Cross-Lingual Bias Transfer, where English is used as a pivot to train low-resource languages. This potentially propagates stereotypes into the target language, as seen in Estonian, where bias is encoded through lexical suffixes \cite{kaukonen2025aunt}.

\section{Ablation Study}

\subsubsection*{Why do current evaluation methods rely on Winograd-style sentences rather than structured free text?}
In this work, we utilize Winograd-style sentences as a controlled experiment (gender bias benchmark), where we can measure the model’s confidence. This type of benchmark helps reveal how a gender token can affect the confidence score. However, when longer, structured free text is used, it introduces noise, making the confidence less reliable.  To validate this, we experimented with an image captioning task (a similar form of synthetic language) leveraging the full gender and bias context of the caption rather than isolating the probability of a single pronoun.

To implement this, we rely on the Vision LLM image captioner Gemma-3 \cite{team2025gemma} (12B and 27B) to create the textual dataset. We filtered 5K images from the COCO Caption dataset validation set \cite{chen2015microsoft} to include only 773 images that contain people. We utilize free-form caption generation, where gender is expressed naturally in the model’s output. Following prior work \cite{sabir2023women}, we extract the mean probability of gendered captions by computing the mean token-level probability for each generated caption.

As shown in Table \ref{tab:bias_image_caption}, the ECE and ICE scores are very similar across the different models, suggesting that the predicted probabilities are tightly clustered. This indicates that ECE may not be sensitive enough to detect subtle differences in model calibration under this experimental setup.

\section{Additional Information}

\subsubsection*{What motivates the introduction of Gender-Aware ECE in this work, and in what ways does it differ from other grouped variants of ECE?} In this section, we compare the proposed Gender-ECE additionally with class-conditioned ECE (cc-ECE) and highlight the advantages of each in the evaluation. 
In the main paper, Gender-ECE was compared with MacroCE, as this was the measure Gender-ECE was inspired by. In this Section, we compare Gender-ECE to class-conditioned calibration error (cc-ECE), which is similar in manner, but divides the instances into two groups based on the true label.
Another variant is classwise-ECE~\cite{kull2019beyond, nixon2019measuring}, which computes ECE separately for each predicted class and then averages the results. However, in the binary setting, classwise-ECE collapses to the standard ECE, making the two metrics equivalent.

Class-conditioned calibration error (cc-ECE) is similar to ECE. Instead of dividing all classes jointly into bins, the bins are also divided based on the true label. 
Formally:
\begin{equation}
{\small
\text{cc-ECE} = \frac{1}{2}\left(\text{ECE}_{\text{male}} + \text{ECE}_{\text{female}}\right)
}
\label{eq:CCE}
\end{equation}
\begin{equation*}
{\small
\begin{aligned}
\text{ECE}_{\text{male}}   &= \sum_{m=1}^{M} \frac{|B_m|}{n} \left|\text{acc}(B_m) - \text{conf}(B_m)\right|, && \forall {y}_i = 1, \\
\text{ECE}_{\text{female}} &= \sum_{m=1}^{M} \frac{|B_m|}{n} \left|\text{acc}(B_m) - \text{conf}(B_m)\right|, && \forall {y}_i = 0.
\end{aligned}
}
\end{equation*}
\noindent{where} ${y}_i$ is true label, and [1/0] correspond to male/female true label.

These two metrics are grouped variants of ECE, distinguished by where they apply the partitioning: cc-ECE groups examples based on the ground-truth labels, whereas Gender‑ECE groups them according to the model’s predicted labels (model preference).

\begin{table}[t!]
    \centering
    \begin{tabular}{lcccc}
        \toprule
        \multirow{2}{*}{\textbf{Model}} 
          & \multicolumn{2}{c}{Gemma-3-9B} 
          & \multicolumn{2}{c}{Gemma-3-27B} \\
        \cmidrule(lr){2-3}\cmidrule(lr){4-5}
        & ECE & ICE & ECE & ICE \\
        \midrule
        GPT-J-6B       
          & \colorBinsCell{0.240} 
          & \colorBinsCell{0.513} 
          & \colorBinsCell{0.179} 
          & \colorBinsCell{0.506} \\
        Llama-3.1-8B   
          & \colorBinsCell{0.178} 
          & \colorBinsCell{0.498} 
          & \colorBinsCell{0.178} 
          & \colorBinsCell{0.497} \\
        Gemma-2-9B     
          & \colorBinsCell{0.181} 
          & \colorBinsCell{0.504} 
          & \colorBinsCell{0.183} 
          & \colorBinsCell{0.503} \\
        Qwen2.5-7B    
          & \colorBinsCell{0.189} 
          & \colorBinsCell{0.501} 
          & \colorBinsCell{0.181} 
          & \colorBinsCell{0.503} \\
        DeepSeek-8B 
          & \colorBinsCell{0.179} 
          & \colorBinsCell{0.495} 
          & \colorBinsCell{0.178} 
          & \colorBinsCell{0.497} \\
        Falcon3-7B     
          & \colorBinsCell{0.176} 
          & \colorBinsCell{0.491} 
          & \colorBinsCell{0.182} 
          & \colorBinsCell{0.497} \\
        \bottomrule
    \end{tabular}
 \caption{ECE and ICE scores across caption generation models using full sentence context with different genders. The results are highly similar, suggesting that ECE is not sensitive enough to capture differences when the only variation between sentences is gender.}

    \label{tab:bias_image_caption}
\end{table}

\begin{table*}[h!]
\centering
\small
    \resizebox{\textwidth}{!}{%
    \begin{tabular}{lccc|ccc|ccc}
    \toprule
    & \multicolumn{3}{c}{WinoBias} & \multicolumn{3}{c}{WinoGender} & \multicolumn{3}{c}{GenderLex} \\
    \cmidrule(lr){2-4} \cmidrule(lr){5-7} \cmidrule(lr){8-10}
    Model & MacroCE & cc-ECE & Gender-ECE &  MacroCE & cc-ECE & Gender-ECE & MacroCE & cc-ECE & Gender-ECE \\
    \midrule
    GPT-J-6B            & \underline{0.444} & \underline{0.356} & \underline{0.164} & 0.473 & 0.427 & \underline{0.118} & \underline{0.453} & 0.375 & 0.076\\
    Llama-3.1-8B       & 0.460 & 0.378 & 0.214 & 0.475 & 0.414 & 0.138 & 0.466 & \underline{0.365} & 0.111\\
    Gemma-2-9B          & \textbf{0.490} & \textbf{0.482} & \textbf{0.297} & \textbf{0.486} & \textbf{0.494} & \textbf{0.396} & \textbf{0.493} & \textbf{0.479} & \textbf{0.267}\\
    Qwen2.5-7B        & 0.442 & 0.362 & 0.190 & \underline{0.461} & \underline{0.411} & 0.129 & 0.476 & 0.422 & 0.107\\
    Falcon3-7B     & 0.452 & 0.357 & 0.149 & 0.474 & 0.429 & 0.176 & 0.491 & 0.435 & 0.149\\
    DeepSeek-8B      & 0.478 & 0.382 & 0.218 & 0.470 & 0.415 & 0.135 & 0.461 & 0.387 & \underline{0.090}\\
    \bottomrule
    \end{tabular}
    }

    \caption{Comparison of group variants of ECE measure: MacroCE, class-conditional ECE, and Gender-ECE.}
    \label{tab:cce_comparison}
    
\end{table*}

\begin{table*}[t!]
    \centering
    \resizebox{\textwidth}{!}{%
    \begin{tabular}{lccc|ccc|ccc|ccc}
        \toprule
         & \multicolumn{3}{c}{GPT-J-6B} 
         & \multicolumn{3}{c}{Llama-3.1-8B} 
         & \multicolumn{3}{c}{DeepSeek-8B} 
         & \multicolumn{3}{c}{Gemma-2-9B} \\
        \cmidrule(lr){2-4} \cmidrule(lr){5-7} \cmidrule(lr){8-10} \cmidrule(lr){11-13}
        \underline{Metric} 
            & Someone & Person & Occ 
            & Someone & Person & Occ 
            & Someone & Person & Occ 
            & Someone & Person & Occ \\
        \midrule
        ECE 
            & \colorBinsCell{0.144} & 0.063 & 0.076
            & 0.134 & \colorBinsCell{0.138} & 0.111
            & \colorBinsCell{0.139} & 0.130 & 0.085
            & 0.364 & \colorBinsCell{0.367} & 0.327 \\
        MacroCE 
            & \colorBinsCell{0.484} & 0.476 & 0.453
            & \colorBinsCell{0.483} & 0.478 & 0.466
            & \colorBinsCell{0.482} & 0.481 & 0.461
            & 0.493 & 0.493 & \colorBinsCell{0.494} \\
        ICE 
            & \colorBinsCell{0.454} & 0.442 & 0.374
            & 0.436 & \colorBinsCell{0.445} & 0.371
            & \colorBinsCell{0.452} & 0.443 & 0.369
            & \colorBinsCell{0.397} & 0.393 & 0.390 \\
        Brier Score 
            & 0.331 & 0.341 & \colorBinsCell{0.432}
            & 0.358 & 0.348 & \colorBinsCell{0.446}
            & 0.352 & 0.361 & \colorBinsCell{0.470}
            & 0.560 & \colorBinsCell{0.581} & 0.574 \\
        Group-ECE
            & \colorBinsCell{0.138} & 0.077 & 0.076
            & \colorBinsCell{0.132} & 0.130 & 0.111
            & \colorBinsCell{0.138} & 0.137 & 0.090
            & \colorBinsCell{0.450} & 0.351 & 0.267 \\     
            + Male
            & \colorBinsCell{0.106} & 0.031 & 0.085
            & \colorBinsCell{0.115} & 0.071 & 0.112
            & 0.048 & 0.065 & \colorBinsCell{0.074}
            & 0.363 & \colorBinsCell{0.367} & 0.330\\  
            + Female
            & \colorBinsCell{0.170} & 0.122 & 0.066
            & 0.148 & \colorBinsCell{0.190} & 0.109
            & \colorBinsCell{0.228} & 0.210 & 0.106
            & \colorBinsCell{0.536} & 0.335 & 0.204 \\ 
        Human alignment 
            & \colorBinsCell{0.598} & 0.616 & 0.715
            & 0.638 & \colorBinsCell{0.598} & 0.727
            & \colorBinsCell{0.578} & 0.596 & 0.687
            & \colorBinsCell{0.603} & 0.606 & 0.618\\
        \bottomrule
    \end{tabular}%
    }
    \caption{(Table 5 full result). LLMs exhibit higher calibration errors when presented with gender-neutral terms (\texttt{Someone}, \texttt{Person}), resulting in higher prediction uncertainty. The red color indicates the highest (worst) calibration error and the (best) human alignment.}

    \label{tab:model_occ_person_someone_full}
\end{table*}

\begin{itemize}
    \item \textbf{cc-ECE} conditions on the \textit{true} label, computing a per‑class ECE over  
    $\mathcal{T}_{y} = \{\, i \mid y_i = y \}$ and then averaging.  
    For each true class $y$, the examples in $\mathcal{T}_y$ are divided into $B$ equal‑width confidence bins, indexed by $b \in \{1, \dots, B\}$, based on their predicted confidence scores.  
    The comparison statistic within each bin is the \textit{recall}  
    $\hat{r}_{y,b} = P(\hat{y} = y \mid y)$,  
    therefore cc-ECE measures whether the model’s confidence correctly anticipates  
    \textbf{false negatives}.
    \item \textbf{Gender‑ECE} (ours) conditions on the \textit{predicted} label, computing an ECE over  
    $\mathcal{S}_{g} = \{\, i \mid \hat{g}_i = g \}$, where $g \in \{\text{male}, \text{female}\}$.  
    As with cc-ECE, the examples in each $\mathcal{S}_g$ are grouped into $B$ confidence bins indexed by $b$, according to their predicted confidence scores.  
    The comparison statistic within each bin is the \textit{precision}  
    $\hat{p}_{g,b} = P(y = \hat{y} \mid \hat{g})$,  
    therefore Gender‑ECE measures whether the model’s confidence correctly anticipates  
    \textbf{false positives}.
\end{itemize}

In our case, cc-ECE evaluates whether, when the \textit{true} token belongs to a given class (\eg male or female), the model’s confidence accurately reflects how often it fails to retrieve it.  This makes cc-ECE a \textbf{recall‑sensitive} calibration measure, particularly relevant in settings where the failure to identify certain classes, such as overlooking male or female entities in coreference resolution, may lead to systematic bias.
Table~\ref{tab:cce_comparison} compares different variants of grouped ECE measures: MacroCE, class-conditioned ECE and Gender-ECE. All the methods follow similar trend, agreeing on the worst performing measure.

\noindent{\textbf{Gender Neutral Experiment.}} Table \ref{tab:model_occ_person_someone_full} shows that the \texttt{Someone} term consistently yields the weakest calibration, producing the highest ECE and Gender-ECE (Group-ECE)  across nearly all models. This gender-neutral formulation leads to increased prediction uncertainty and the largest Female–Male calibration gaps, particularly for Gemma-2-9B and the distillation model DeepSeek-8B. Human-alignment scores are also lowest under the gender-neutral terms \texttt{person} and \texttt{someone}, indicating that models deviate most from human gender bias distributions when given minimal contextual cues.

\noindent\textbf{DeepSeek-based Distillation Models.} As shown in Table \ref{tab:ece_deepseek}, the impact of distillation on calibration and fairness is highly scale-dependent. For the 8B and 14B models, the results are inconsistent: the significant decrease in Female-ECE is offset by a rising Male-ECE. This implies that smaller models struggle to balance group-specific fairness with general calibration. In contrast, the 32B and 70B models show a consistent trend, with a lower Gender-ECE and overall ECE score.

\noindent{\textbf{What is the relationship between calibration performance and actual bias mitigation?}} Calibration and bias mitigation are related but not equivalent. A better-calibrated model reflects more accurate confidence scores, but this doesn’t guarantee reduced bias. A model can be well calibrated yet still exhibit bias in its predictions across gender groups. In our work, we use calibration to analyze how model confidence shifts with gendered inputs, but we do not assume that calibration alone mitigates bias.

\noindent{\textbf{Input Representation.}} In this work, “calibration” is not based on prompting, but rather from the direct extraction of token probabilities from the model’s logits. This \textbf{provides deterministic probability estimates} for the pronouns of interest, which are subsequently evaluated for calibration using ECE. Here, calibration refers to assessing whether the probability assigned by the model (\eg 0.8 for “she”) aligns with the empirical frequency with which the corresponding prediction is correct.

\noindent{\textbf{Bias Definition.} Following the taxonomy of \cite{gallegos2024bias}, we frame bias as a form of representational harm arising from miscalibration in pronoun resolution. Such miscalibration manifests when the model produces systematically over‑ or under‑confident predictions for one gender group compared to another. We quantify this effect using Gender-ECE (\ie Group‑ECE) and the calibration bias gap.

\begin{table}[t]
\centering
\resizebox{\columnwidth}{!}{%
\begin{tabular}{lccccc}
\toprule
Model & ECE & \multicolumn{3}{c}{Gender-ECE} & Human \\
\cmidrule(lr){3-5}
              &              & Group & M & F &            \\ \midrule
Llama-3.1-8B   & 0.193 & 0.214 & 0.179 & \colorBinsCell{0.249} &  \colorBinsCell{0.662} \\
+ DeepSeek-8B & \colorBinsCell{0.240} & \colorBinsCell{0.218} & \colorBinsCell{0.225} & 0.182 &0.648 \\ \midrule
Qwen2.5-14B   & 0.223 & \colorBinsCell{0.218} & 0.226 & \colorBinsCell{0.209} &  0.644 \\
+ DeepSeek-14B    & \colorBinsCell{0.226} & 0.179 & \colorBinsCell{0.262} & 0.095 & \colorBinsCell{0.664} \\ \midrule
Qwen2.5-32B   & \colorBinsCell{0.186} & \colorBinsCell{0.153} & \colorBinsCell{0.222} & \colorBinsCell{0.085} & \colorBinsCell{0.669} \\
+ DeepSeek-32B    & 0.176 & 0.139 & 0.216 & 0.062 & 0.665 \\ \midrule
Llama-3.3-70B-Inst  & \colorBinsCell{0.224} & \colorBinsCell{0.218} & \colorBinsCell{0.230} & \colorBinsCell{0.205} & \colorBinsCell{0.680} \\
+ DeepSeek-70B-Inst   & 0.203 & 0.177 & 0.229 & 0.124 & 0.667 \\ \bottomrule
\end{tabular}%
}

\caption{ECE and Gender-ECE results on the WinoBias dataset for each base model and its corresponding version after distillation via DeepSeek-R1. While Female-ECE consistently decreases across all models, improvements in overall ECE are only realized in the larger 32B and 70B models.}

\label{tab:ece_deepseek}
\end{table}

\section{Extra Calibration Result}

Figure~\ref{fig:calibration_winobias_es} shows the results of uncalibrated and calibrated reliability diagrams of the Winobias dataset using equal-size binning. Equal-size binning means that each of the bins has an equal number of instances. This is useful, because some probability ranges (\eg [0.9–1.0]) may have far more predictions than others. Equal-size binning ensures that each bin has statistical weight, avoiding empty or low-count bins that skew ECE. The results remain very similar to those obtained with equal-width binning. 

Figure~\ref{fig:calibration_winobias_iso} compares uncalibrated and calibrated results of the Winobias dataset using isotonic regression~\cite{zadrozny2002transforming}. The results are similar to Beta calibration~\cite{kull2017beta}. The uncalibrated confidences deviate substantially from the diagonal, showing that the classifier’s predicted probabilities do not match the observed accuracies. In most bins, the model exhibits over-confidence, assigning higher confidence scores than the corresponding empirical accuracy. After applying isotonic calibration, the bin accuracies align much more closely with the diagonal, indicating improved probability calibration.

In addition, we experimented with temperature scaling \cite{guo2017calibration}, but the optimization converged to an excessively large temperature value, effectively collapsing all predicted probabilities toward 0.5. Platt scaling~\cite{platt2000probabilistic} showed similar, but opposite behavior, driving the confidence scores toward the 0 and 1 rather than producing a meaningful calibration.

\begin{figure*}[h!]
\small
    \centering
    \begin{minipage}{\textwidth}
        \centering
        \subfloat[ECE: 0.143]{\includegraphics[width=0.20\textwidth]{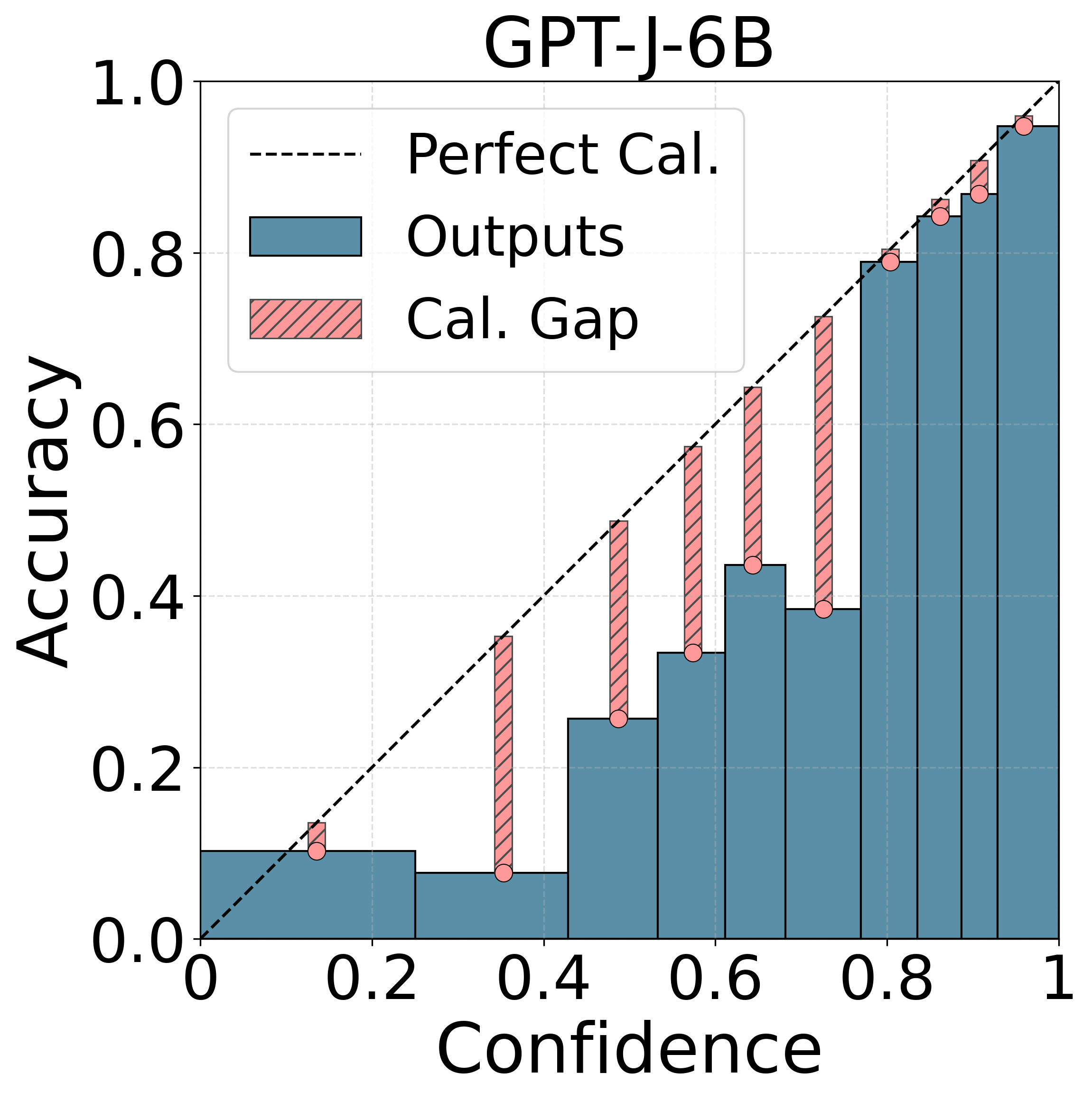}}\hfill
        \subfloat[ECE: 0.178]{\includegraphics[width=0.20\textwidth]{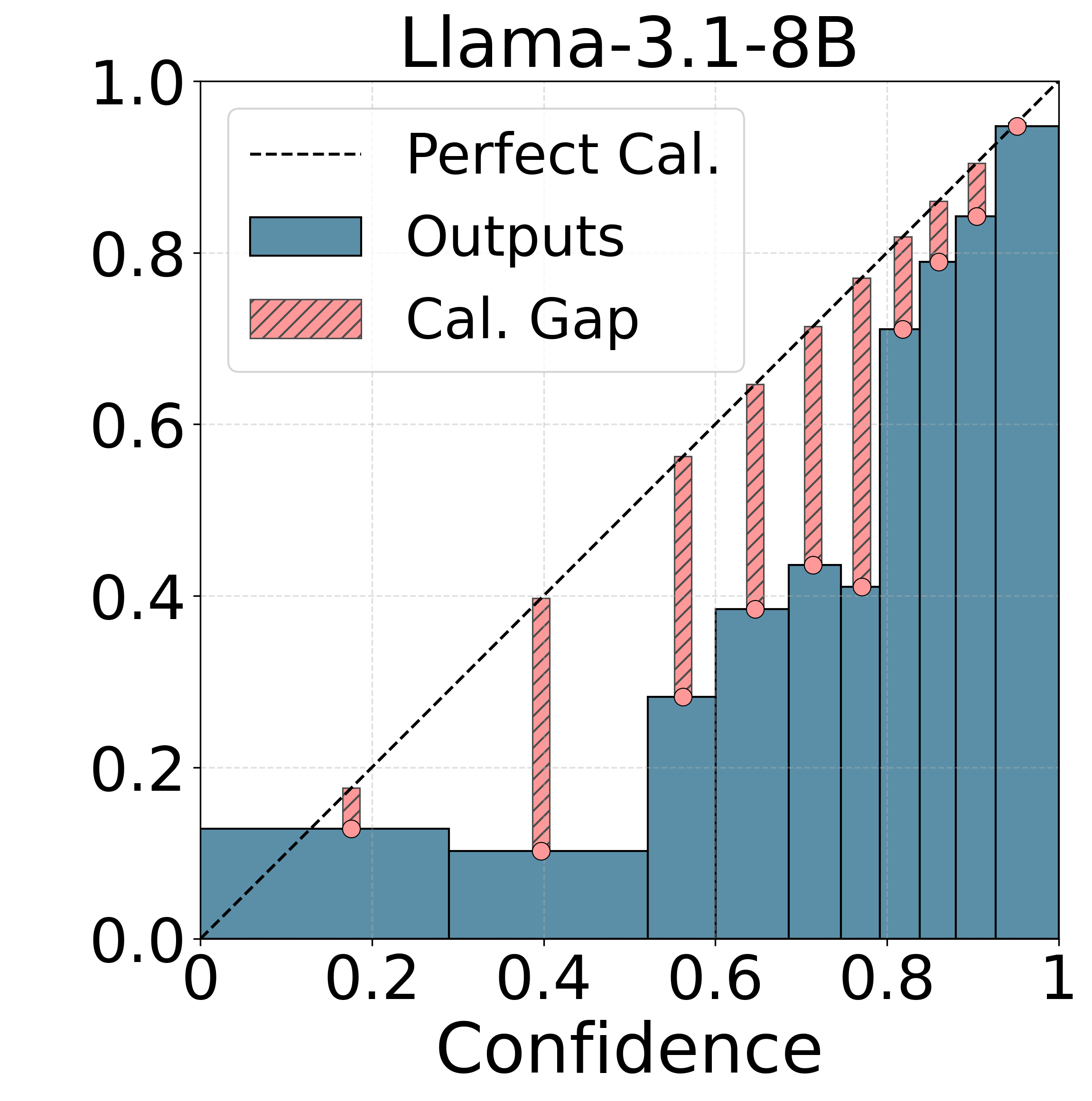}}\hfill
        \subfloat[ECE: 0.224]{\includegraphics[width=0.20\textwidth]{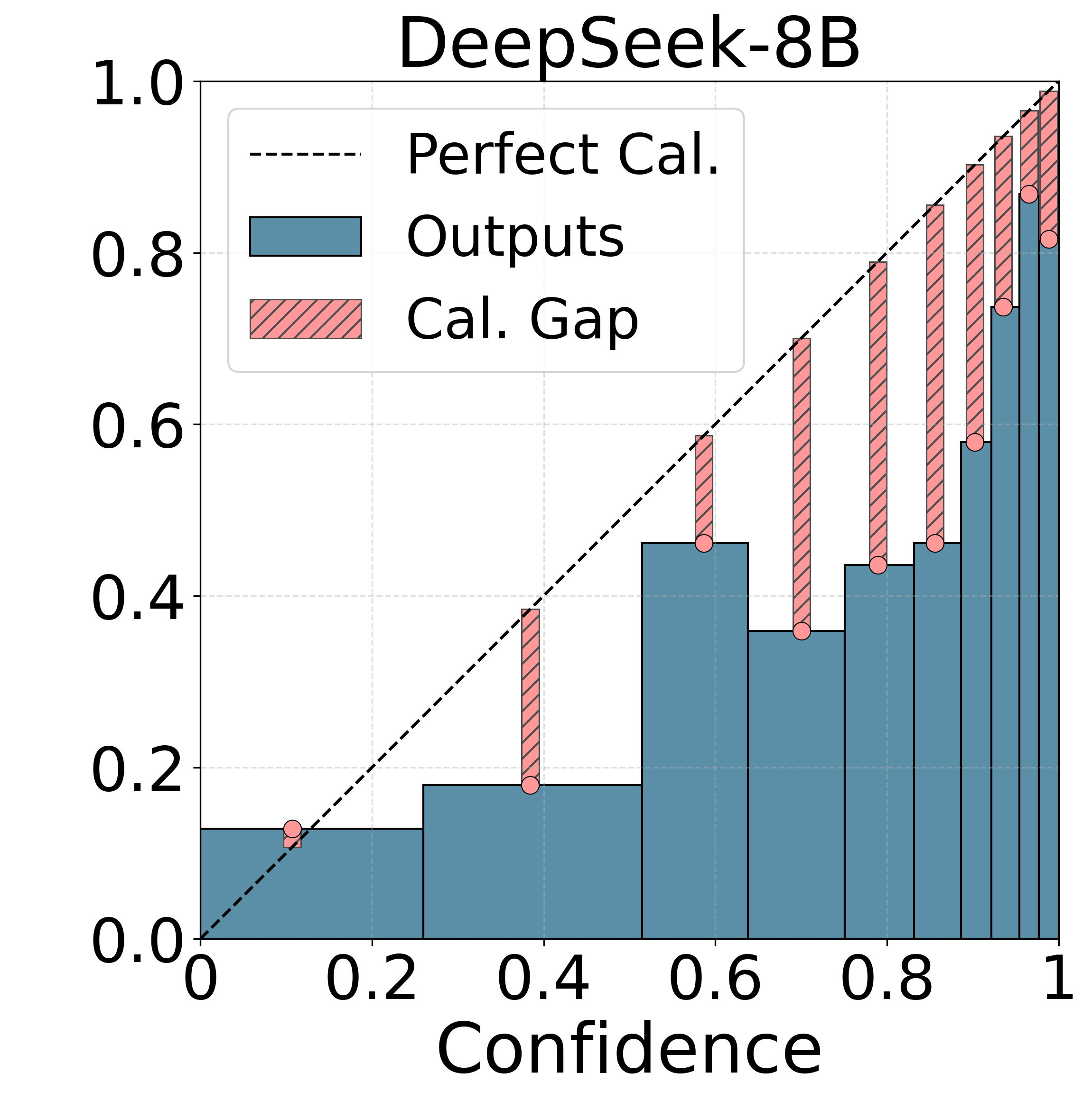}}\hfill
        \subfloat[ECE: 0.423]{\includegraphics[width=0.20\textwidth]{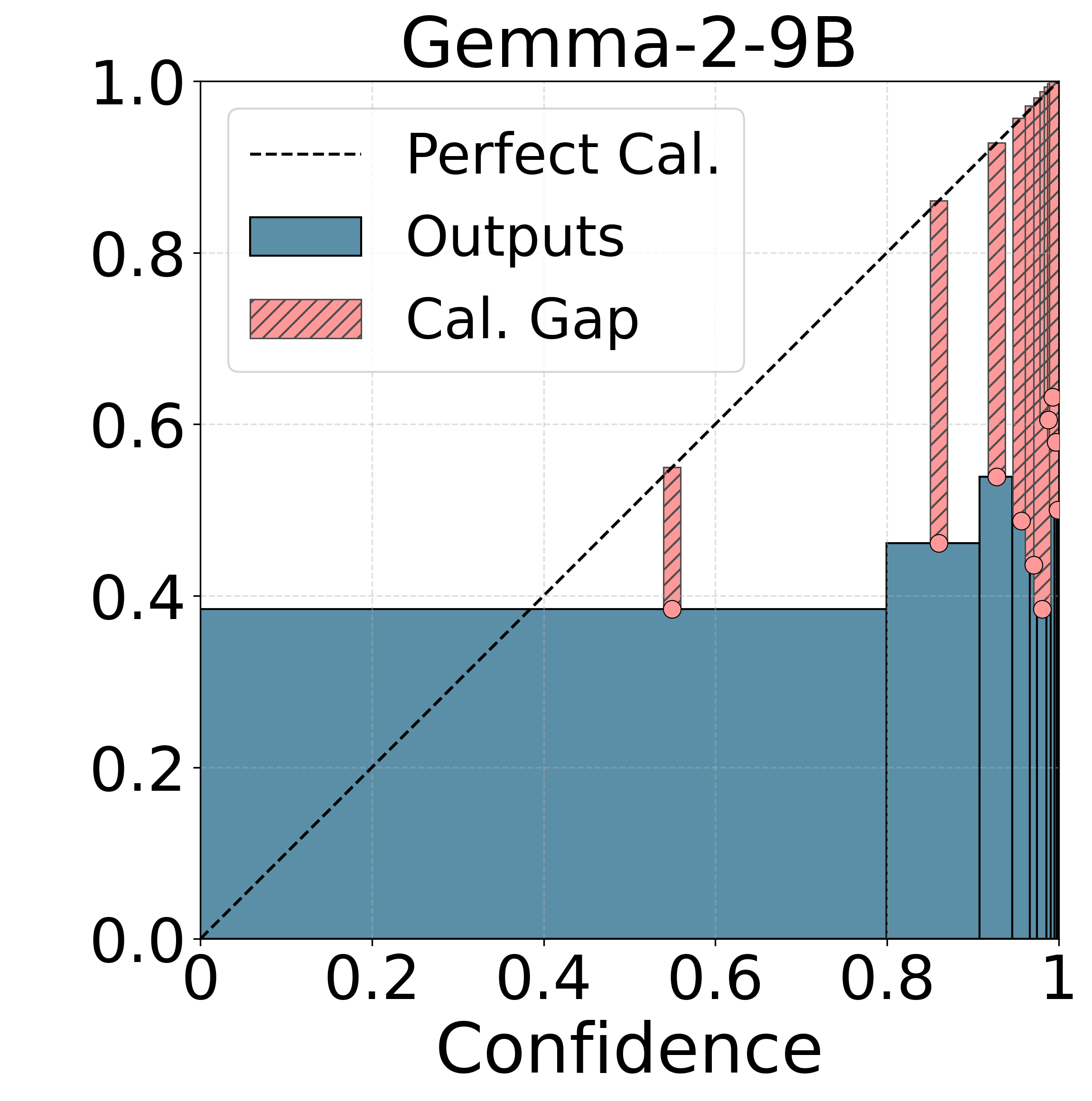}}\hfill
        \subfloat[ECE: 0.234]{\includegraphics[width=0.20\textwidth]{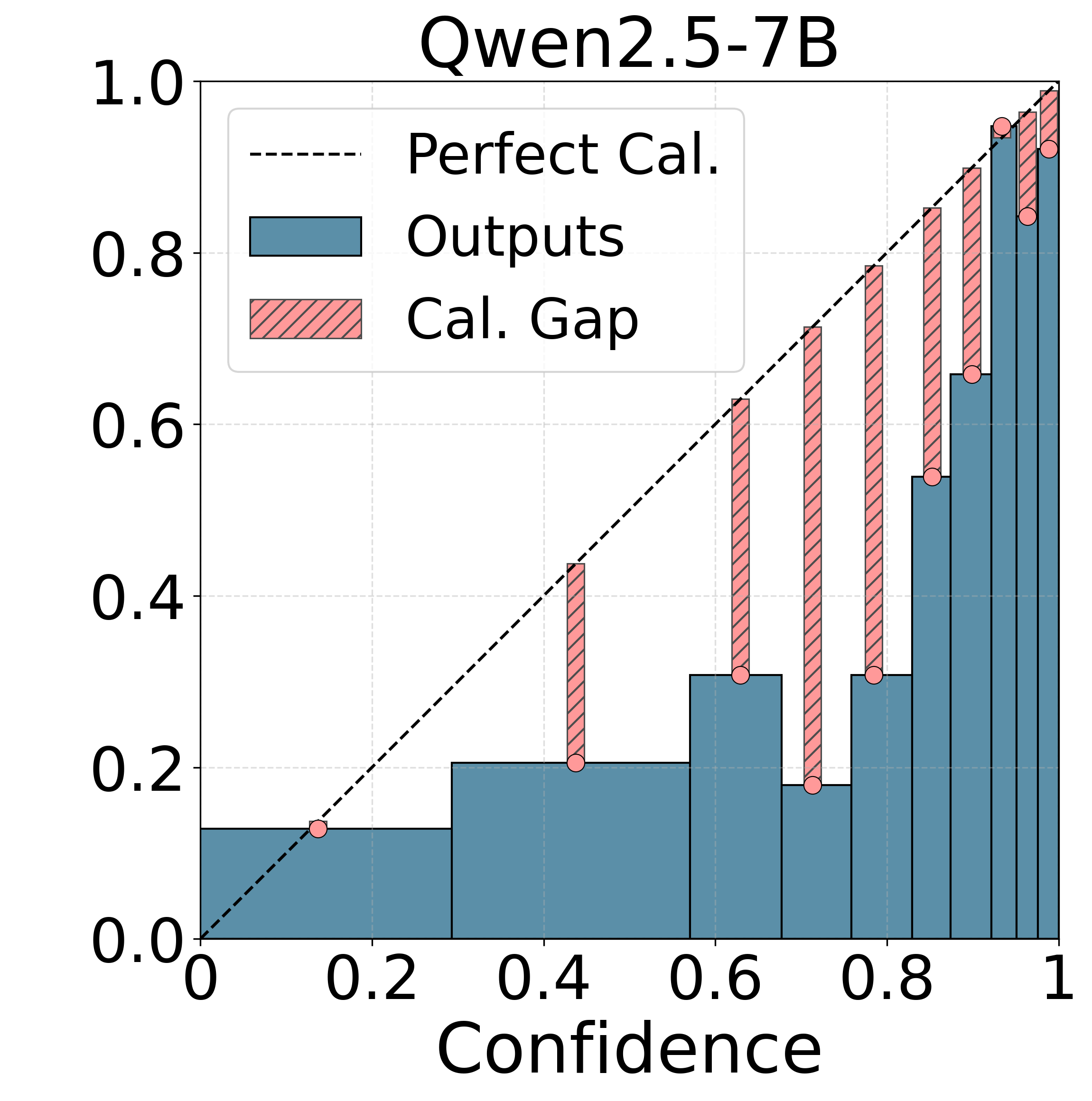}}
    \end{minipage}

    \vspace{-5pt} 
    
    \begin{minipage}{\textwidth}
        \centering
        \subfloat[ECE: 0.064]{\includegraphics[width=0.20\textwidth]{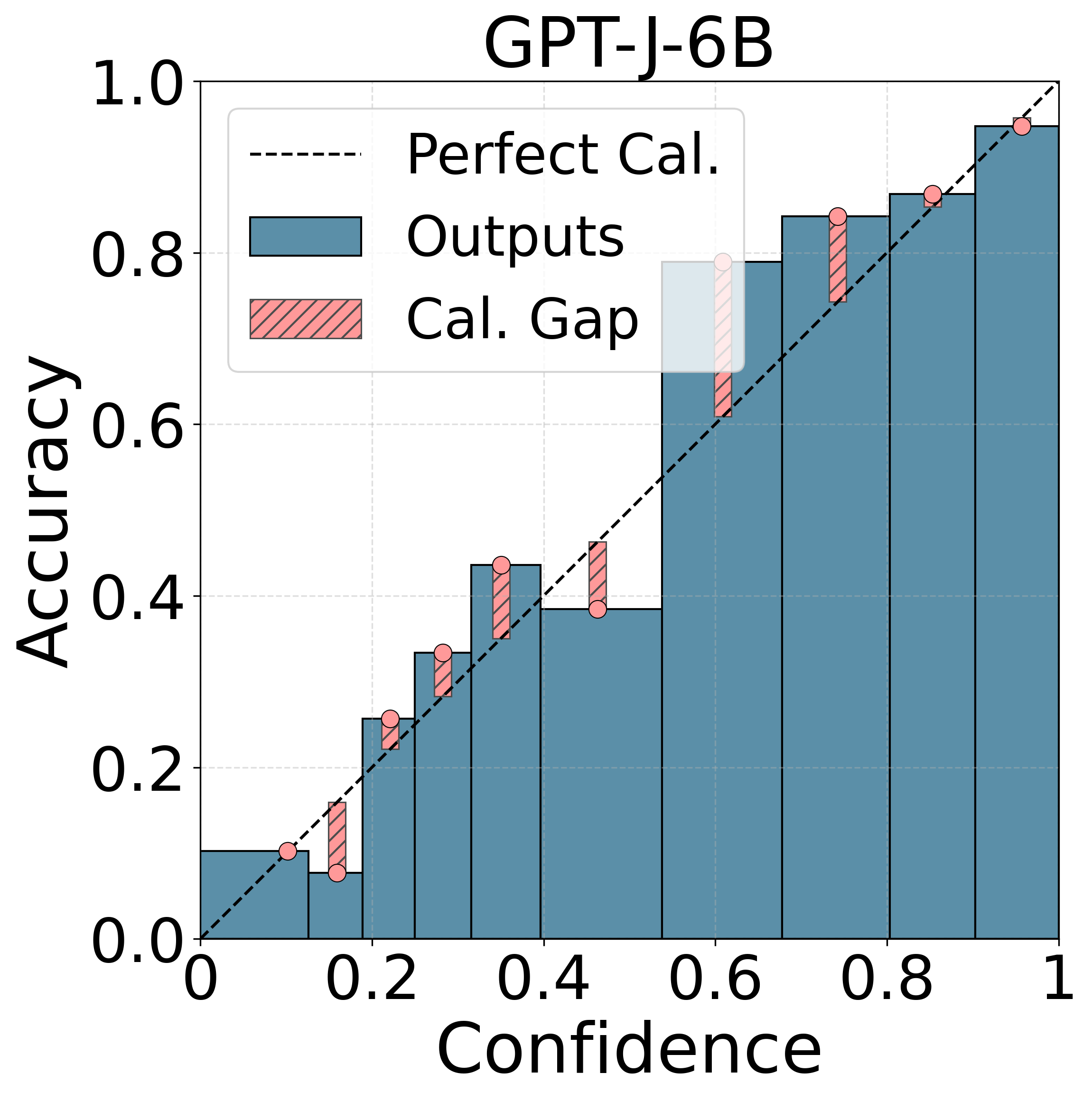}}\hfill
        \subfloat[ECE: 0.048]{\includegraphics[width=0.20\textwidth]{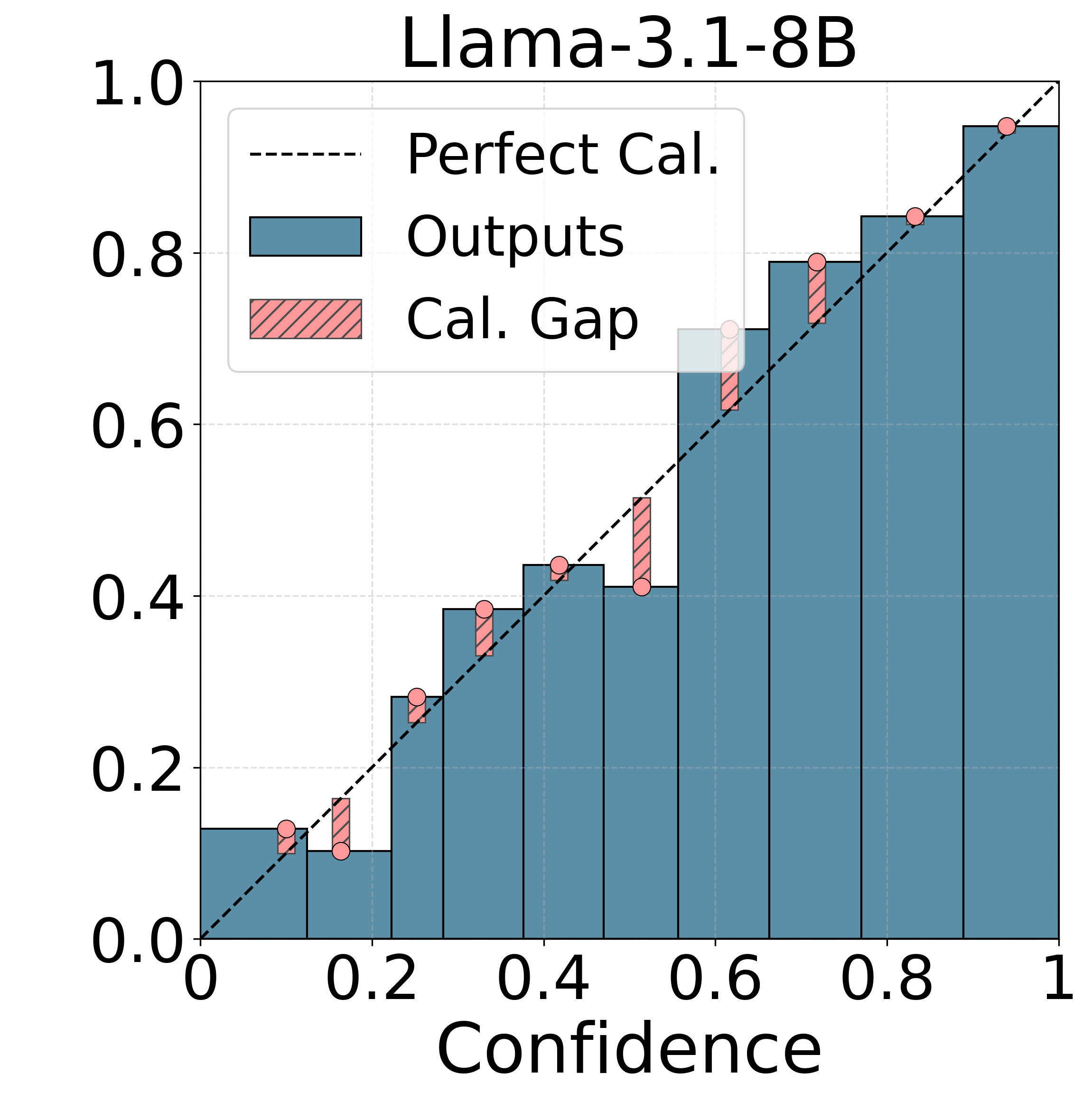}}\hfill
        \subfloat[ECE: 0.061]{\includegraphics[width=0.20\textwidth]{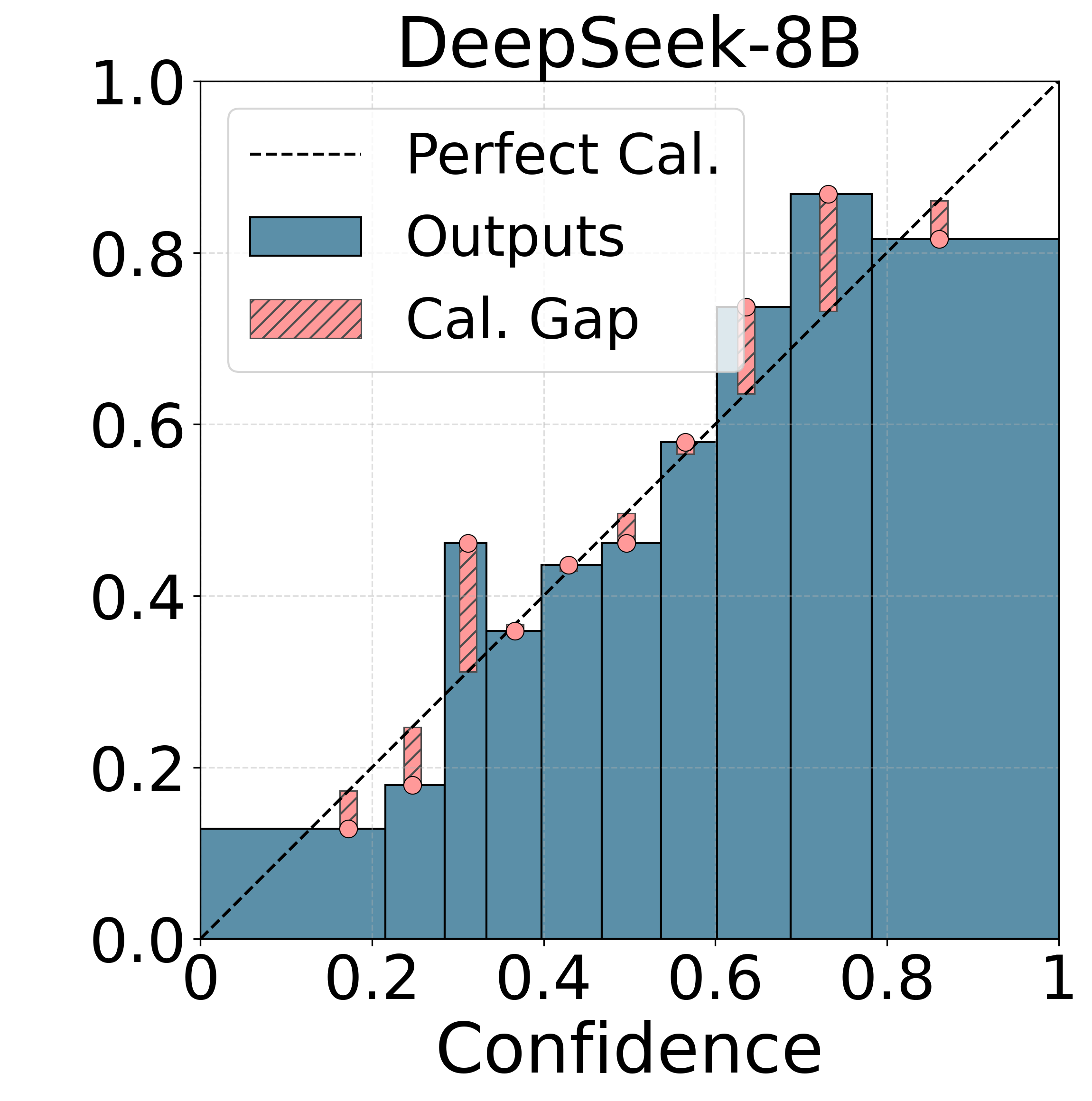}}\hfill
        \subfloat[ECE: 0.063]{\includegraphics[width=0.20\textwidth]{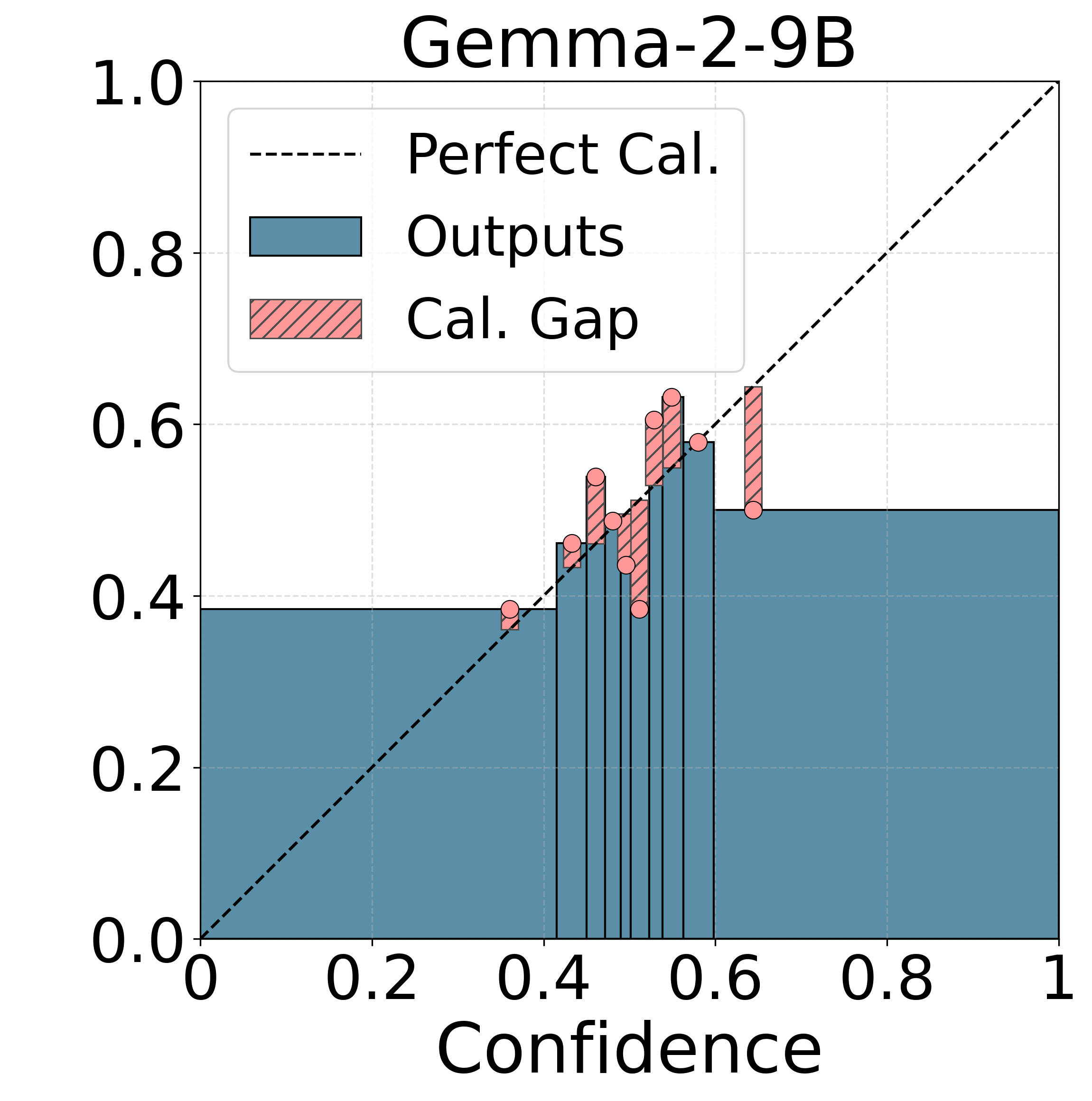}}\hfill
         \subfloat[ECE: 0.071]{\includegraphics[width=0.20\textwidth]{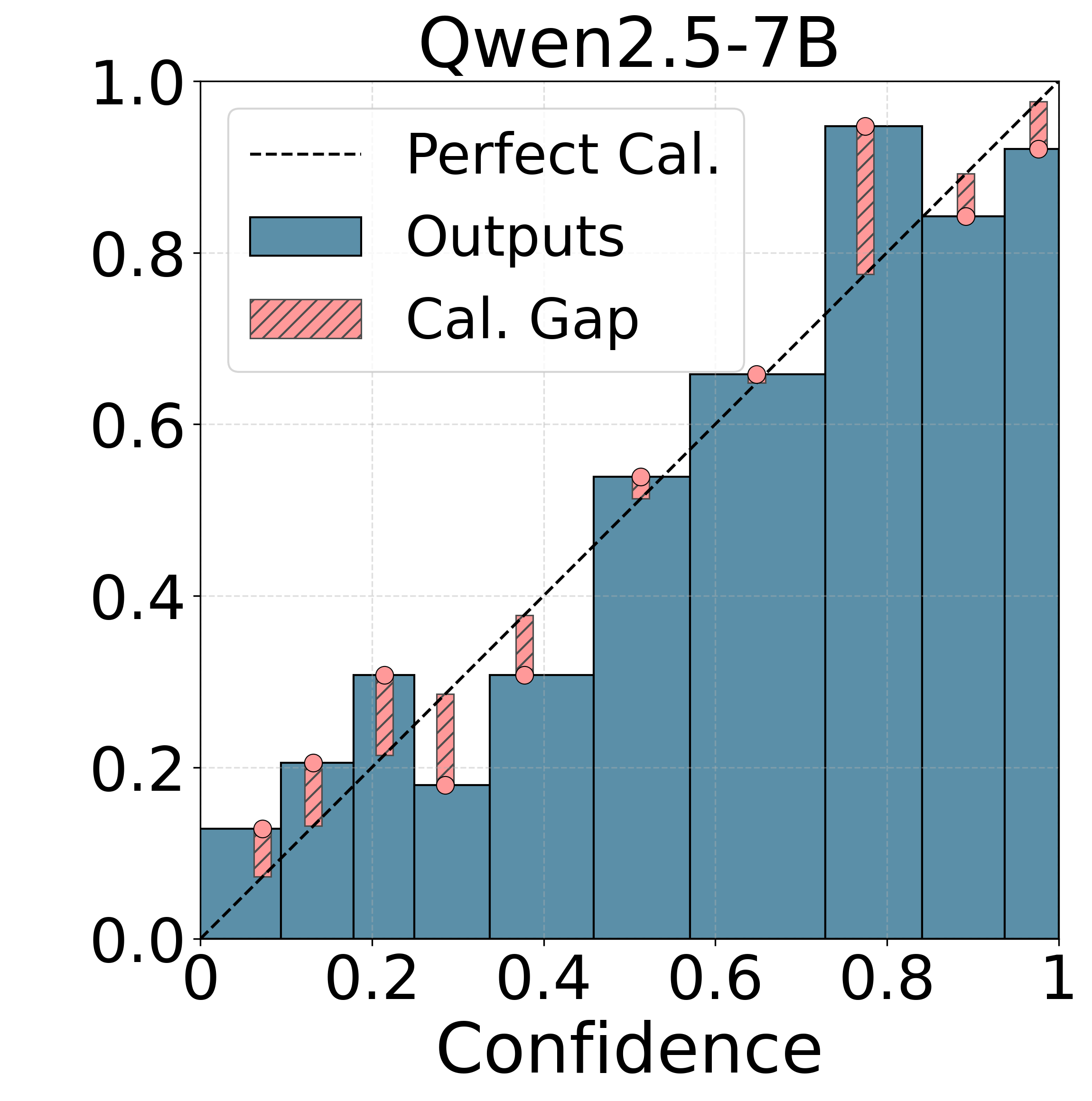}}
    \end{minipage}

      \caption{ Comparison of uncalibrated (Top) and Beta calibrated~\cite{kull2017beta} (Bottom) gender bias probabilities for WinoBias dataset. Instances are divided into 10 equal-size bins instead of equal-width bins. The red bar shows the distance to perfectly calibrated probabilities.}
    \label{fig:calibration_winobias_es}

    \begin{minipage}{\textwidth}
        \centering
        \subfloat[ECE: 0.148]{\includegraphics[width=0.20\textwidth]{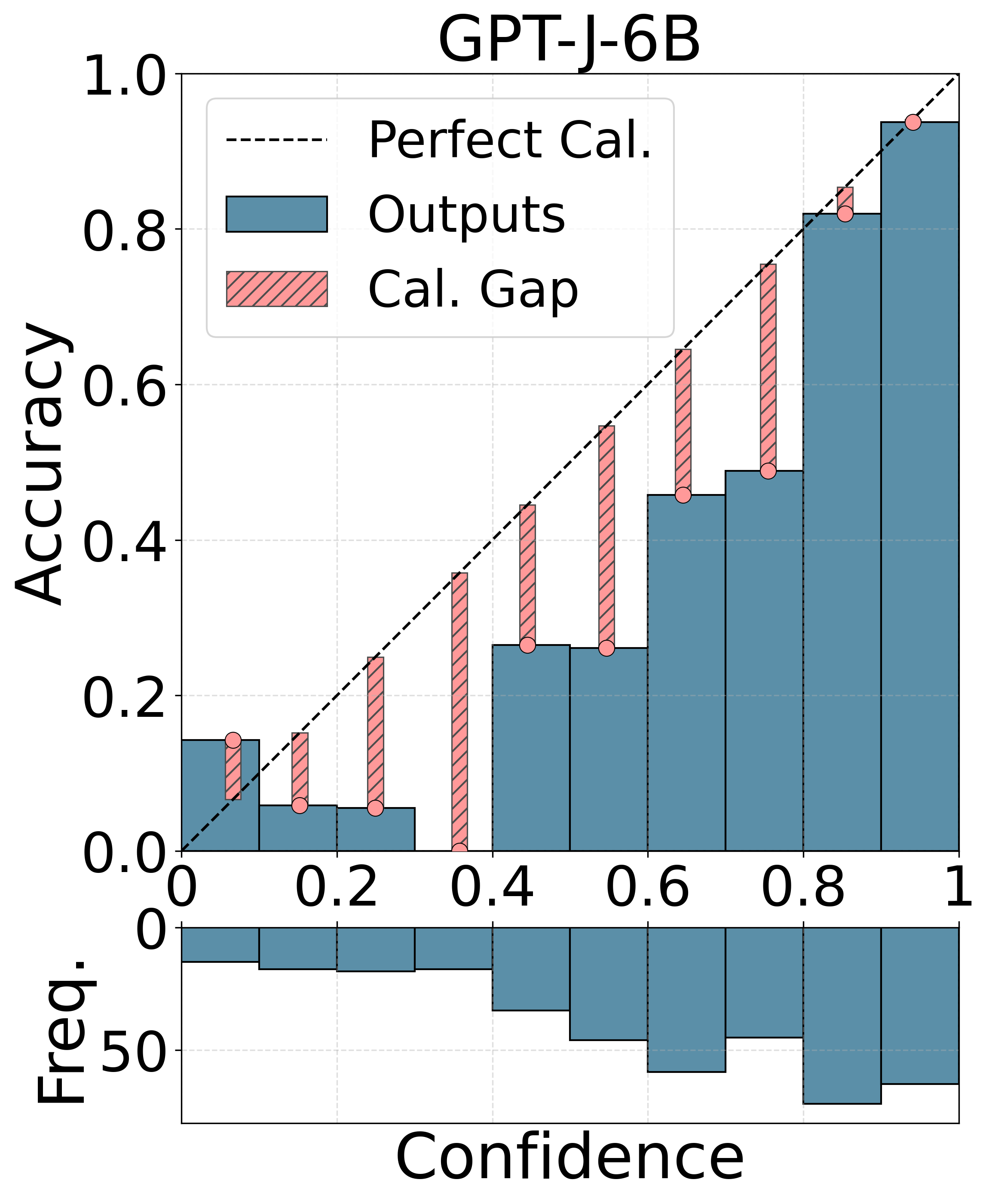}}\hfill
        \subfloat[ECE: 0.196]{\includegraphics[width=0.20\textwidth]{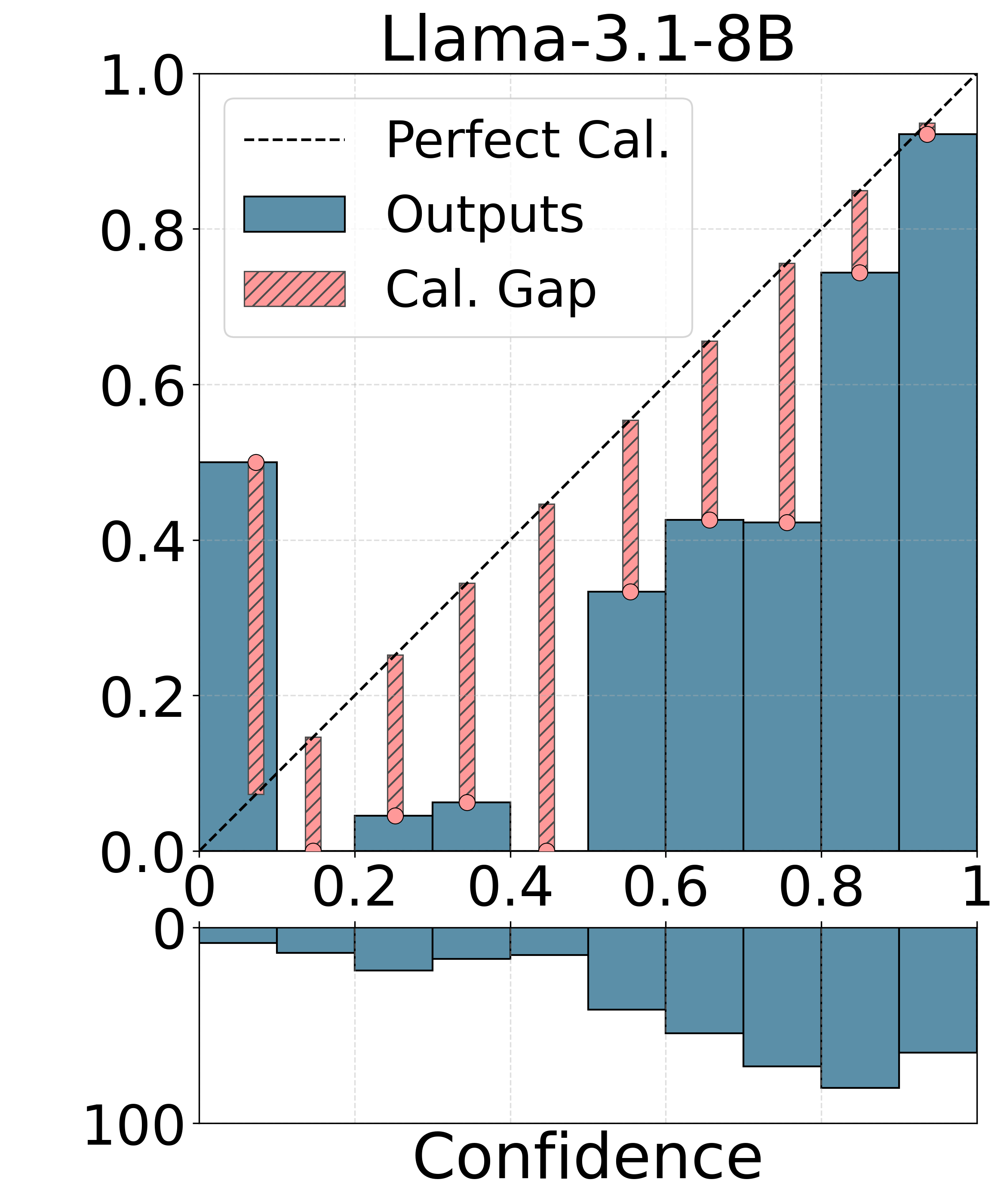}}\hfill
        \subfloat[ECE: 0.235]{\includegraphics[width=0.20\textwidth]{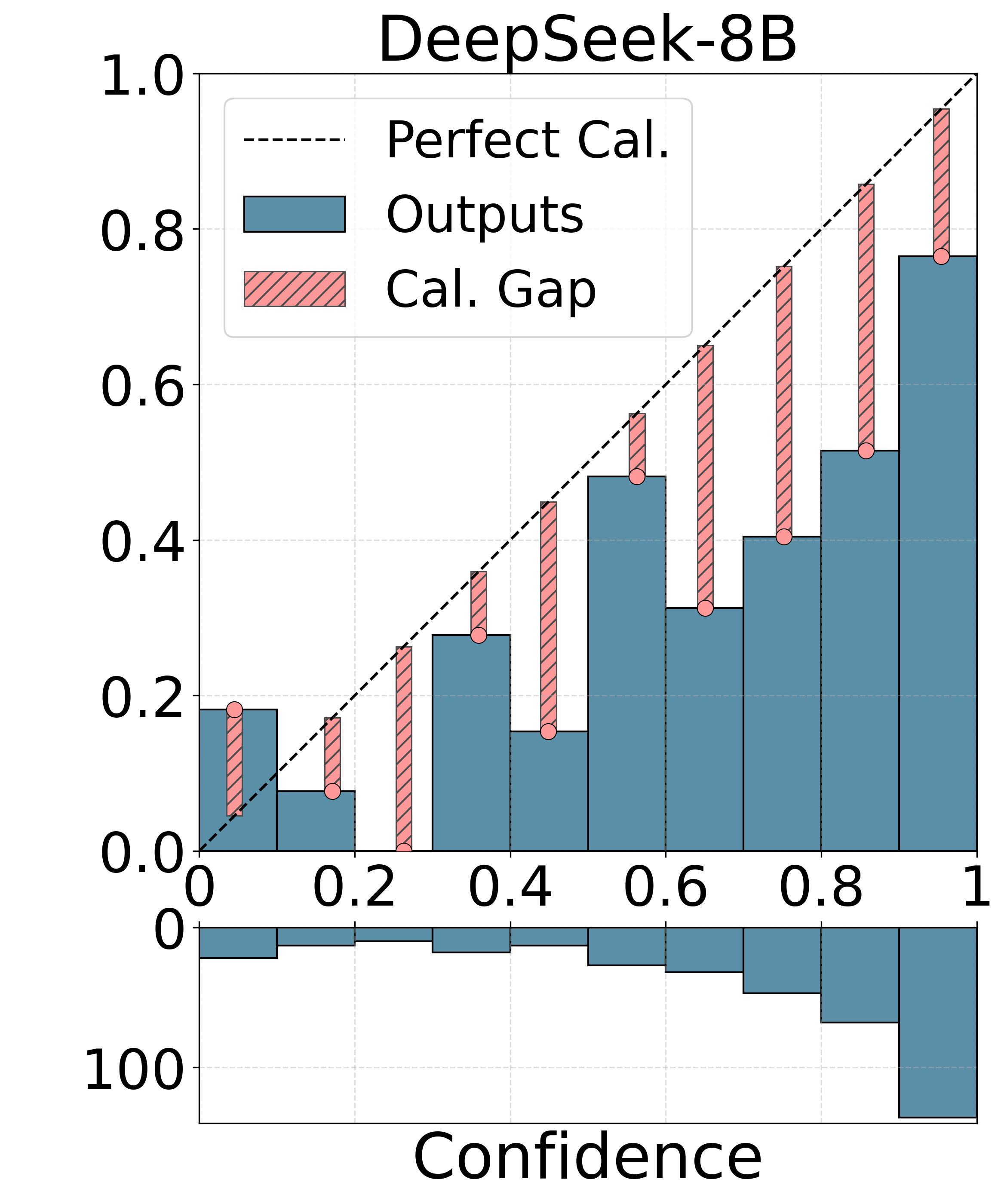}}\hfill
        \subfloat[ECE: 0.429]{\includegraphics[width=0.20\textwidth]{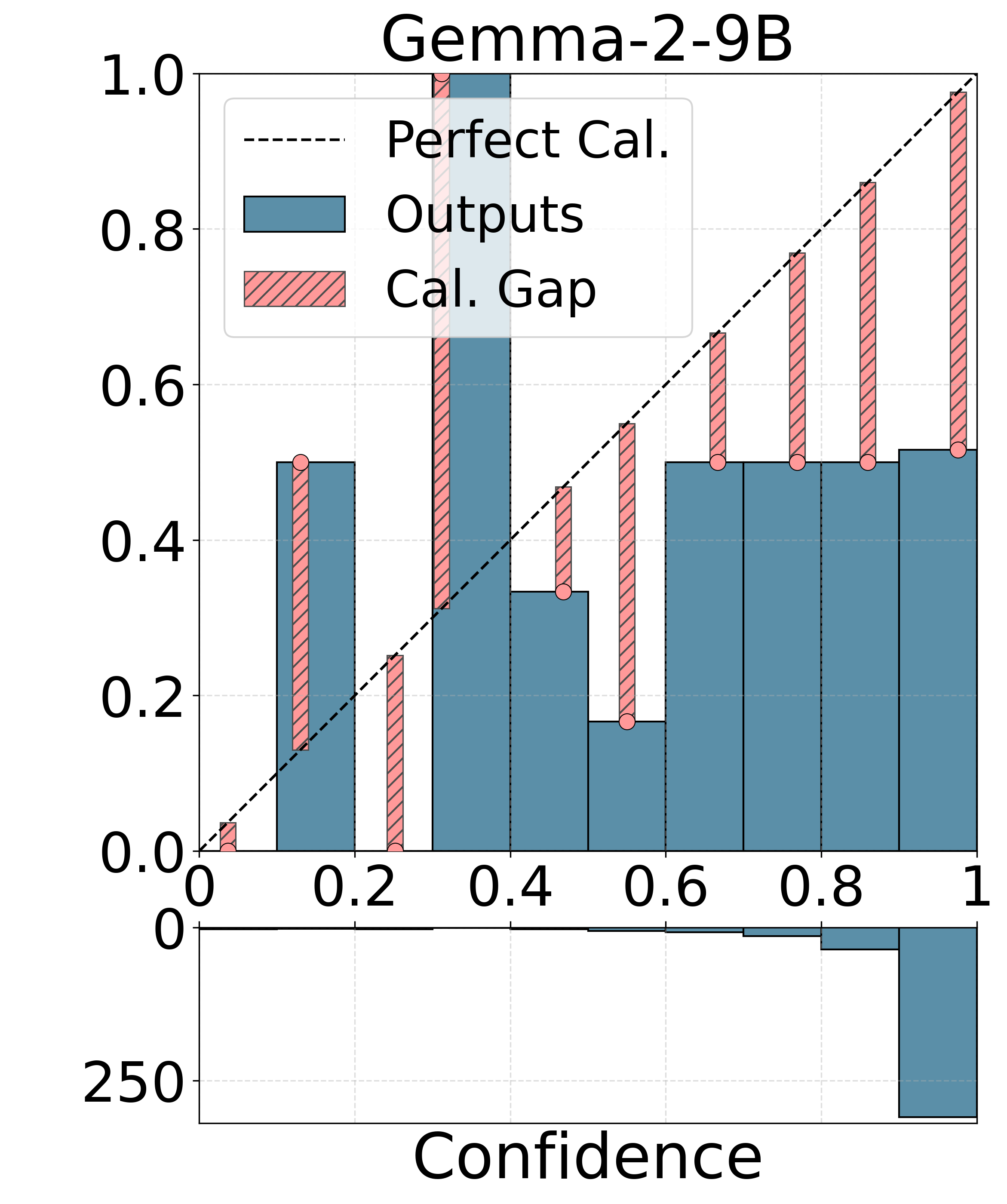}}\hfill
        \subfloat[ECE: 0.234]{\includegraphics[width=0.20\textwidth]{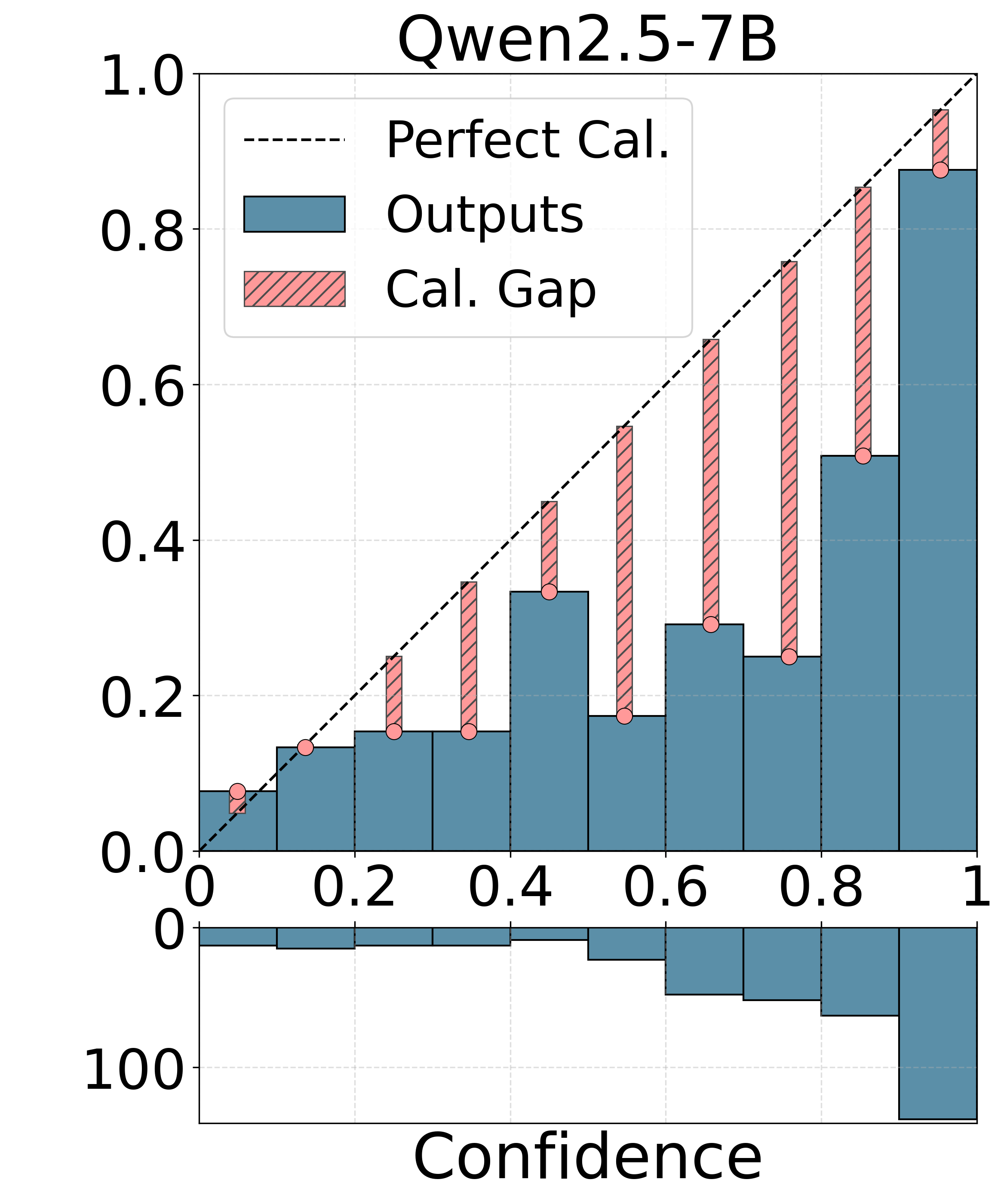}}
    \end{minipage}

    \vspace{-10pt} 

    \begin{minipage}{\textwidth}
        \centering
        \subfloat[ECE: 0.048]{\includegraphics[width=0.20\textwidth]{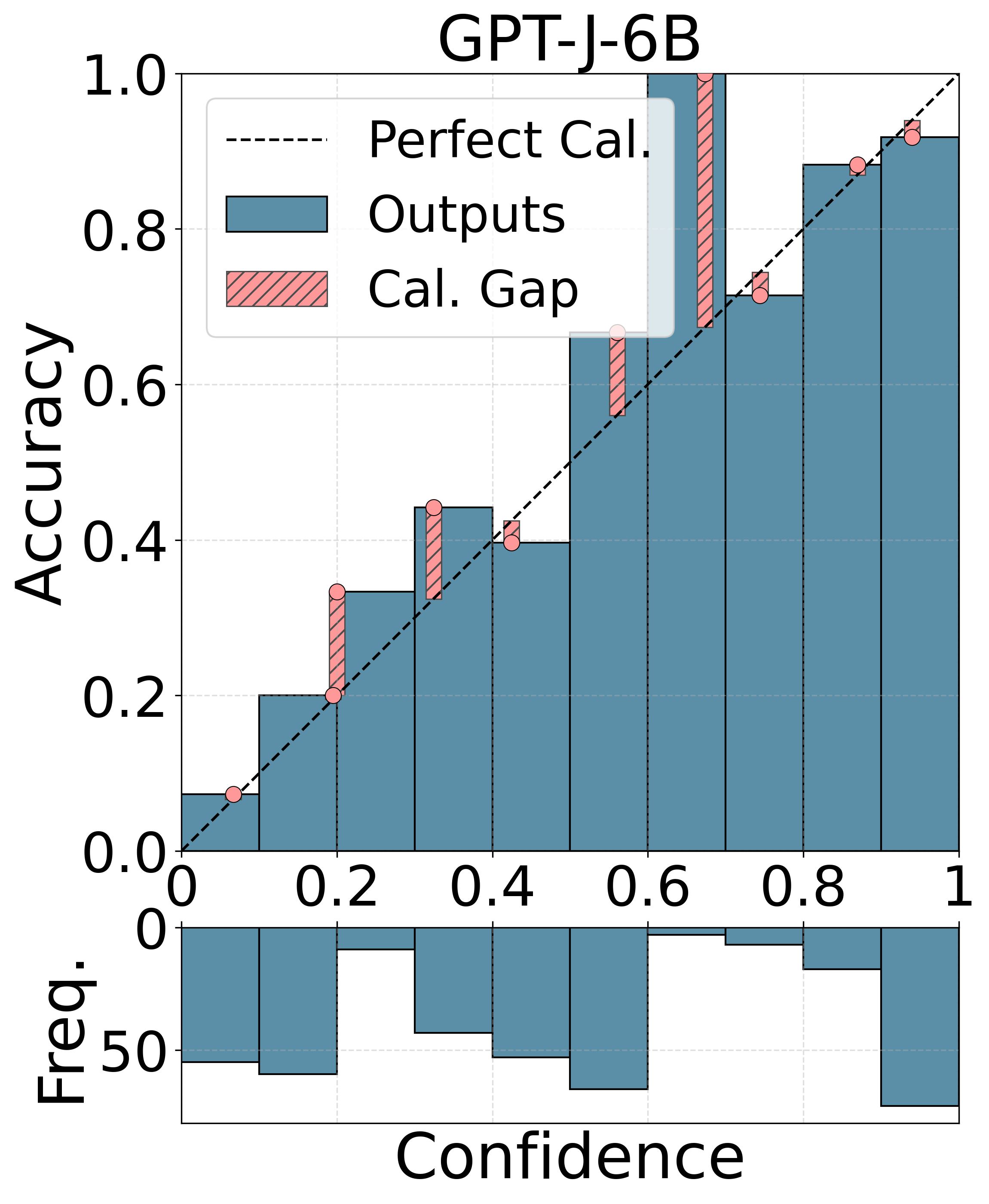}}\hfill
        \subfloat[ECE: 0.032]{\includegraphics[width=0.20\textwidth]{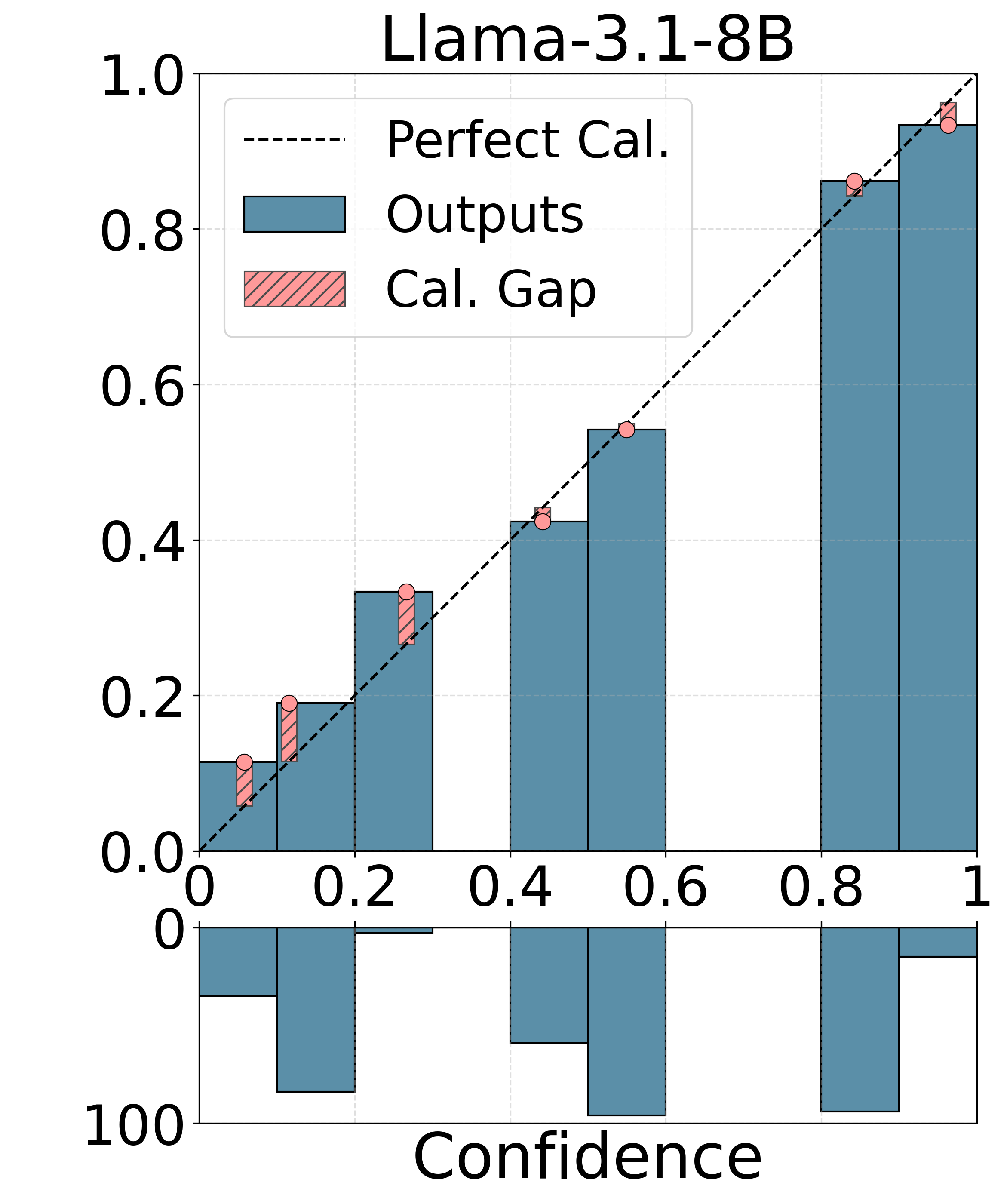}}\hfill
        \subfloat[ECE: 0.055]{\includegraphics[width=0.20\textwidth]{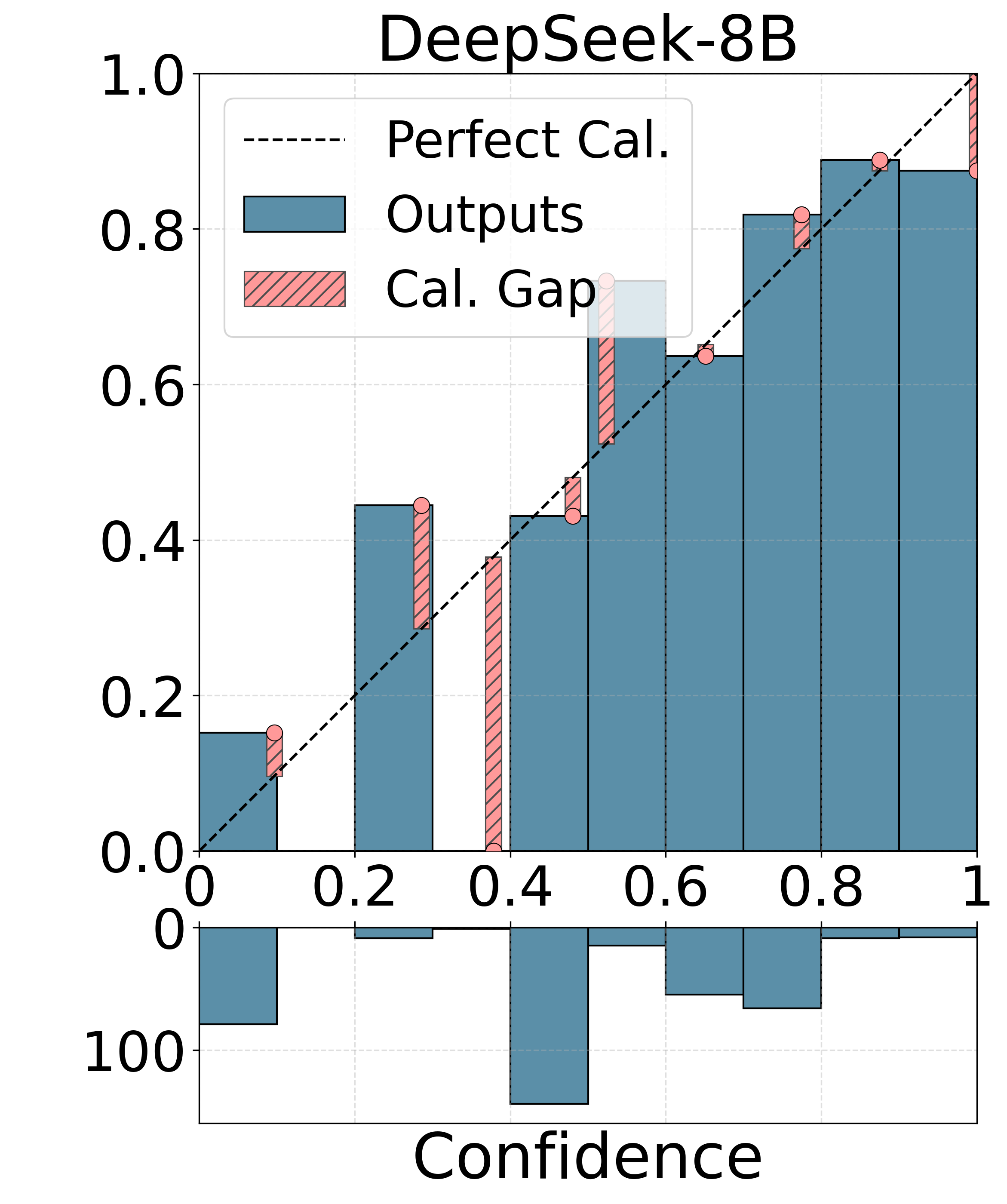}}\hfill
        \subfloat[ECE: 0.042]{\includegraphics[width=0.20\textwidth]{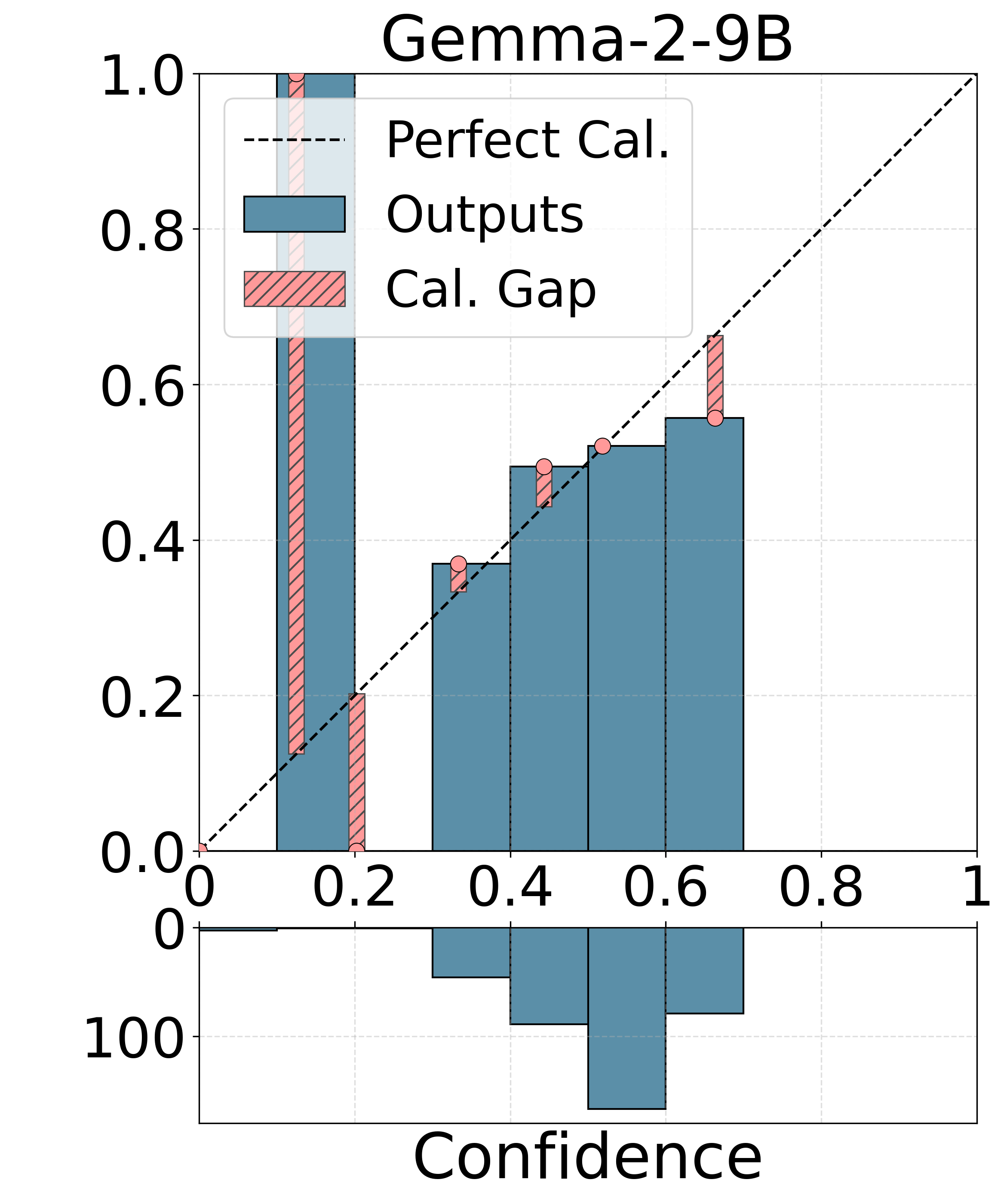}}\hfill
         \subfloat[ECE: 0.083]{\includegraphics[width=0.20\textwidth]{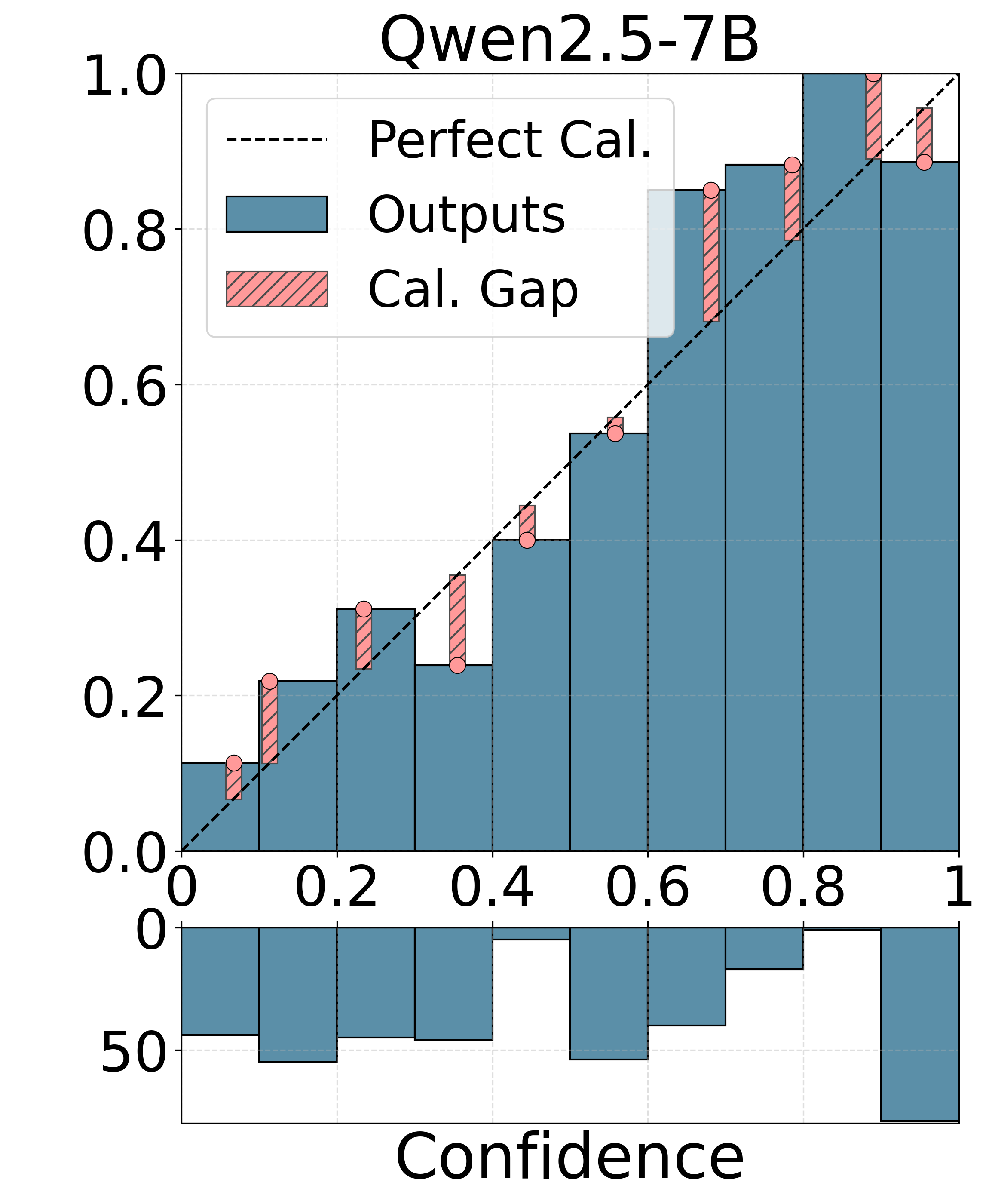}}
    \end{minipage}
      \caption{ Comparison of uncalibrated (Top) and calibrated with isotonic regression~\cite{zadrozny2002transforming} (Bottom) gender bias probabilities for WinoBias dataset. Instances are divided into 10 equal-width bins. The red bar shows the distance to perfectly.}
    \label{fig:calibration_winobias_iso}
\end{figure*}

\bibliography{aaai2026}

@article{platt2000probabilistic,
author = {Platt, John},
year = {2000},
month = {06},
pages = {},
title = {Probabilistic Outputs for Support Vector Machines and Comparisons to Regularized Likelihood Methods},
volume = {10},
journal = {Adv. Large Margin Classif.}
}

@inproceedings{nixon2019measuring,
  title={Measuring Calibration in Deep Learning.},
  author={Nixon, Jeremy and Dusenberry, Michael W and Zhang, Linchuan and Jerfel, Ghassen and Tran, Dustin},
  booktitle={CVPR workshops},
  volume={2},
  number={7},
  year={2019}
}

@article{chen2015microsoft,
  title={Microsoft coco captions: Data collection and evaluation server},
  author={Chen, Xinlei and Fang, Hao and Lin, Tsung-Yi and Vedantam, Ramakrishna and Gupta, Saurabh and Doll{\'a}r, Piotr and Zitnick, C Lawrence},
  journal={arXiv preprint arXiv:1504.00325},
  year={2015}
}

@article{sabir2023women, 
title={Women Wearing Lipstick: Measuring the Bias Between an Object and Its Related Gender}, 
author={Sabir, Ahmed and Padr{\'o}, Llu{\'\i}s}, 
journal={arXiv preprint arXiv:2310.19130},
year={2023}
}

@article{zhao2018gender,
  title={Gender bias in coreference resolution: Evaluation and debiasing methods},
  author={Zhao, Jieyu and Wang, Tianlu and Yatskar, Mark and Ordonez, Vicente and Chang, Kai-Wei},
  journal={arXiv preprint arXiv:1804.06876},
  year={2018}
}

@article{OpenAI,
  title={Chatgpt: Optimizing LMs for dialogue},
  author={OpenAI},
 url= {https://openai.com/blog/chatgpt},
  year={2022}
}

@InProceedings{rudinger-EtAl:2018:N18,
  author    = {Rudinger, Rachel  and  Naradowsky, Jason  and  Leonard, Brian  and  {Van Durme}, Benjamin},
  title     = {Gender Bias in Coreference Resolution},
  booktitle = {NAACL},
  month     = {},
  year      = {2018},
  address   = {},
  publisher = {}
}

@inproceedings{levesque2012winograd,
  title={The winograd schema challenge},
  author={Levesque, Hector and Davis, Ernest and Morgenstern, Leora},
  booktitle={Thirteenth international conference on the principles of knowledge representation and reasoning},
  year={2012}
}

@inproceedings{nangia2020crows,
  title={CrowS-Pairs: A Challenge Dataset for Measuring Social Biases in Masked Language Models},
  author={Nangia, Nikita and Vania, Clara and Bhalerao, Rasika and Bowman, Samuel},
  booktitle={EMNLP},
  pages={},
  year={2020}
}

@article{Staczak2021ASO,
  title={A Survey on Gender Bias in Natural Language Processing},
  author={Karolina Stańczak and Isabelle Augenstein},
  journal={ArXiv},
  year={2021},
  volume={},
}

@inproceedings{felkner2023winoqueer,
  title={WinoQueer: A Community-in-the-Loop Benchmark for Anti-LGBTQ+ Bias in Large Language Models},
  author={Felkner, Virginia K and Chang, Ho-Chun Herbert and Jang, Eugene and May, Jonathan},
  booktitle={ACL},
  year={2023}
}

@misc{gpt-j,
  author = {Wang, Ben and Komatsuzaki, Aran},
  title = {{GPT-J-6B: A 6 Billion Parameter Autoregressive Language Model}},
  howpublished = {\url{https://github.com/kingoflolz/mesh-transformer-jax}},
  year = 2021,
  month = May
}

@article{llama3modelcard,
  title = {Llama 3 Model Card},
  author = {{AI@Meta}},
  year = {2024},
  journal = {GitHub Repository},
  url = {https://github.com/meta-llama/llama3/blob/main/MODEL_CARD.md}
}

@misc{deepseekai2025,
      title={DeepSeek-R1: Incentivizing Reasoning Capability in LLMs via Reinforcement Learning}, 
      author={DeepSeek-AI},
      year={2025},
      eprint={2501.12948},
      archivePrefix={arXiv},
      primaryClass={cs.CL},
}

@inproceedings{cheng2023marked,
  title={Marked Personas: Using Natural Language Prompts to Measure Stereotypes in Language Models},
  author={Cheng, Myra and Durmus, Esin and Jurafsky, Dan},
  booktitle={ACL},
 pages={},
  year={2023}
}

@inproceedings{si2022re,
  title={Re-Examining Calibration: The Case of Question Answering},
  author={Si, Chenglei and Zhao, Chen and Min, Sewon and Boyd-Graber, Jordan},
  booktitle={Findings EMNLP},
  pages={},
  year={2022}
}

@inproceedings{guo2017calibration,
  title={On calibration of modern neural networks},
  author={Guo, Chuan and Pleiss, Geoff and Sun, Yu and Weinberger, Kilian Q},
  booktitle={ICML},
  pages={},
  year={2017},
  organization={}
}

@inproceedings{niculescu2005predicting,
  title={Predicting good probabilities with supervised learning},
  author={Niculescu-Mizil, Alexandru and Caruana, Rich},
  booktitle={ICML},
  pages={},
  year={2005}
}

@article{gallegos2024bias,
  title={Bias and fairness in large language models: A survey},
  author={Gallegos, Isabel O and Rossi, Ryan A and Barrow, Joe and Tanjim, Md Mehrab and Kim, Sungchul and Dernoncourt, Franck and Yu, Tong and Zhang, Ruiyi and Ahmed, Nesreen K},
  journal={Computational Linguistics},
  volume={},
  number={},
  pages={},
year={2024}
}

@article{stanovsky2019evaluating,
  title={Evaluating gender bias in machine translation},
  author={Stanovsky, Gabriel and Smith, Noah A and Zettlemoyer, Luke},
  journal={arXiv preprint arXiv:1906.00591},
  year={2019}
}

@article{team2024gemma,
  title={Gemma 2: Improving open language models at a practical size},
  author={Team, Gemma and Riviere, Morgane and Pathak, Shreya and Sessa, Pier Giuseppe and Hardin, Cassidy and Bhupatiraju, Surya and Hussenot, L{\'e}onard and Mesnard, Thomas and Shahriari, Bobak and Ram{\'e}, Alexandre and others},
  journal={arXiv preprint arXiv:2408.00118},
  year={2024}
}

@misc{Falcon3,
    title = {Falcon 3 family of Open Foundation Models},
    author = {TII Team},
    month = {December},
    year = {2024}
}

@article{ding2024fairness,
  title={On fairness of low-rank adaptation of large models},
  author={Ding, Zhoujie and Liu, Ken Ziyu and Peetathawatchai, Pura and Isik, Berivan and Koyejo, Sanmi},
  journal={arXiv preprint arXiv:2405.17512},
  year={2024}
}

@article{qwen2,
  title={Qwen2 Technical Report},
  author={An Yang and others},
  journal={arXiv preprint arXiv:2407.10671},
  year={2024}
}

@article{kadavath2022language,
  title={Language models (mostly) know what they know},
  author={Kadavath, Saurav and Conerly, Tom and Askell, Amanda and Henighan, Tom and Drain, Dawn and Perez, Ethan and Schiefer, Nicholas and Hatfield-Dodds, Zac and DasSarma, Nova and Tran-Johnson, Eli and others},
  journal={arXiv preprint arXiv:2207.05221},
  year={2022}
}

@article{xiong2023can,
  title={Can llms express their uncertainty? an empirical evaluation of confidence elicitation in llms},
  author={Xiong, Miao and Hu, Zhiyuan and Lu, Xinyang and Li, Yifei and Fu, Jie and He, Junxian and Hooi, Bryan},
  journal={arXiv preprint arXiv:2306.13063},
  year={2023}
}

@article{tian2023just,
  title={Just ask for calibration: Strategies for eliciting calibrated confidence scores from language models fine-tuned with human feedback},
  author={Tian, Katherine and Mitchell, Eric and Zhou, Allan and Sharma, Archit and Rafailov, Rafael and Yao, Huaxiu and Finn, Chelsea and Manning, Christopher D},
  journal={arXiv preprint arXiv:2305.14975},
  year={2023}
}

@inproceedings{kapoor2024calibration,
  title={Calibration-tuning: Teaching large language models to know what they don’t know},
  author={Kapoor, Sanyam and Gruver, Nate and Roberts, Manley and Pal, Arka and Dooley, Samuel and Goldblum, Micah and Wilson, Andrew},
  booktitle={UncertaiNLP 2024},
  pages={},
  year={2024}
}

@article{team2025gemma,
  title={Gemma 3 technical report},
  author={Team, Gemma and Kamath, Aishwarya and Ferret, Johan and Pathak, Shreya and Vieillard, Nino and Merhej, Ramona and Perrin, Sarah and Matejovicova, Tatiana and Ram{\'e}, Alexandre and Rivi{\`e}re, Morgane and others},
  journal={arXiv preprint arXiv:2503.19786},
  year={2025}
}

@inproceedings{kaukonen2025aunt,
  title={How aunt-like are you? exploring gender bias in the genderless estonian language: A case study},
  author={Kaukonen, Elisabeth and Sabir, Ahmed and Sharma, Rajesh},
  booktitle={Proceedings of the Joint 25th Nordic Conference on Computational Linguistics and 11th Baltic Conference on Human Language Technologies (NoDaLiDa/Baltic-HLT 2025)},
  pages={296--301},
  year={2025}
}

@inproceedings{kull2017beta,
  title={Beta calibration: a well-founded and easily implemented improvement on logistic calibration for binary classifiers},
  author={Kull, Meelis and Silva Filho, Telmo and Flach, Peter},
  booktitle={Artificial intelligence and statistics},
  pages={623--631},
  year={2017},
  organization={PMLR}
}

@inproceedings{zadrozny2002transforming,
  title={Transforming classifier scores into accurate multiclass probability estimates},
  author={Zadrozny, Bianca and Elkan, Charles},
  booktitle={Proceedings of the eighth ACM SIGKDD international conference on Knowledge discovery and data mining},
  pages={694--699},
  year={2002}
}

@article{chen2022close,
  title={A close look into the calibration of pre-trained language models},
  author={Chen, Yangyi and Yuan, Lifan and Cui, Ganqu and Liu, Zhiyuan and Ji, Heng},
  journal={arXiv preprint arXiv:2211.00151},
  year={2022}
}

@inproceedings{krause2023confidently,
  title={Confidently wrong: exploring the calibration and expression of (Un) certainty of large language models in a multilingual setting},
  author={Krause, Lea and Tufa, Wondimagegnhue and Santamar{\'\i}a, Selene B{\'a}ez and Daza, Angel and Khurana, Urja and Vossen, Piek},
  booktitle={MM-NLG 2023},
  pages={},
  year={2023}
}

@inproceedings{roelofs2022mitigating,
  title={Mitigating bias in calibration error estimation},
  author={Roelofs, Rebecca and Cain, Nicholas and Shlens, Jonathon and Mozer, Michael C},
  booktitle={International Conference on Artificial Intelligence and Statistics},
  pages={4036--4054},
  year={2022},
  organization={PMLR}
}

@inproceedings{naeini2015obtaining,
  title={Obtaining well calibrated probabilities using bayesian binning},
  author={Naeini, Mahdi Pakdaman and Cooper, Gregory and Hauskrecht, Milos},
  booktitle={AAAI},
  volume={},
  number={},
  year={2015}
}

@article{brier1950verification,
  title={Verification of forecasts expressed in terms of probability},
  author={Brier, Glenn W},
  journal={Monthly weather review},
  volume={78},
  number={1},
  pages={1--3},
  year={1950}
}

@article{kull2019beyond,
  title={Beyond temperature scaling: Obtaining well-calibrated multi-class probabilities with dirichlet calibration},
  author={Kull, Meelis and Perello Nieto, Miquel and K{\"a}ngsepp, Markus and Silva Filho, Telmo and Song, Hao and Flach, Peter},
  journal={NeurIPS},
  volume={},
  year={2019}
}

@article{murphy1977reliability,
  title={Reliability of subjective probability forecasts of precipitation and temperature},
  author={Murphy, Allan H and Winkler, Robert L},
  journal={Journal of the Royal Statistical Society Series C: Applied Statistics},
  volume={26},
  number={1},
  pages={41--47},
  year={1977},
  publisher={Oxford University Press}
}

@article{sabir2025exploring,
  title={Exploring Gender Bias Beyond Occupational Titles},
  author={Sabir, Ahmed and Sharma, Rajesh},
  journal={arXiv preprint arXiv:2507.02679},
  year={2025}
}

@article{minderer2021revisiting,
  title={Revisiting the calibration of modern neural networks},
  author={Minderer, Matthias and Djolonga, Josip and Romijnders, Rob and Hubis, Frances and Zhai, Xiaohua and Houlsby, Neil and Tran, Dustin and Lucic, Mario},
  journal={NeurIPS},
  year={2021}
}

@inproceedings{lum2025bias,
  title={Bias in Language Models: Beyond Trick Tests and Towards RUTEd Evaluation},
  author={Lum, Kristian and Anthis, Jacy Reese and Robinson, Kevin and Nagpal, Chirag and D’Amour, Alexander Nicholas},
  booktitle={ACL},
  pages={},
  year={2025}
}

@article{cohen1960coefficient,
  title={A coefficient of agreement for nominal scales},
  author={Cohen, Jacob},
  journal={Educational and psychological measurement},
  volume={},
  number={},
  pages={},
  year={1960},
  publisher={Sage Publications Sage CA: Thousand Oaks, CA}
}

@article{yoon2025reasoning,
  title={Reasoning models better express their confidence},
  author={Yoon, Dongkeun and Kim, Seungone and Yang, Sohee and Kim, Sunkyoung and Kim, Soyeon and Kim, Yongil and Choi, Eunbi and Kim, Yireun and Seo, Minjoon},
  journal={arXiv preprint arXiv:2505.14489},
  year={2025}
}

\end{document}